%% file: acl_latex.tex
\pdfoutput=1

\documentclass[11pt]{article}

\usepackage[preprint]{acl}

\usepackage{times}
\usepackage{latexsym}

\usepackage{graphicx} 
\usepackage{tabularray}
\usepackage{tabularx}
\usepackage{booktabs}
\usepackage{multicol}
\usepackage{multirow}

\usepackage{tcolorbox}
\usepackage{multicol}
\usepackage{todonotes}
\usepackage{xcolor}
\usepackage{amsmath}
\usepackage{tcolorbox}
\usepackage{graphicx}
\usepackage{amssymb}

\usepackage[T1]{fontenc}

\usepackage[utf8]{inputenc}

\usepackage{microtype}

\usepackage{inconsolata}

\usepackage{graphicx}

\usepackage[whole]{bxcjkjatype}

\usepackage{CJKutf8}
\newcommand{\Chinese}[1]{{\begin{CJK*}{UTF8}{gbsn}#1\end{CJK*}}}

\usepackage[T2A,T1]{fontenc}

\title{Towards Cross-Lingual Explanation of Artwork \\ in Large-scale Vision Language Models}

\author{%
  Shintaro Ozaki${ }^{\alpha}$  \hspace{30pt}
  Kazuki Hayashi${ }^{\alpha}$ \hspace{30pt}
  Yusuke Sakai${ }^{\alpha}$ \hspace{30pt} 
  \\ [4pt]
  \hspace{-10pt}
  \textbf{Hidetaka Kamigaito}${}^{\alpha}$ \hspace{15pt}
  \textbf{Katsuhiko Hayashi}${}^{\beta}$ \hspace{15pt}
  \textbf{Taro Watanabe}${}^{\alpha}$  \\  [10pt]
    ${}^{\alpha}$ Nara Institute of Science and Technology (NAIST)\hspace{7pt}
${}^{\beta}$ The University of Tokyo \\ [2pt]
\texttt{\{ozaki.shintaro.ou6, kamigaito.h, taro.watanabe\}@naist.ac.jp} \\ [3pt]
}

\begin{document}
\maketitle
\begin{abstract}
As the performance of Large-scale Vision Language Models (LVLMs) improves, they are increasingly capable of responding in multiple languages, and there is an expectation that the demand for explanations generated by LVLMs will grow.
However, pre-training of Vision Encoder and the integrated training of LLMs with Vision Encoder are mainly conducted using English training data, leaving it uncertain whether LVLMs can completely handle their potential when generating explanations in languages other than English.
In addition, multilingual QA benchmarks that create datasets using machine translation have cultural differences and biases, remaining issues for use as evaluation tasks.
To address these challenges, this study created an extended dataset in multiple languages without relying on machine translation. 
This dataset that takes into account nuances and country-specific phrases was then used to evaluate the generation explanation abilities of LVLMs. 
Furthermore, this study examined whether Instruction-Tuning in resource-rich English improves performance in other languages. 
Our findings indicate that LVLMs perform worse in languages other than English compared to English. 
In addition, it was observed that LVLMs struggle to effectively manage the knowledge learned from English data\footnote{Our data is publicly available at \url{https://huggingface.co/datasets/naist-nlp/MultiExpArt}.}.
\end{abstract}

\begin{figure}[t]
    \centering
    \includegraphics[width=\linewidth]{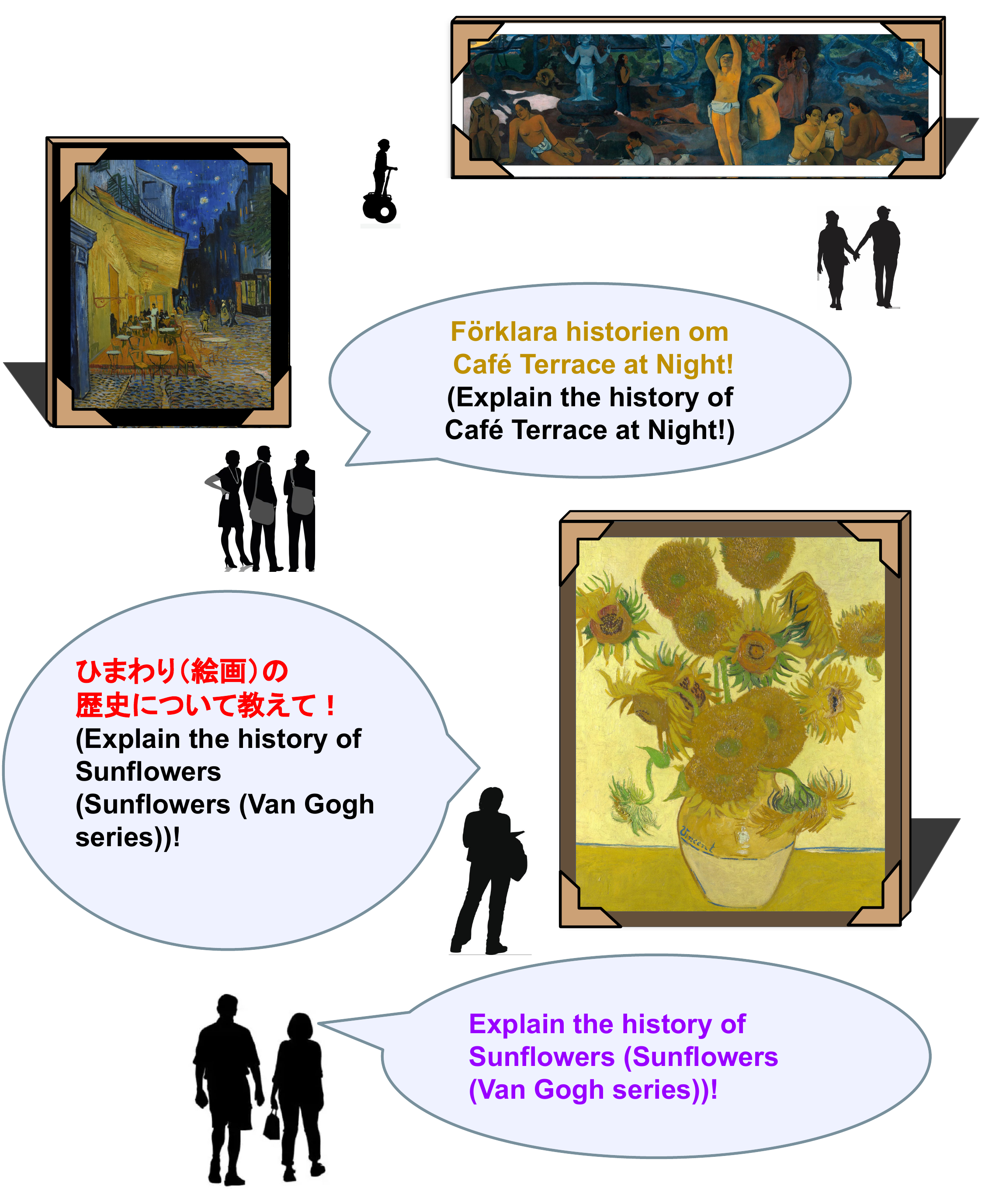}
    \caption{An example of situations that require multilingual and explanation skills.}
    \label{fig:top}
\end{figure}

\section{Introduction}
Each artwork, e.g., image, has a unique title, making it suitable for evaluating Large-scale Vision Language Models (LVLMs) that handle both the image and the text.
\citet{hayashi2024artwork} focused on artwork explanation generation to investigate the relationship between language-based and vision-based knowledge of LVLMs using English data. 
When using LVLMs for creative support,  explanation generation abilities are required based on the composition and ingenuity of the image, e.g., comparisons with other works, historical background, and deep artistic knowledge.
LVLMs enable image and text aware tasks exactly, e.g., determining the color of traffic lights in the image and judging if it is possible to proceed, by integrating Vision Encoder~\cite{li2023blip}, e.g., Vision Transformer (ViT)~\cite{dosovitskiy2020image}, which processes image data into high-dimensional features, and Large Language Models (LLMs)~\cite{llmjp2024llmjpcrossorganizationalprojectresearch}, which can handle natural language, through additional training.
This allows LVLMs to understand instructions with image inputs by humans and generate responses based on those instructions and they have archived remarkable performance on Vision \& Language (V\&L) benchmarks~\cite{liu2023mmbench, li2023seed, ozaki2024bqabodylanguagequestion}.

However, there are remaining issues with training current LVLMs when dealing with multilingual data. 
Training and evaluation of LVLMs often use English data, leaving questions on the performance on other languages, and there exists no standard protocol especially when evaluating the performance of multilingual image understanding tasks.
Several multilingual image QA tasks do exist~\cite{changpinyo-etal-2023-maxm,nguyen2023vlsp2022}, but they primarily rely on machine translation, making it uncertain whether country-specific cultural nuances or biases are completely considered.
Specifically, when creating multilingual QA tasks, \citet{sakai2024mcsqa} pointed out that multiple concepts e.g., 'roast', 'grill', 'broil', 'toast', and 'bake' in English could be potentially translated into only one expression e.g., '焼く' in Japanese.
Thus, it is necessary to construct a completely fair multilingual evaluation dataset for explanation generation abilities.
The issue is, in particular, compounded in the field of art, since an explanation of an image may vary across countries, leading to different explanations due to the impression of the image in other countries.
Simply translating from resource-rich languages like English into other languages using machine translation to create datasets fails to account for cultural nuances.
For example, ``Mona Lisa'' is translated directly into Chinese and Japanese correctly, but in Spanish, it is translated as ``Mona Lisa'' even though it is called “La Gioconda” in Spanish.
Moreover, since these QA datasets do not evaluate the ability to generate explanations, there are no appropriate metrics to evaluate the explanation generation abilities of LVLMs across different languages.

To solve the lack of datasets that can evaluate the ability to generate explanations in other languages and the inability to account for country-specific cultural nuances or biases by simply using machine translation to create datasets, we created datasets that allow you to evaluate the ability to generate explanations in other languages without machine translation using Wikipedia.
\citet{hayashi2024artwork} focused only on English, but our study expanded this work to ten languages (Chinese, Dutch, English, French, German, Italian, Japanese, Russian, Spanish, and Swedish).

We utilized these datasets to analyze the multilingual performance of current LVLMs in generation explanation abilities related to artworks with three settings which are Alignment-10, Alignment-5, and Full tasks, and investigated whether LVLMs can maintain equal generation explanation abilities in artworks when extended to ten languages.
To investigate LVLMs' multilingual generation explanation abilities, we hypothesized that ``the integrated training of LVLMs and the pre-training of Vision Encoder are mainly trained in English data, limiting their ability to achieve optimal performance when handling other languages.''
Moreover, we also conducted Instruction-Tuning in English-only training data for two of the models so that validate the extent to which these two models can acquire explanation generation capabilities in other languages solely from English training data.

We found that LVLMs perform best when given instructions in English and generating output in English, while their performance declines when instructions or output are in languages other than English. Moreover, we observed that outputting in the same language as the instructions like Japanese instruction with its Japanese response leads to better performance than the response in English for Japanese instruction, indicating that LVLMs struggle to effectively utilize the knowledge learned in English when applied to other languages. 
The result also showed that performance was further worse with Instruction-Tuning conducted in English.
These findings support our hypothesis and suggests that it is necessary to let Vision Encoder train not only English training data but also other language data.

\input{tables/table2}

\section{Related Work}
\label{related work}

\paragraph{LVLMs}
In general, an LVLM comprises a Vision Encoder that processes visual information and an LLM pre-trained on a large amount of textual data.
They are trained using contrastive learning~\cite{chen2020simple}, aiming to integrate visual and linguistic information. 
Vision Encoder is a model trained to encode images and visual data, typically using architectures such as ResNet~\cite{he2015deep} or Vision Transformer (ViT)~\cite{dosovitskiy2020image}. 
On the other hand, LLMs are models pre-trained on a large text dataset, with prominent examples including Qwen~\cite{qwen, qwen2}, LLaMA~\cite{touvron2023llama, touvron2023llama2, dubey2024llama}, Gemini~\cite{team2023gemini, reid2024gemini} and GPT~\cite{brown2020language, ouyang2022training, dettmers2022gpt3}
LVLMs such as Qwen-VL~\cite{Qwen-VL}, and LLaVA-NeXT~\cite{liu2024llavanext} are examples of integrated models. These models achieve visual and natural language integration by acquiring features from images through the Vision Encoder and textual features through LLMs and then performing additional training with the goal of integrating vision and language.

\paragraph{LVLMs \& Knowledge}
Whether the visual knowledge learned by the Vision Encoder and the linguistic knowledge learned by LLMs are properly aligned remains mostly unclear~\cite{li2022blip, li2023blip}. 
Especially for generating explanations involving knowledge about artwork, which this study focuses on, it is essential to systematically align and utilize both types of knowledge~\cite{hayashi2024artwork}.
This requires the integration of visual knowledge (e.g., visual features of specific artworks) and linguistic knowledge (e.g., historical background and technical details about those artworks).
In LVLMs, the integration of Vision Encoder and LLMs are achieved by adding partial networks, but this alone makes it challenging to properly align visual and linguistic knowledge.
In domains requiring sophisticated knowledge, such as artwork, improper alignment can degrade the quality of generated explanations.
Thus, while this study aims to integrate visual and linguistic information and build efficient models using contrastive learning, it also indicates that further research is necessary to achieve proper alignment of visual and linguistic knowledge.

\begin{figure*}[t]
\centering
\includegraphics[width=\textwidth]{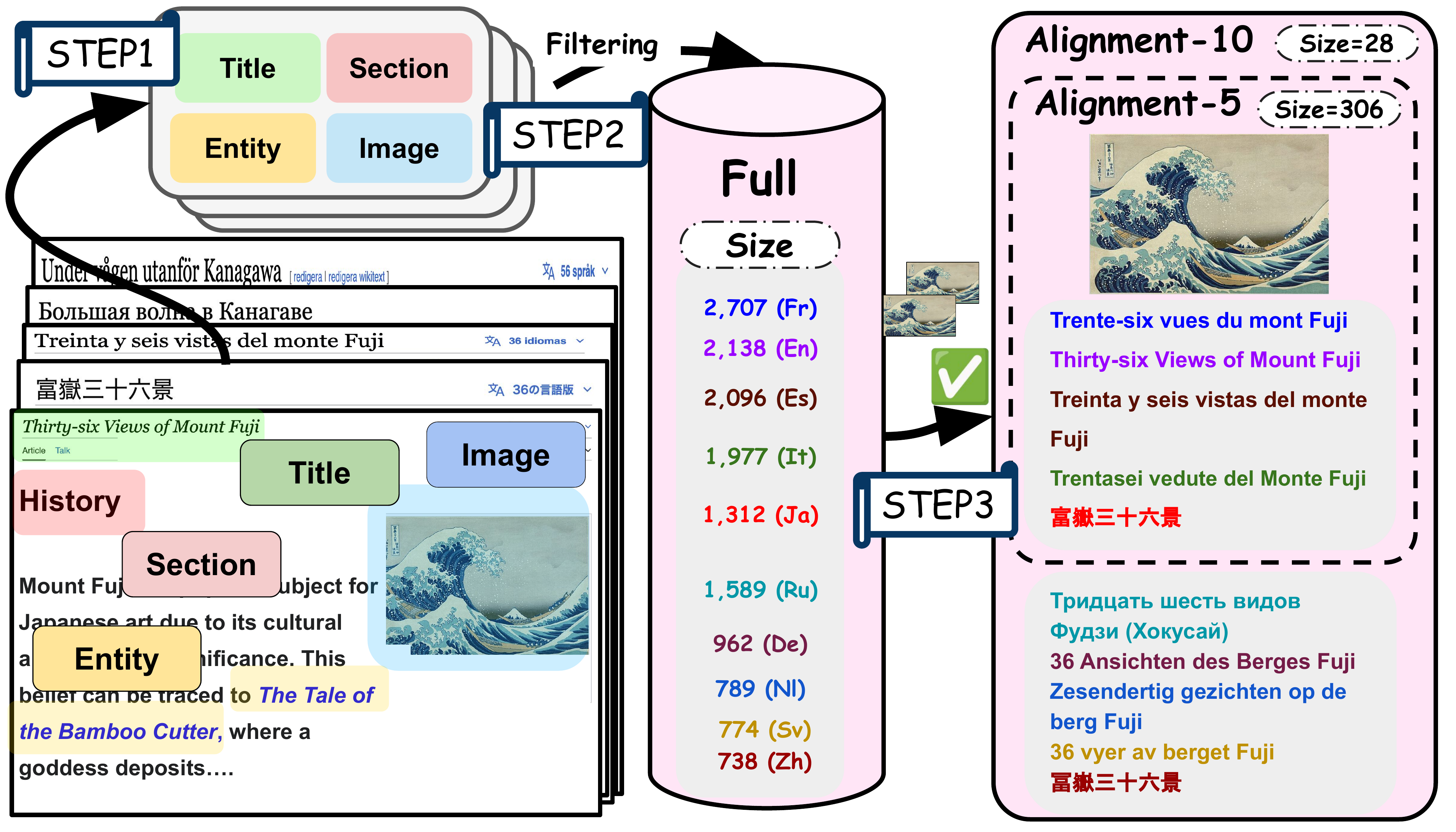}
\caption{How to make datasets from Wikipedia. As shown in Section~\ref{dataset creation}, we extracted and filtered Wikipedia pages about artworks. We then manually identified pages with titles and images common across ten languages.}
\label{fig:make datasets}
\end{figure*}

\paragraph{LVLMs \& Multilingual}
As we mentioned earlier, LVLMs follow human instructions through integrated learning of Vision Encoder~\cite{li2023blip} and LLMs trained by a large amount of English training data. However, it is unclear whether LVLMs are able to really understand and output properly when input from languages other than English.
On the other hand, as far as evaluation tasks such as XGQA~\cite{pfeiffer-etal-2022-xgqa} they expanded the English GQA dataset into seven languages through translation. However, because this expansion relies on translations from English, it likely includes QA pairs that do not consider the cultural contexts of the target languages.
For instance, MaXM~\cite{changpinyo-etal-2023-maxm} collects large data sets by translating non-English language data into English, which is then back-translated into seven languages.
Similarly, EVJVQA~\cite{nguyen2023vlsp2022} creates around 33,000 QA pairs from approximately 5,000 images taken in Vietnam, but the translations still retain biases unique to Vietnamese culture and norms.
In our research, we mitigated these biases by focusing on artworks, preventing the introduction of a specific culture to any country within the images. (i.e., There are countries where cars drive on the right lane and others where they drive on the left.)
Since artworks have unique and definitive relationships between the title and its image, we also create datasets from relatively resource-rich Wikipedia in various languages without relying on machine translation. Our study is not a Question Answering task, such as VQA~\cite{VQA}, but an explanation task, which requires LVLMs to explain images correctly.
We evaluated an explanation-generaton task in ten languages expanding~\citet{hayashi2024artwork}.

\section{Task}
Our task is ``Analyzing the multilingual performance of LVLMs in explaining artworks''.
To tackle this, we measured explanatory capabilities using three settings (Alignment-10, Alignment-5, Full) which were described below.
In addition, we prepared templates for input when evaluating explanation abilities with datasets we created.

\paragraph{Alignment-10}
We created datasets composed only of data with the same images and titles across ten languages from the Full task, which have a total of 28 pages.
In other words, this dataset contains the same images with titles represented in the language of each country, allowing for an equal evaluation of description generation capabilities across the ten languages.

\paragraph{Alignment-5}
To mitigate the data scarcity issue in Alignment-10, Alignment-5 restricts the target languages to five specific languages.
The total number of data is 306, and this dataset is used to compare explanation generation abilities across the five languages.
To cover a diverse range of language families, we selected English, Spanish, French, Italian, and Japanese as Alignment-5 task.

\paragraph{Full}
To further mitigate the data scarcity issues in the above settings, Full ignores the correspondence of artworks between languages and treats each language independently.
For details on the number of data, refer to Table~\ref{tab:full_task_data} or Figure~\ref{dataset creation}.
By using the Full task, we aimed to evaluate the differences in performance.

\input{results/a10-en-en}

\paragraph{Templates}
We prepared templates for evaluating explanation generation abilities using the datasets created from three tasks mentioned above. The process is as follows:
1) We prepared four patterns of templates for each of ten languages to mitigate the variance issue in performance evaluation \cite{sakai-etal-2024-toward}. 
In templates, we referred to the study by~\citet{hayashi2024artwork}, selecting four patterns with clearly different grammatical structures to avoid a lack of diversity.
~\citet{sakai2024mcsqa} noted that not choosing distinctly different patterns may result in differences originally present in English being lost in translation;
2) We let ChatGPT\footnote{\url{https://openai.com/chatgpt/}} translate the obtained templates into ten languages.
We chose to use LLMs rather than translation tools because LLMs are thought to better understand and translate including nuances; 
3) Even with translations taking into nuances by ChatGPT, there may be variations in quality between languages.
To solve this, we asked nine native speakers of ten languages, to check whether the templates translated back into English maintained the same nuance and level of difficulty.
This process ensured that all 10 language templates created in this study have the same level of difficulty; Of course, it might be possible to crowdsource this task using platforms like MTurk\footnote{\url{https://www.mturk.com/}}, but asking annotators simply ``Is this translation correct including nuances?'' may not lead to serious engagement with the translation checking task.
For examples of the each language template, refer to Table~\ref{tab:prompt_template}.
During inference, the model runs as many inferences as the total number of combinations of section, subsection, and subsubsection. The appendix provides the exact number of inferences in Table~\ref{tab:full_task_data}.

\input{results/a10-en-lang}
\input{results/a10-lang-lang}

\section{Dataset Creation}
\label{dataset creation}
For each of ten languages, the following steps were taken to create the dataset.
Ten languages were determined based on having a higher number of Wikipedia articles than the total number of articles.

\paragraph{\textbf{STEP1: Extracting Data from Wikipedia}}
We collected Artwork articles from the English Wikipedia Infobox. Articles with the same title in nine other languages are identified to create corresponding articles in those languages.
Hyperlinked strings within the articles are extracted as entities related to artworks. The description includes four types of information: the image, the title, hierarchical information from the article (Section, Subsection, Sub subsection), and the extracted entities.

\paragraph{\textbf{STEP2: Filtering and Formatting}}
From the collected articles, those without images were excluded.
Any articles that had domains but no actual pages on Wikipedia were also removed.
This process completes the dataset used for the Full task.

\paragraph{\textbf{STEP3: Adjusting}}
For the Alignment-10 and Alignment-5 tasks, we extracted pages from other languages that have the same titles as the English Wikipedia pages to ensure alignment across ten or five languages. 
These non-English pages include the English titles in their metadata, and alignment is successful when these English titles match exactly.
To eliminate differences between languages, a manual verification is conducted to ensure that all articles contain images of the same artwork.
Variations in image size are permitted, but all images must represent the same artwork across languages.
The datasets for Alignment-10 and Alignment-5 are prepared accordingly, using images from the English articles for alignment.

\paragraph{\textbf{STEP4: Data Splitting}}
To measure the explanation generation abilities of LVLMs, the following approach is used:
(1) For the Alignment task, all data was treated as test set.
(2) For the Full task, nine non-English languages are used for test set, while English data is divided into train, dev, and test sets.
To avoid biases arising from the popularity of artworks in the LVLM's training data, we shuffled the English data based on six indicators: page views, number of links, number of edits, number of references, number of language versions, and article length~\cite{hayashi2024artwork}.
The data was ranked according to these indicators, and the test, valid, and train data were split in a 2:2:6 ratio to maintain average rankings.
The data used in the Alignment task was included in the test set.

\begin{figure*}[t]
    \centering
    \includegraphics[width=\textwidth]{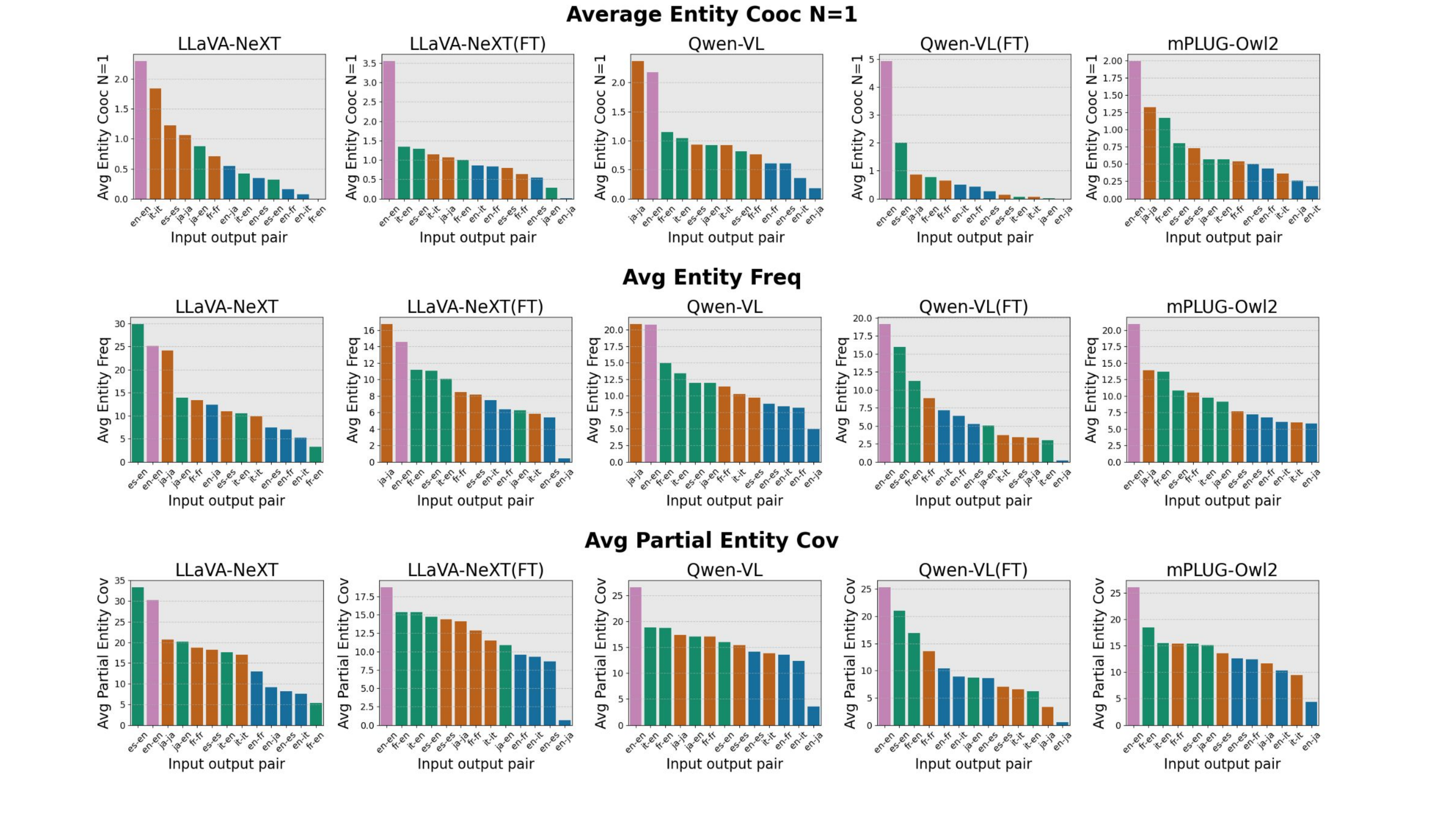}
    \caption{Some of the results in the Alignment-5 task.
    Purple bin indicates the method which is the instruction and the output in English (\{En\}-\{En\}), Green bin indicates the instruction in languages other than English and the output in English (\{Lang\}-\{En\}), Brown bin indicates the instruction and output in languages other than English (\{Lang\}-\{Lang\}) and Blue bin indicates the instruction in English and the output in languages other than English (\{En\}-\{Lang\}).
    From this figure, it can be seen that the English instructions are optimal, even if the number of data is expanded. We described further detailed results in Table~\ref{tab:result-score-5} including Phi-3 and XComposer2. You can see the rest of the results in Figure~\ref{fig:a5-figure-appendix} in the Appendix.}
    \label{fig:a5-figure}
\end{figure*}

\section{Experiments}
\label{Setup}
\subsection{Evaluation Metrics}
This study adopted three evaluation metrics proposed by~\citet{hayashi2024artwork} and also described these metrics more details in Appendix~\ref{sec:appendix:metric}. We also utilize popular metrics in NLG for evaluation, i.e., BLEU~\cite{papineni2002bleu}, ROUGE~\cite{lin2004rouge}, and BERTScore~\cite{zhang2019bertscore}.

\paragraph{Entity Coverage, Entity F1, and Entity Cooccurrence}  
These metrics evaluate how well the generated text incorporates entities related to the artwork and how accurately it reflects the relationships between these entities proposed by~\citet{hayashi2024artwork}.
Entity Coverage measures the inclusion of relevant entities in both exact and partial matches.
Entity F1 assesses the frequency and appropriateness of entity usage by comparing the generated text with reference explanations, inspired by the ROUGE metric.
Entity Cooccurrence goes a step further by examining how entities are contextually combined across sentences, considering their co-occurrence within the entire text, and applying brevity penalties to avoid inflated coverage in longer explanations.

\subsection{Models and Others}
We chose five models with relatively high performance: mPLUG-Owl2~\cite{ye2024mplug}, LLaVA-NeXT~\cite{liu2023llava, liu2024llavanext, liu2023improvedllava}, XComposer2~\cite{dong2024internlm}, Phi-3~\cite{abdin2024phi}, and Qwen-VL~\cite{Qwen-VL}. In addition, LLaVA-NeXT and Qwen-VL were conducted LoRA Tuning~\cite{hu2022lora} with English train data and included in the evaluation. Detailed experimental settings are described in Appendix~\ref{experiment-settings}.
This approach is based on the observation that current LLMs perform better when instructions are given in English~\cite{putri2024can}.
As far as Alignment tasks, we validated four patterns of input: \{En, Lang\}-\{En, Lang\}.
This indicates that when the input is English, the output can be directed to English or another language.
The same thing can also be done when the input is another language, and these four patterns were tested in this study.
By testing these patterns, we verify whether LVLMs perform better when supported in English, and whether having the output in English is a meaningful instruction.
As far as tokenizing words, we used SpaCy\footnote{\url{https://spacy.io/}} as a multilingual tokenizer, tokenizing each language to perform segmentation.
Thus, each language is expected to be divided into optimal token units.

\subsection{Results}
\label{results}
From the experiments conducted with Alignment-10, the method let LVLMs generate in English with English (\{En\}-\{En\}) results are listed in Table~\ref{tab:result-score-10-en-en}, the method which is instruction in English and output in other languages (\{En\}-\{Lang\}) results in Table~\ref{tab:result-score-10-en-lang}, and the instruction and output in other same languages (\{Lang\}-\{Lang\}) results in Table~\ref{tab:result-score-10-lang-lang}. The results for Phi-3 and XComposer2 are described in the Table~\ref{tab:result-score-10-appendix} in Appendix.
Overall, the results confirm that giving instructions in English and letting them generate output in English (i.e., \{En\}-\{En\}) maximizes the performance of LVLMs.
On the other hand, LoRA Tuning increased the value of Entity Cooccurrance, while other values decreased. 
This suggests that LoRA Tuning enabled LVLMs to understand and explain the context, but prevented entities from appearing in the generated sentences.
Furthermore, looking at the results of Alignment-5 in Table~\ref{tab:result-score-5} in Appendix, where the number of data was expanded, the outputs that used English instructions and outputs were generally higher, followed by those using instructions and outputs in other languages. This is consistent with the results of Alignment-10.
In addition, Figure~\ref{fig:a5-figure} includes results where instructions were given in other languages and outputs were produced in English.

\section{Analysis and Discussion}
\paragraph{Which Instruction and Output Language is Best?}
We confirmed that the pattern which instruction and output are English (\{En\}-\{En\}) performed the best ability, whereas the performance is lower for the pattern in which instruction in English and output in other languages (\{En\}-\{Lang\}, i.e., Please generate the output in Chinese).
This suggests that ``LVLMs have a poor ability to successfully transfer knowledge learned in English to other languages''.
We also confirmed that this effect was more pronounced 
in the LoRA-Tuning model (LLaVA-NeXT(FT) and Qwen-VL(FT)).

\paragraph{LVLMs' Ability to Explain Artworks in Other Languages}
Considering the multilingual explanation generation capabilities of LVLMs, a comparison between Table~\ref{tab:result-score-10-en-lang} and Table~\ref{tab:result-score-10-lang-lang} reveals that performing the method is instruction and output in other same languages (\{Lang\}-\{Lang\}) generally yields better results than in the instruction in English and output in other language (\{En\}-\{Lang\}).
When explaining in the native language using data trained in that language, the model effectively manages the knowledge.
However, when explaining in other languages using knowledge trained in English, the model struggles to handle the information adequately. 
This result shows particularly clear in the cases of Qwen-VL's results between the method is instruction and output in Chinese (\{Zh\}-\{Zh\}) and the instruction in English and output in English (\{Zh\}-\{En\}) pairs.
In addition, using English training data for LoRA Tuning likely leads to the forgetting of original performance, resulting in a decline in effectiveness.
From these observations, it is clear that LVLMs currently exhibit their maximum capabilities only when instructed and output in English (\{En\}-\{En\}).
Thus, future research should focus on training LVLMs in multiple languages.

\paragraph{Comparison between En-En and Lang-Lang capabilities}
In this result, En-En performance was the highest. It indicates that existing LVLMs have been primarily pre-trained on English data, and providing both input and output in English allows the model to perform at its best. 
\{Lang\}-\{Lang\} performance was the next best. While it is evident that LVLMs have been pre-trained on languages other than English and can handle multilingual tasks, their performance still falls short of En-En. We believe this is due to insufficient pre-training on non-English languages.

\paragraph{Comparison between En-\{Lang\} and \{Lang\}-\{Lang\}}
We also compared En-\{Lang\} and \{Lang\}-\{Lang\} results, finding that \{Lang\}-\{Lang\} outperformed En-\{Lang\}.
This suggests that providing instructions in English yields better results than in other languages. Interestingly, this contrasts with previous research~\cite{putri-etal-2024-llm}, which found that instructions in English led to better performance. 
While general QA tasks typically assess the model's knowledge, where multilingual knowledge can be effectively utilized, our explanation generation task, which requires the model to provide appropriate explanations using its knowledge, suggests that it struggles to deliver adequate explanations in languages other than English.

\section{Conclusion}
This study focused on artworks, which have a unique image and name regardless of the language, to evaluate the explanation generation abilities of LVLMs in multilingual contexts.
We created datasets compiled from Wikipedia pages in ten languages without using machine translations to evaluate their abilities across multilingual languages.
The results indicate that LVLMs perform optimally when input and output are both in English, while their performance declines when using languages other than English. Thus, our hypothesis, that ``Vision Encoder needs to be learned in other languages as part of its pre-training,'' is correct, and might need to train Vision Transformer using multilingual data.

\section* {Limitations}
\subsection* {Data Collection and Crawling Consistency}
Our initial data collection was conducted through web crawling on June 30th, 2024. It is important to note that subsequent crawls may yield different results due to page updates, such as an increase in the number of pages or the addition of images. As a result, the data retrieved through repeated crawling may not consistently match the original dataset. This introduces a level of variability in the data, which must be considered when replicating or extending this research.

\subsection* {Necessity of Human Evaluation Across Multiple Languages}
To validate the effectiveness and accuracy of LLMs, especially when dealing with complex and diverse linguistic features across multiple languages, human evaluation is indispensable. In this study, we conducted manual evaluations across ten languages. This step is crucial for assessing the model's real-world applicability and ensuring that automated evaluations do not overlook nuanced errors that only human evaluators can identify.

\subsection*{Coverage of Entity Knowledge in LVLMs}

Artwork explanation requires an understanding of entities related to given artworks. However, similar to small-scale models \citep{kamigaito-etal-2023-table}, LVLMs sometimes lack proper alignment of entity knowledge between LLMs and vision encoders. One possible way to address this issue is by leveraging external knowledge. For example, Vision RAG \citep{faysse2024colpaliefficientdocumentretrieval} can search text from given images whereas it requires a retriever capable of interpreting artistic knowledge and LVLMs that can handle lengthy exemplars. Meanwhile, Multimodal Knowledge Graphs (KGs) \citep{chen2024knowledgegraphsmeetmultimodal} are concise yet sparse. Although PLM-based KG completion (KGC) \citep{yao2019kgbertbertknowledgegraph} can potentially alleviate sparsity, its performance is uncertain due to the risk of data leakage \citep{sakai-etal-2024-pre}. By contrast, traditional embedding-based KGC \citep{10.5555/3104482.3104584} is robust and reliable, grounded in theoretical studies \citep{kamigaito-hayashi-2021-unified,pmlr-v162-kamigaito22a,kamigaito22erratum,feng-etal-2023-model,feng-etal-2024-unified}, but its performance is generally lower than that of PLM-based KGC. Consequently, there is no decisive approach for integrating external knowledge into LVLMs for artwork explanation. The similar challenge arises in image review evaluation as reported by \citet{hayashi-etal-2025-irr}.

\subsection*{Length of Generated Explanation}

Our experimental results demonstrate that the length of the generated explanations can vary drastically across different LVLMs, potentially hindering fair comparisons. An instruction-based length control method \cite{juseon-do-etal-2024-instructcmp} may mitigate this issue. However, such methods are primarily designed for summarization and their appropriateness for explanation generation is uncertain.

\section* {Ethical Considerations}
\subsection* {Linguistic Considerations and Ethical Implications}
In several languages, nouns are gendered, meaning they are classified as either masculine or feminine such as Spanish and Italian. For this study, we assumed that LLMs are capable of accurately distinguishing between these gendered forms. This assumption is crucial, as it reflects the model's ability to handle linguistic nuances, particularly in gendered languages. This raises ethical considerations, as any failure of the model to accurately represent gendered language could result in biased or incorrect outputs.

\subsection* { Wikipedia Resources among Ten Languages}
Regarding Wikipedia pages, non-English versions are often less well-maintained, and whether entities are as well-organized as in English is debatable. 
In addition, Chinese Wikipedia contains a mix of traditional and simplified characters, which seems less standardized.
In this study, since we crawled pages from Wikipedia and evaluated using their entities, it's possible that the correct answers are included in the outputs of LVLMs.

\section*{Acknowledgement}

This work was supported by JSPS KAKENHI Grant Number JP23H03458.

\bibliography{custom}

\appendix

\section{Appendix}
\label{sec:appendix}
\subsection{Inference and LoRA Tuning Settings}
\label{experiment-settings}
\subsubsection* {Inference setting}
In this study, as far as inference which needs to use GPUs, all experiments were conducted on a single NVIDIA RTX A6000 GPU and NVIDIA A100-SXM4-40GB, with 8-bit quantization utilized for
model generation.
However, there is no InternLM-XComposer-2 with 8-bit, this model was loaded and inferred in 4-bit mode.
To standardize the length of tokens generated across all models, the maximum token length was set to 1024.
The same settings were applied to each model for performance comparison purposes.

\subsubsection* {LoRA Tuning setting}
We conducted LoRA~\cite{hu2022lora} Tuning with two models: LLaVA-NeXT and Qwen-VL. Both were trained using two NVIDIA A100-SXM4-40GB GPUs. Detailed parameters are provided in Table~\ref{tab:qwen-vl-lora} and Table~\ref{tab:llava-next-lora}.

\section{Explanation Generation Abilities from Other Languages to English (\{Lang\}-\{En\})}
When considering output in English from other languages, we found this method also performs less abilities. 
This suggests that LVLMs have relatively less training data in languages other than English, and they may not properly understand instructions given in other languages. 
Thus, it is difficult to say that the integrated learning of LLMs and Vision Encoder work properly.

\begin{table}[h!]
\small
\centering
\begin{tabular}{p{4cm}p{2.5cm}}
\toprule
Hyper Parameter & Value \\
\midrule
torch\_dtype & bfloat16 \\
seed & 42 \\
max length & 2,048 \\
warmup ratio & 0.01 \\
learning rate & 1e-5 \\
batch size & 4 \\
epoch &1 \\
lora r & 64 \\
lora alpha & 16 \\
lora dropout & 0.05 \\
lora target modules & c\_attn, attn.c\_proj, w1, w2 \\
\bottomrule
\end{tabular}
\caption{The hyper-parameters of Qwen-VL used in the experiment, and others, were set to default settings. The implementation used Transformers~\cite{wolf-etal-2020-transformers} and bitsandbytes~\cite{dettmers2022gpt3}.}
\label{tab:qwen-vl-lora}
\end{table}

\begin{table}[h!]
\centering
\small
\begin{tabular}{p{4cm}p{2.5cm}}
\toprule
Hyper Parameter & Value \\
\midrule
seed & 42 \\
max length & 2048 \\
lora enable & True \\
learning rate & 2e-5 \\
warmup ratio & 0.05 \\
lora r & 16 \\
lora alpha & 32 \\
torch\_dtype & float16 \\
\bottomrule
\end{tabular}
\caption{The hyper-parameters of LLaVA-NeXT used in the experiment, and others were also set to default settings.}
\label{tab:llava-next-lora}
\end{table}

\section{Details of experimental settings}
\begin{center}
\resizebox{\columnwidth}{!}{%
    \begin{tabular}{@{}lll@{}}
        \toprule
        Model  & Base Model    & HuggingFace Name  \\
        \midrule
        mPLUG-Owl2 & LLaMA2-7B    & MAGAer13/mplug-owl2-llama2-7b     \\
        Qwen-VL-Chat  & Qwen  & Qwen/Qwen-VL-Chat   \\
        LLaVA-NeXT & LLaMA3-8B  & lmms-lab/llama3-llava-next-8b   \\
        Phi-3 & Phi-3-Vision-128K-Instruct & microsoft/Phi-3-vision-128k-instruct \\
        XComposer2 & internlm-xcomposer2-7B & internlm/internlm-xcomposer2-7B  \\
        \bottomrule
    \end{tabular}
}
\end{center}

\begin{table*}[h!]
\setlength{\tabcolsep}{4pt}
\resizebox{\textwidth}{!}{
\centering
\begin{tabular}{lccccccccccccc}
\toprule
\multirow{2}{*}{\textbf{Language}} & \textbf{A-10} & \textbf{A-5} & \multicolumn{3}{c}{\textbf{Full}} & \multicolumn{3}{c}{\textbf{\#NUM in A-10}} & \multicolumn{3}{c}{\textbf{\#NUM in A-5}} & \textbf{\#NUM in Full} \\
\cmidrule(lr){4-6}\cmidrule(lr){7-9}\cmidrule(lr){10-12}
& \textbf{Test} & \textbf{Test} & \textbf{Train} & \textbf{Valid} & \textbf{Test} & \textbf{En-\{L\}} & \textbf{\{L\}-En} & \textbf{\{L\}-\{L\}} & \textbf{En-\{L\}} & \textbf{\{L\}-En} & \textbf{\{L\}-\{L\}}  & \textbf{En-\{L\}}\\
\midrule
English  & 28  & 306 & 6, 413 & 2, 138 & 2, 138 & 864 & -- & -- & 5, 924 & -- & -- & 29, 064 \\
\midrule
French   & 28  & 306 & --     & --     & 2, 707 & 1, 780 & 1, 180 & 856 & 11, 020 & 6, 184 &  6, 176 & 57, 916 \\
Spanish  & 28  & 306 & --     & --     & 2, 096 & 964 & 1, 176 & 668 & 5, 744 & 6, 056 & 4, 504 & 40, 960 \\
Italian  & 28  & 306 & --    & --     & 1, 977 & 1, 092 & 1, 244 & 596 & 6, 396 & 7, 032 & 4, 900 & 39, 092 \\
Japanese & 28  & 306 & --     & --     & 1, 312 & 1, 012  & 1, 244 & 572 & 6, 156 & 6, 928 & 4, 720 & 23, 760 \\
Russian  & 28  & -- & --     & --     & 1, 589 & 668 & 1, 184 & 604 & -- & -- & -- & 27, 572 \\
German   & 28  & -- & --     & --     & 962 & 732 & 1, 208 & 1, 132 & -- & -- & -- & 23, 792 \\
Dutch    & 28  & -- & --     & --     & 789 & 556 & 1, 284 & 468 & -- & -- & -- & 13, 940 \\
Swedish  & 28  & -- & --     & -- & 774 & 912 & 1, 164 & 444 & -- & -- & -- & 7, 228 \\
Chinese  & 28  & -- & --     & --     & 738 & 1, 092 & 1, 304 & 612 & -- & -- & -- & 8, 624 \\
\midrule
\textbf{ALL} & 280 & 1,530 & - & - & 15, 082 &  9, 672 & 9, 704 & 5, 952 & 35, 240 & 26, 200  & 20, 300 & 271, 948 \\
\bottomrule
\end{tabular}
}
\caption{The number of each language data in Alignment-10, Alignment-5, and Full task, split by train, valid, and test sets. We split train, valid and test sets only English due to the number of data in English.
\#NUM represents the total number of inferences, and ``A'' indicates Alignment, showing the number of aligned artworks across languages. Additionally, 
\{L\} is an abbreviation for \{Lang\}, representing nine languages other than English.
}
\label{tab:full_task_data}
\end{table*}

\section{Details of Creating Datasets or Training Data}
\subsection{How to Choose Ten Languages?}
We selected ten languages with the highest number of articles from the statistics of all language versions of Wikipedia\footnote{\url{https://en.wikipedia.org/wiki/Wikipedia:Multilingual_statistics}}. Of the top 10 prefectures, Cebuano, Egyptian dialects of Arabic, and Polish were deemed difficult to identify by sampling during the evaluation, so we added the runners-up, Chinese and Japanese.

\subsection{How to Split Train, Valid, and Test Data in English?}
For English, a language rich resource, we split the data into train, valid, and test data using six metrics proposed by ~\citet{hayashi2024artwork} (six metrics: page views, number of links, number of edits, number of references, number of language versions, and article length.) were used in this study as well, and the data were divided equally considering famous artworks. 
All data included in the alignment were used as test data so that data used in the alignment task were not included in the train. We described the number of all data in Table~\ref{tab:full_task_data}.

\subsection{License}
In our study, we created a dataset from Wikipedia articles regarding artworks.
Each image is available under the Creative Commons License (CC) or other licenses. Specific license information for each image can be found on the Wikipedia page or the image description page for that image. The images in this study are used under the terms of these licenses, and links to the images are provided in the datasets we publish so that users can download the images directly. The images themselves are not directly published.
Thus, our data does not infringe upon the licenses.

\section{Evaluation Metrics Formulation}
\label{sec:appendix:metric}

This section describes on the evaluation metrics used in Section~\ref{Setup} using mathematical expressions~\cite{hayashi2024artwork, ozaki2024understandingimpactconfidenceretrieval}. An explanation consisting of \( n \) sentences generated by the model is denoted as \( G = \{g_{1},\cdots, g_{n}\} \), and a reference explanation consisting of \( m \) sentences is denoted as \( R = \{r_{1},\cdots, r_{m}\} \). The function \( \text{Entity}(\cdot) \) is defined to extract entities contained in the input text. The notation \( |G| \) represents the total number of tokens in the generated explanation, and \( |R| \) represents the total number of tokens in the reference explanation.

\paragraph{Entity Coverage (EC)} is calculated as follows:
\begin{equation}
EC(G, R) = Cov(G, R)
\end{equation}
Here, \( Cov(G, R) \) is a function returning the proportion of entities in \( R \) that are covered by \( G \). For partial matches, the Lowest Common Subsequence (LCS) is employed to calculate the longest matching length ratio in the generated explanation relative to the length of the reference entity.

\paragraph{Entity F1 (EF$_1$)} is computed as follows:
\begin{align}
    EF_1 &= \frac{2 \times P \times R}{P + R} \\
    P &= \frac{\sum_{e_i \in Entity(G)} \text{Count}_{\text{clip}}(e_i, G, R)}{\sum_{e_j \in Entity(G)} \#(e_j, G)}\\
    R &= \frac{\sum_{e_i \in Entity(R)} \text{Count}_{\text{clip}}(e_i, G, R)}{\sum_{e_j \in Entity(R)} \#(e_j, R)},
\end{align}
where \( \#(e_j, G) \), \( \#(e_j, R) \) are functions that count the occurrences of entity \( e_j \) in \( G \) and \( R \) respectively, and \( \text{Count}_{\text{clip}}(e_i, G, R) \) returns the lesser frequency of occurrence of \( e_i \) in either \( G \) or \( R \).

\paragraph{Entity Cooccurrence (ECooc)} is calculated using \( BP \) from equation (\ref{eq:bp}) as follows:
\begin{align}
&ECooc(G, R) \nonumber\\
=& BP(G, R) \times  Cov(Co(G), Co(R)),
\end{align}
where \( BP(G, R) \) is given by:
\begin{equation}
BP(G, R) = \exp(\max(0.0, \frac{|G|}{|R|} - 1)) \label{eq:bp}
\end{equation}
and the function \( Co(\cdot) \) returns pairs of co-occurring entities within a context window comprising a sentence and its adjacent \( n \) sentences. Sentence segmentation was performed using the nltk sentence splitter for this purpose.\footnote{Sentence segmentation was performed using the NLTK sentence splitter.}

\section{Filtered  Sections}
The following section was filtered in this study. Approximately 30 instances from the Alignment-10 task were reviewed, and sections without informative content.

\begin{tcolorbox}[colback=white,  colframe=red!50!blue, boxrule=1pt, title=English]
\textbf{References, See also, External links, Sources, Further reading, Bibliography, Gallery, Footnotes, Notes References, References Sources, Bibliography (In Spanish), Bibliography (In Italian), Bibliography (In German), Bibliography (In French), Images, Note, Links, Notes, List, Notes and references, List by location}
\end{tcolorbox}

\begin{tcolorbox}[colback=white, colframe=red!80!black, boxrule=1pt, title=Japanese]
\textbf{外部リンク, 参考文献, 関連項目, 脚注, 出典, ギャラリー, バージョン, 注釈, 関連する作品}
\end{tcolorbox}

\begin{tcolorbox}[colback=white, colframe=green!50!black, boxrule=1pt, title=Italian]
\textbf{Collegamenti esterni, Altri progetti, Bibliografia, Note, Omaggi, Voci correlate, Bibliografia, Musica, Fumetti, Letteratura, Filmografia, Nella cultura di massa, Altri progetti, Galleria d'immagini, Curiosità, Calendario}
\end{tcolorbox}

\begin{tcolorbox}[colback=white, colframe=blue, boxrule=1pt, title=French]
\textbf{Liens externes, Articles connexes, Bibliographie et ressources en ligne, Annexes, Notes et références, Divers, Littérature, Peinture et sculpture, Déclinaisons et détournements, Bases de données et dictionnaires, Italien, Français, Ouvrages, Articles, Bibliographie, Théâtre, Cinéma, Article connexe, Annexe, Notes et référence, Voir aussi, Divers, Pour approfondir, Versions, Références, Sources secondaires, Sources originales, Références de l'expression dans l'art, Ouvrages, Ailleurs, Notes, Films, Dans la culture, Postérité, Données techniques, Galerie, Historique}
\end{tcolorbox}

\begin{tcolorbox}[colback=white, colframe=brown!70!black, boxrule=1pt, title=Spanish]
\textbf{Enlaces externos, Bibliografía, Referencias, Fuentes, Enlaces externos, Bibliografía, Véase también, Notas, Información, Galería, Galería de imágenes, Filmografía}
\end{tcolorbox}

\begin{tcolorbox}[colback=white, colframe=red!80!white, boxrule=1pt, title=Chinese (Traditional)]
\textbf{外部連結, 延伸阅读, 参考文献, 參考文獻, 參見, 参見, 書目, 注释与参考文献, 來源, 擴展閱讀, 参考来源, 外部链接, 延伸阅读, 引用, 注释, 參考資料, 参考资料, 相關條目, 參考來源, 參見條目, 其他事項, 参考, 註解, 媒體, 紀錄片, 書籍, 近似作品, 相關作品, 德文, 注释, 擴展閱讀, 吉米·威尔士的声明}
\end{tcolorbox}

\begin{tcolorbox}[colback=white, colframe=red!80!white, boxrule=1pt, title=Chinese (Simplified)]
\textbf{\Chinese{外部链接, 延伸阅读, 参考文献, 参见, 注释与参考文献, 来源, 扩展阅读, 参考来源, 引用, 注释, 参考资料, 相关条目, 参见条目, 其他事项, 参考, 近似作品, 媒体, 纪录片, 书籍, 注释, 吉米·威尔士的声明}}
\end{tcolorbox}

\begin{tcolorbox}[colback=white, colframe=yellow!70!black, boxrule=1pt, title=Swedish]
\textbf{Noter, Referenser, Se även, Externa länkar, Allmänna källor, Galleri, Källor, Bilder, Kalenderfunktionen, Relaterade målningar}
\end{tcolorbox}

\begin{tcolorbox}[colback=white, colframe=blue!50!white, boxrule=1pt, title=Dutch]
\textbf{Zie ook, Literatuur, Externe links, Bewerkingen, Andere, Latere edities, Trivia, Zie ook, Galerij, Originele gietingen, Stanza dell’incendio del Borgo, Stanza della Segnatura, Noten, Literatuur en bronnen}
\end{tcolorbox}

\begin{tcolorbox}[colback=white, colframe=cyan!50!black, boxrule=1pt, title=Russian]
{\fontencoding{T2A}\selectfont \textbf{Ссылки, Примечания, См. также, Документалистика, Литература, Источники, Отражение в искусстве}}
\end{tcolorbox}

\begin{tcolorbox}[colback=white, colframe=red!30!black, boxrule=1pt, title=German]
\textbf{Anmerkungen, Weblinks, Literatur, Anmerkungen und Einzelnachweise, Einzelbelege, Einzelnachweise, Chronologie, Quellen, Übersicht, Literatur (Auswahl), Siehe auch, Rezeption, Dokumentarfilme, Ausstellungen, Siehe auch}
\end{tcolorbox}

\onecolumn
\section {Instruction to Native Speakers}
We asked native speaker to prepare the instruction to check if the above template is equal in difficulty compared to the English text.
\label{appendix:instruction}
\begin{tcolorbox}[title=Instruction, boxrule=1pt]
\# What we research \\
We are conducting a study to measure LLMs' ability to understand the arts.
Previous studies have been done only for English, and we are now trying to extend and validate it for multiple languages. \\

The text presented has been translated from English into your language using DeepL. \\

I want you to make sure that the sentence you translate has the same meaning as the English sentence. \\
The time I assume will not take more than 5 minutes and that's about OK for a check. I also use back translation to check it, so I believe it is not that broken. \\

\# Keep in mind \\
\begin{itemize}
    \item My final goal is to have the sentences corrected to be as natural as English sentences.
    \item Depending on \{title\} and \{section\}, and in some countries, you may need to be concerned about masculine and feminine nouns. If that is the case, choose whichever you type into the LLM in your native language (i.e., the more natural one). 
    \item Please do not change the entire text. 
    \item Changing, deleting or adding words is acceptable.
\end{itemize}

\# Examples of \{title\} and \{section\}, \{subsection\} and \{subsubsection\} \\
We use Wikipedia for our research. \\
Here is one of the example: \url{https://en.wikipedia.org/wiki/Mona_Lisa} \\
In this case, \{title\} will contain ``Mona Lisa''. \\
In addition, \{section\} contains ``Description'', ``History'', and so on. \\
\{subsection\} refers to a smaller frame within \{section\}, such as ``Creation and date''. \\

↓ Below is the text I would like you to review. \\
(\{lang\}\_temp1\_sec is translated from en\_temp1\_sec using DeepL)
(\{lang\}\_temp2\_subsec is translated from en\_temp2\_subsec using DeepL) \\

\#English (source) \\
This sentence is a sample. \\

\# Your native language (target I translated from DeepL.) \\
This sentence is a sample.
\end{tcolorbox}

\clearpage
\section {Other Results and Visualizations}
\input{results/full}

\input{results/a5}

\input{results/a10-appendix}

\input{results/a10-lang-en}

\begin{figure*}[h!]
    \centering
    \includegraphics[width=0.9\textwidth]{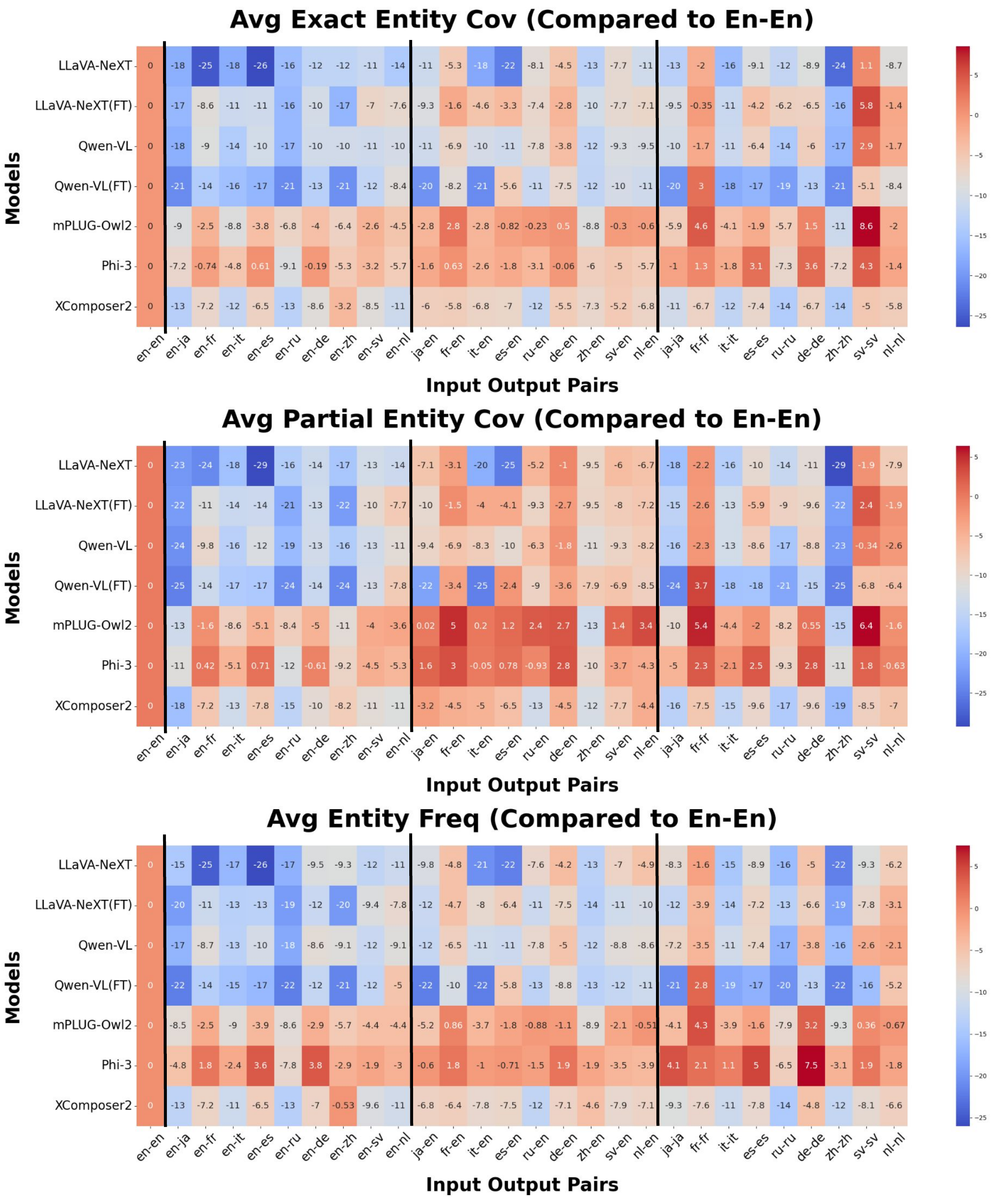}
    \caption{Visualization of Alignment-10 results in a heat map. We made the visualization based on when we had LVLMs give instructions and output in English.}
    \label{fig:a10-heatmap1}
\end{figure*}

\begin{figure*}[h!]
    \centering
    \includegraphics[width=0.9\textwidth]{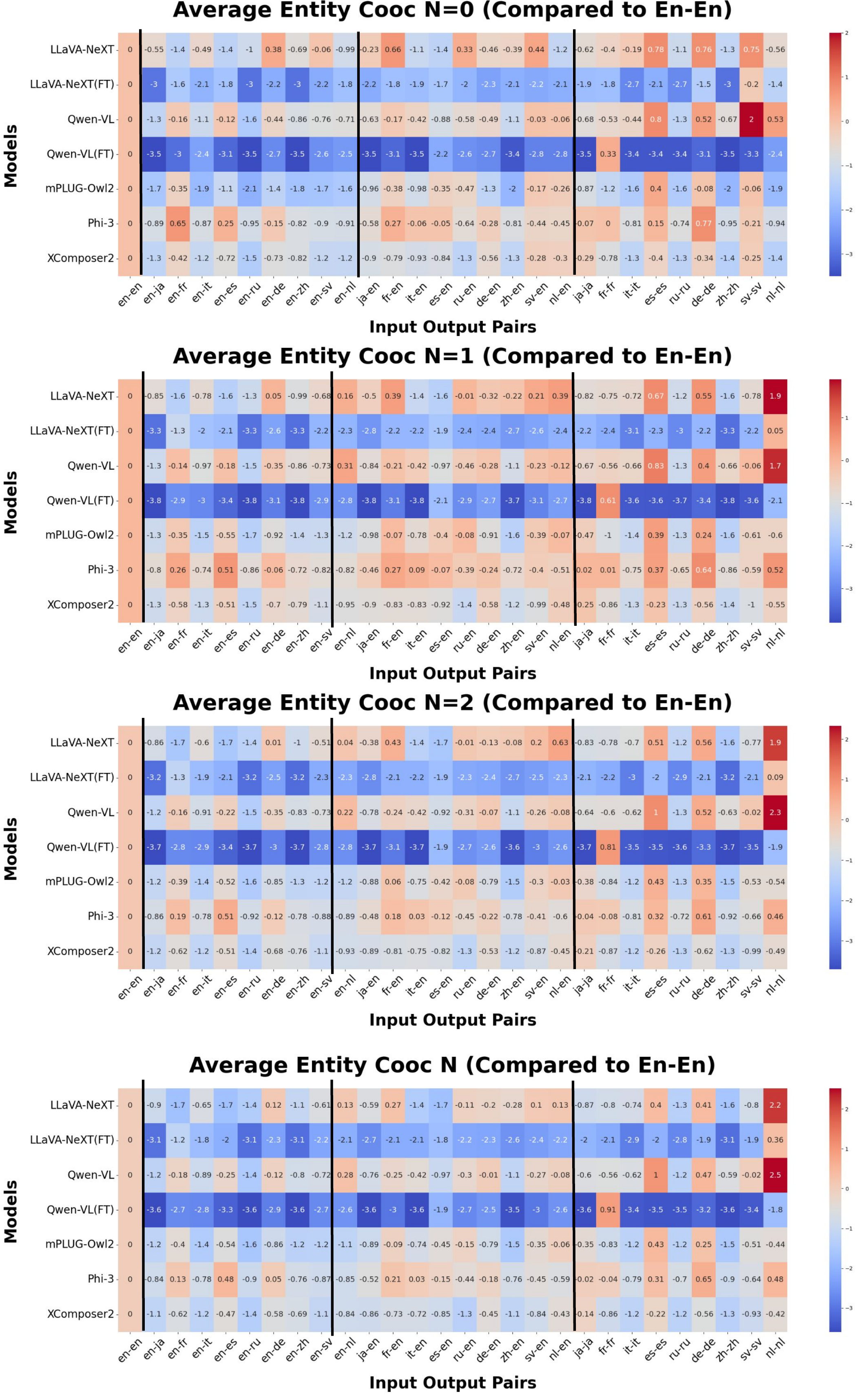}
    \caption{Visualization of Alignment-10 results in a heat map. We made the visualization based on when we had LVLMs give instructions and output in English.}
    \label{fig:a10-heatmap2}
\end{figure*}

\begin{figure*}[h!]
    \centering
    \includegraphics[width=\textwidth]{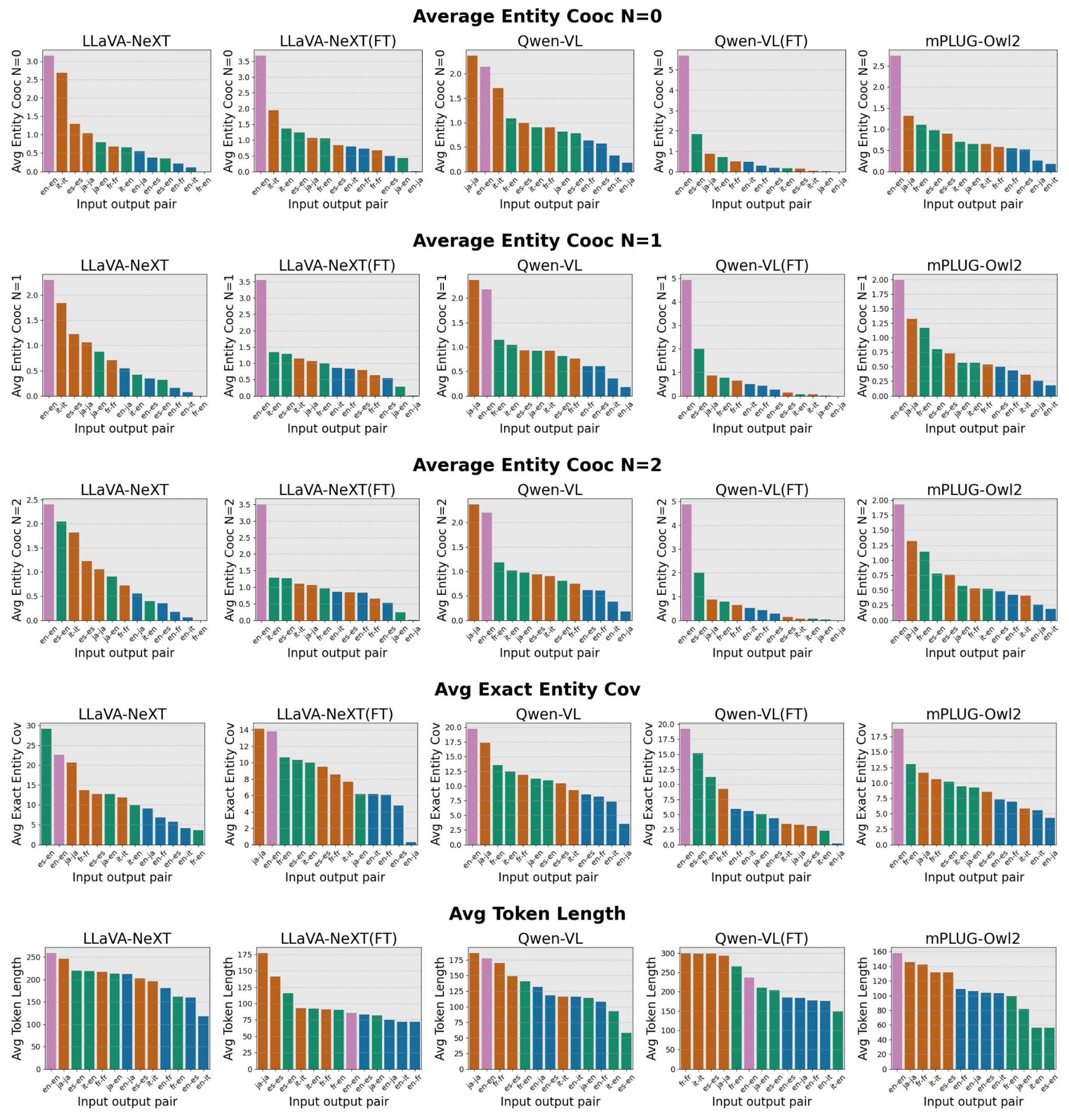}
    \caption{The rest of the results in the Alignment-5 task. From this figure, it can also be seen that the English instructions are optimal, even if the number of data is expanded.
    Purple bin indicates the method which is the instruction and the output in English (\{En\}-\{En\}), Green bin indicates the instruction in languages other than English and the output in English (\{Lang\}-\{En\}), Brown bin indicates the instruction and output in languages other than English (\{Lang\}-\{Lang\}) and Blue bin indicates the instruction in English and the output in languages other than English (\{En\}-\{Lang\}).}
    \label{fig:a5-figure-appendix}
\end{figure*}

\input{results/full-misc}

\clearpage
\section {Comparison with OpenAI API}
We also conducted Alignment-5 and Alignment-10 task by \texttt{gpt-4o-mini} (gpt-4o-mini-2024-07-18).
\vspace{50pt}

\input{results/additional-openai-a10}

\input{results/additional-openai-a5}

\section{Details of Each Language Templates}
As indicated in Table~\ref{tab:prompt_template}, we created Templates for ten languages. Ten language templates are shown below. These templates were modified by nine native speakers of the country's first language, who were asked to modify the sentences to have the same nuance and level of detail as in English. We described these templates from Table~\ref{tab:template-en} to Table~\ref{tab:template-ru}.

\input{tables/appendix_template}

\clearpage
\section{Dataset Example}
The below block is an example of datasets we created.
The full dataset is available at \url{https://huggingface.co/datasets/naist-nlp/MultiExpArt}.
\\ 
\begin{tcolorbox}[title=Dataset Example, boxrule=1pt, colback=white, coltitle=white, colframe=blue!50, colbacktitle=blue!60]
\texttt{\{}\\
\texttt{"text\_id": 1,}\\
\texttt{"title": "Mona Lisa",}\\
\texttt{"en\_title": "Mona Lisa",}\\
\texttt{"lang": "en",}\\
\texttt{"source\_lang": "en",}\\
\texttt{"target\_lang": "en",}\\
\texttt{"image\_url": "IMAGE\_URL\_FOR\_MONA\_LISA",}\\
\texttt{"en\_image\_url": "IMAGE\_URL\_FOR\_MONA\_LISA",}\\
\texttt{"template": "en\_template1\_section",}\\
\texttt{"prompt": "Focus on Mona Lisa and explain the Abstract.",}\\
\texttt{"gold": "The Mona Lisa is a half-length portrait painting by Italian artist Leonardo da Vinci. Considered an archetypal masterpiece of the Italian Renaissance, it has been described as \"the best known, the most visited, the most written about, the most sung about, [and] the most parodied work of art in the world\". The painting's novel qualities include the subject's enigmatic expression, monumentality of the composition, the subtle modelling of forms, and the atmospheric illusionism. The painting has been traditionally considered to depict the Italian noblewoman Lisa del Giocondo. It is painted in oil on a white poplar panel. Leonardo never gave the painting to the Giocondo family.",}\\
\texttt{"gold\_entity": ["Francis I of France", "Italian Renaissance", "Italy", "Francis I of France", "Leonardo da Vinci", "Lisa del Giocondo", "Louvre", "Louvre", "Nat King Cole"]}\\
\texttt{\}}
\end{tcolorbox}

\newpage
\section{Examples of LVLM outputs}
Tables~\ref{tab:llava-output-examples} and~\ref{tab:llava-output-examples2} present output examples from LLaVA-NeXT.

\begin{table}[h!]
    \centering
    \resizebox{\textwidth}{!}{
    \begin{tabular}{p{3cm}ccp{10cm}}
    \toprule
    \textbf{Question} & \textbf{Input} & \textbf{Output} & \textbf{LVLM output} \\ 
    \midrule
    \multirow{40}{*}{\rotatebox{90}{\shortstack{Focus on Venus de Milo and explain the Description.}}} & En & En &  The image you've provided appears to be of the Venus de Milo, a famous marble sculpture of the Greek goddess Aphrodite (Venus in Roman mythology). The sculpture is known for its missing arms, which has led to various interpretations and speculations about its original form and the story it was meant to tell. The Venus de Milo is characterized by its draped clothing, which is typical of classical Greek sculpture, and the flowing drapery that suggests movement. The figure is depicted in a standing pose, which is common for depictions of the goddess, and the facial expression is serene and contemplative.The sculpture is missing its arms, which has led to much speculation about the original intent of the artist. Some believe that the missing arms were removed to prevent the statue from being used for religious purposes, while others think that the arms were removed for artistic reasons, to emphasize the beauty of the torso and the elegance of the pose.The Venus de Milo is one of the most celebrated sculptures in the world and is housed in the Louvre Museum in Paris, France. It is a symbol of beauty and has been the subject of numerous reproductions and adaptations throughout history. \\
    \cmidrule{2-4}
    & En & Ja & Venus de Miloは、ギリシャの古代彫刻で有名な女性像の一つで、世界中で愛される名彫刻の一つです。Venus de Miloは、ギリシャの島であるミロス島で発見されましたが、その後、様々な謎に包まれています。2015年4月25日から9月6日まで、三菱地所美術館で「Venus de Miloとその謎」という展覧会が開催されます。この展覧会では、Venus de Miloの謎を解くための最新の科学技術が活用され、Venus de Miloの美しさと魅力を再び発見することができます。Venus de Miloは、ギリシャの島であるミロス島で発見されました。 \\
    \cmidrule{3-4}
    & En & Ja (DeepL) & Venus de Milo is one of the most famous female figures in ancient Greek sculpture, and one of the most beloved masterpieces of sculpture in the world.Venus de Milo was discovered on the Greek island of Milos, but has since been shrouded in mystery.From April 25 to September 6, 2015, the Mitsubishi Jisho Art Museum will host the exhibition “Venus de Milo and its Mysteries”. The exhibition will utilize the latest scientific techniques to solve the mystery of Venus de Milo, allowing visitors to rediscover the beauty and charm of Venus de Milo, which was discovered on the Greek island of Milos.  \\
    \bottomrule    
    \end{tabular}
    }
    \caption{Comparison of output examples from LLaVA-NeXT. (DeepL) represents translations from other languages into English.}
    \label{tab:llava-output-examples}
\end{table}

\begin{table}[h!]
    \centering
    \resizebox{\textwidth}{!}{
    \begin{tabular}{p{3cm}ccp{10cm}}
    \toprule
    \textbf{Question} & \textbf{Input} & \textbf{Output} & \textbf{LVLM output} \\ 
    \midrule
    \multirow{40}{*}{\rotatebox{90}{\shortstack{Focus on Venus de Milo and explain the Description.}}}  & Ja & Ja & この彫刻は、ローマのサン・ピエトロ大聖堂の前庭に立つ。ミロのヴィーナスは、ローマのサン・ピエトロ大聖堂の前庭に立つ。 \\
    \cmidrule{3-4}
    & Ja & Ja (DeepL) & This sculpture stands in the vestibule of St. Peter's Basilica in Rome. The Venus de Milo stands in the vestibule of St. Peter's Basilica in Rome. \\
    \cmidrule{2-4}
    & Ja & En & The statue is a nude, with a contrapposto pose. The statue is 1.5 meters tall, and is 0.8 meters wide. The statue is on a marble base that is 0.8 meters tall, and 0.9 meters wide. The statue is on display in the Louvre Museum. \\
    \cmidrule{2-4}
    & -- & Gold & The Venus de Milo is an over tall. Brill's New Pauly says . The Louvre's online catalogue states . Christofilis Maggidis says . Alain Pasquier says excluding the plinth, or including it. Parian marble statue of a Greek goddess, most likely Aphrodite, depicted with a bare torso and drapery over the lower half of her body. The figure stands with her weight on her right leg, and the left leg raised; her head is turned to the left. The statue is missing both arms, the left foot, and the earlobes. There is a filled hole below her right breast that originally contained a metal tenon that would have supported the right arm. The Venus' flesh is polished smooth, but chisel marks are still visible on other surfaces. The drapery is more elaborately carved on the right-hand side of the statue than the left, perhaps because on the left-hand side it was originally obscured from view. Likewise the Venus is less finely-finished from behind, suggesting that it was originally intended to be viewed only from the front. While the body of the Venus is depicted in a realistic style, the head is more idealised. The lips are slightly open, the eyes and mouth are small. The sculpture has been minimally restored: only the tip of the nose, lower lip, big toe on the right foot, and some of the drapery.Stylistically, the sculpture combines elements of classical and Hellenistic art. Features such as the small, regular eyes and mouth, and the strong brow and nose, are classical in style, while the shape of the torso and the deeply carved drapery are Hellenistic.Kenneth Clark describes the figure as "the last great work of antique Greece", and "of all the works of antiquity one of the most complex and the most artful. ...[the sculptor] has consciously attempted to give the effect of a 5th-century work", while also using "the inventions of his own time"; "the planes of her body are so large and calm that at first we do not realise the number of angles through which they pass. In architectural terms, she is a baroque composition with classic effect".  \\
    \bottomrule
    \end{tabular}
    }
    \caption{Comparison of output examples from LLaVA-NeXT. (DeepL) represents translations from other languages into English.}
    \label{tab:llava-output-examples2}
\end{table}

\clearpage

\section{Pre-trained data list}
This study specifies the pretraining data used for the open models employed. 
As of September 2024, Phi-3 and LLaVA-NeXT have not been released.

\begin{table}[h!]
    \centering
    \resizebox{\textwidth}{!}{
    \begin{tabular}{ccccc}
    \toprule
    \multirow{3}{*}{\textbf{Data Type}} & \multirow{3}{*}{\textbf{Data Name}} & \multicolumn{3}{c}{\textbf{Model}} \\
    \cmidrule(lr){3-5}
    & & \shortstack{\textbf{mPLUG-Owl2} \\ \citep{ye2024mplug}} & \shortstack{\textbf{Qwen-VL} \\ \citep{bai2023qwen}} & \shortstack{\textbf{XComposer2} \\ \citep{dong2024internlm}} \\
    \midrule
        \multirow{3}{*}{\textbf{Text}} & ShareGPT~\citep{chen2025sharegpt4v} & $\checkmark$ & $\checkmark$ & $\checkmark$ \\
        & SlimOrca~\citep{SlimOrca} & $\checkmark$ & & \\
        & In-house Data & $\checkmark$ & $\checkmark$ & $\checkmark$ \\
        \cmidrule(lr){2-5}
        \multirow{7}{*}{\textbf{Dialogue Caption}} & LLAVA~\citep{liu2023improvedllava} & $\checkmark$ & & \\
        & COCO~\citep{chen2015microsoft} & $\checkmark$ & $\checkmark$ & $\checkmark$ \\
        & TextCaps~\citep{sidorov2019textcaps} & $\checkmark$ & & $\checkmark$ \\
        & SBU~\citep{NIPS2011_5dd9db5e} & & $\checkmark$ & \\
        & DataComp~\citep{datacomp} & $\checkmark$ & & \\
        & CC12M \& 3M~\citep{changpinyo2021cc12m} & $\checkmark$ & & $\checkmark$ \\
        & LAION-en \& zh~\citep{schuhmann2022laion} & $\checkmark$ & & $\checkmark$ \\
        \cmidrule(lr){2-5}
        \multirow{7}{*}{\textbf{VQA}} & VQA2~\citep{goyal2017making}& $\checkmark$ & $\checkmark$ & \\
        & GQA~\citep{hudson2019gqa} & $\checkmark$ & $\checkmark$ & \\
        & OKVQA~\citep{marino2019ok} & $\checkmark$ & $\checkmark$ & \\
        & OCRVQA~\citep{mishra2019ocr} & $\checkmark$ & & \\
        & A-OKVQA~\citep{schwenk2022okvqa} & $\checkmark$ & & \\
        & DVQA~\citep{kafle2018dvqa} & $\checkmark$ & & \\
        & TextVQA~\citep{VQA} & $\checkmark$ & & \\
        \cmidrule(lr){2-5}
        \multirow{7}{*}{\textbf{Grounding}} & A12D & $\checkmark$ & & \\
        & Ref Grounding & & $\checkmark$ & \\
        & GRIT~\citep{peng2023kosmos} & $\checkmark$ & & \\
        & VisualGenome~\citep{krishna2016visual} & $\checkmark$ & $\checkmark$ & \\
        & RefCOCO~\citep{decoding} & & $\checkmark$ & \\
        & RefCOCO+~\citep{decoding} & & $\checkmark$ & \\
        & RefCOCOg & & $\checkmark$ & \\
        \cmidrule(lr){2-5}
        \multirow{2}{*}{\textbf{OCR}} & SynthDoG-en \& zh~\citep{kim2022ocr} & $\checkmark$ & & \\
        & Common Crawl pdf \& HTML & $\checkmark$ & & \\
        \cmidrule(lr){2-5}
        \multirow{2}{*}{\textbf{Image Captioning}} & Web CapFilt~\citep{pmlr-v162-li22n} & $\checkmark$ & $\checkmark$ & \\
        & NoCaps & $\checkmark$ & & $\checkmark$ \\
        \cmidrule(lr){2-5}
        \multirow{2}{*}{\textbf{Visual Spatial Reasoning}}  & Flickr30K~\citep{hambardzumyan2022deeplakelakehousedeep} & & $\checkmark$ & $\checkmark$ \\
        & IconQA~\citep{lu2021iconqa} & & & \\
        \cmidrule(lr){2-5}
        \multirow{3}{*}{\textbf{Video Question Answering}} & MSVD-QA & & $\checkmark$ & \\
        & MSRVT-QA & $\checkmark$ & & \\
        & iVQA~\citep{liu2018ivqa}& & $\checkmark$ & \\
        \cmidrule(lr){2-5}
        \textbf{Image Classification} & VizWiz~\citep{gurari2018vizwiz}& & $\checkmark$ & \\
        \cmidrule(lr){2-5}
        \textbf{Knowledge-Grounded Image QA} & ScienceQA~\citep{lu2022learn} & & $\checkmark$ & \\
        \bottomrule
        \end{tabular}
    }
    \caption{Pretraining dataset list.}
    \label{tab:pretraindata-list}
\end{table}

\end{document}

%% file: tables/table2.tex
\begin{table*}[t]
    \centering
    \resizebox{\textwidth}{!}{ 
    \begin{tabular}{@{}llp{6cm}p{6cm}p{6cm}@{}}
        \toprule
        \textbf{Language} & \textbf{Type} & \textbf{Template} & \textbf{Instruction} & \textbf{Output}\\
        \midrule
        \multirow{6.5}{*}{English} & \multirow{3}{*}{Section} & 
        \texttt{Explain the \textbf{\textcolor[HTML]{AD0000}{\{Section\}}} of this artwork, \textbf{\textcolor[HTML]{1400AD}{\{Title\}}}.} & 
        \texttt{Explain the \textbf{\textcolor[HTML]{AD0000}{History}} of this artwork, \textbf{\textcolor[HTML]{1400AD}{Mona Lisa}}.} & Of Leonardo da Vinci's works, the Mona Lisa is the only portrait whose authenticity...\\
        \cmidrule{2-5}
         & \multirow{3}{*}{Subsection} & 
        \texttt{Explain the \textbf{\textcolor[HTML]{AD0000}{\{Subsection\}}} regarding the \textbf{\textcolor[HTML]{AD0000}{\{Section\}}} of this artwork, \textbf{\textcolor[HTML]{1400AD}{\{Title\}}}.} & 
        \texttt{Explain the \textbf{\textcolor[HTML]{AD0000}{Creation and date}} regarding the \textbf{\textcolor[HTML]{AD0000}{History}} of this artwork, \textbf{\textcolor[HTML]{1400AD}{Mona Lisa}}.} & The record of an October 1517 visit by Louis d'Aragon states that the Mona Lisa...\\
        \midrule
        \multirow{7.5}{*}{Japanese} & \multirow{3}{*}{Section} & 
        \texttt{\textbf{\textcolor[HTML]{1400AD}{\{Title\}}}の作品に関して、この作品の\textbf{\textcolor[HTML]{AD0000}{\{Section\}}}を説明してください。} & 
        \texttt{\textbf{\textcolor[HTML]{1400AD}{モナリザ}}の作品に関して、\textbf{この作品の\textcolor[HTML]{AD0000}{歴史}}について説明してください。} & レオナルド・ダ・ヴィンチの作品の中で、「モナ・リザ」は唯一、その真偽が不確かな肖像画であり...\\
        \cmidrule{2-5}
         & \multirow{4}{*}{Subsection} & 
        \texttt{\textbf{\textcolor[HTML]{1400AD}{\{Title\}}}の作品に関して、この作品の\textbf{\textcolor[HTML]{AD0000}{\{Section\}}}に関する\textbf{\textcolor[HTML]{AD0000}{\{Subsection\}}}}を説明してください。 & 
        \texttt{\textbf{\textcolor[HTML]{1400AD}{モナリザ}}の作品に関して、この作品の\textbf{\textcolor[HTML]{AD0000}{歴史に関する\textcolor[HTML]{AD0000}{制作と日付}}}を説明してください。} & 1517年10月のルイ・ド・アラゴンの訪問の記録には、「モナ・リザ」について...\\
        \midrule
        \multirow{5.5}{*}{Chinese} & \multirow{2}{*}{Section} & 
        \texttt{\Chinese{解释这件艺术品的}\textcolor[HTML]{AD0000}{\{Section\}}，\textbf{\textcolor[HTML]{1400AD}{\{Title\}}}。} &
        \texttt{\Chinese{解释这件艺术品的}\textcolor[HTML]{AD0000}{\Chinese{历史}}，\textbf{\textcolor[HTML]{1400AD}{\Chinese{蒙娜丽莎}}}。} & \Chinese{在达芬奇的作品中，蒙娜丽莎是唯一一幅真伪有争议的肖像画...}\\
        \cmidrule{2-5}
         & \multirow{3}{*}{Subsection} & 
        \texttt{\Chinese{解释关于这件艺术品的}\textcolor[HTML]{AD0000}{\{Section\}}\Chinese{的}\textcolor[HTML]{AD0000}{\{Subsection\}}，\textbf{\textcolor[HTML]{1400AD}{\{Title\}}}。} &
        \texttt{\Chinese{解释关于这件艺术品的}\textbf{\textcolor[HTML]{AD0000}{\Chinese{历史的}}\textbf{\textcolor[HTML]{AD0000}{\Chinese{创作和日期}}}}，\textbf{\textcolor[HTML]{1400AD}{\Chinese{蒙娜丽莎}}}。} & \Chinese{路易·德·阿拉贡在1517年10月访问的记录中提到，“蒙娜丽莎”...}\\
        \midrule
        \multirow{6.5}{*}{Spanish} & \multirow{3}{*}{Section} & 
        \texttt{Explica la \textbf{\textcolor[HTML]{AD0000}{\{Seccion\}}} de esta obra de arte, \textbf{\textcolor[HTML]{1400AD}{\{Title\}}}.} &
        \texttt{Explica la \textbf{\textcolor[HTML]{AD0000}{Historia}} de esta obra de arte, \textbf{\textcolor[HTML]{1400AD}{Mona Lisa}}.} & De las obras de Leonardo da Vinci, la Mona Lisa es el único retrato cuya autenticidad...\\
        \cmidrule{2-5}
         & \multirow{3}{*}{Subsection} & 
        \texttt{Explica la \textbf{\textcolor[HTML]{AD0000}{\{Subsection\}}} sobre la \textbf{\textcolor[HTML]{AD0000}{\{Section\}}} de esta obra de arte, \textbf{\textcolor[HTML]{1400AD}{\{Title\}}}.} &
        \texttt{Explica la \textbf{\textcolor[HTML]{AD0000}{Creación y fecha}} sobre la \textcolor[HTML]{AD0000}{Historia} de esta obra de arte, \textbf{\textcolor[HTML]{1400AD}{Mona Lisa}}.} & El registro de una visita en octubre de 1517 de Luis de Aragón menciona que la Mona Lisa...\\
        \bottomrule
    \end{tabular}
    }
    \caption{Examples of templates and instructions for the proposed task. The blue part indicates the artwork's title and the red part indicates the names of sections and subsections in the original Wikipedia articles that correspond to their explanations. We prepared such templates for ten languages and asked native speakers to make sure they are on the same level as English.}
    \label{tab:prompt_template}
\end{table*}

%% file: results/a10-en-en.tex
\begin{table*}[t]
\centering
\resizebox{\textwidth}{!}{
\footnotesize
\begin{tabular}{@{}cccccccccccc@{}}
\toprule 
\multicolumn{1}{c}{\multirow{2.5}{*}{Input}} & \multirow{2.5}{*}{Output} & \multirow{2.5}{*}{LVLM} & \multicolumn{2}{c}{Entity Cov.} & \multirow{2.5}{*}{Entity F1} & \multicolumn{4}{c}{Entity Cooccurrence} & \multirow{2}{*}{Avg. Length} \\
\cmidrule(lr){4-5} \cmidrule(lr){7-10}
& & & exact & partial & & n=0 & n=1 & n=2 & n=$\infty$ & \\
\midrule
\multirow{5}{*}{En} & \multirow{5}{*}{En} & LLaVA-NeXT &  \textbf{26.49} & \textbf{31.54} & \textbf{26.07} & 1.35 & 1.65 & 1.66 & 1.70 & 252 \\ 
& & LLaVA-NeXT (FT) &  16.98 & 22.70 & 19.95 & 3.02 & 3.33 & 3.23 & 3.11 & 83 \\ 
& & Qwen-VL &  21.11 & 27.03 & 21.78 & 1.60 & 1.59 & 1.56 & 1.52 & 155 \\ 
& & Qwen-VL (FT) &  21.12 & 24.87 & 21.95 & \textbf{3.57} & \textbf{3.83} & \textbf{3.78} & \textbf{3.68} & 177 \\ 
& & mPLUG-Owl2 &  12.79 & 17.08 & 13.48 & 2.07 & 1.68 & 1.59 & 1.56 & 151 \\ 
 
\bottomrule
\end{tabular}
}
\caption{Results of LVLMs in Alignment-10 Task (the instruction and the output in English, \{En\}-\{En\}). Bold fonts indicate the best scores. The red and blue figures shown in the following figures are the different figures compared to this Table. "(FT)" indicates the model conducted LoRA-Tuning.}
\label{tab:result-score-10-en-en}
\end{table*}

%% file: results/a10-en-lang.tex
\begin{table*}[t]
\centering
\resizebox{\textwidth}{!}{
\footnotesize
\setlength{\tabcolsep}{5pt}
\begin{tabular}{@{}cccrrrrrrrc@{}}
\toprule 
\multicolumn{1}{c}{\multirow{2.5}{*}{Input}} & \multirow{2.5}{*}{Output} & \multirow{2.5}{*}{LVLM} & \multicolumn{2}{c}{Entity Cov.} & \multicolumn{1}{c}{\multirow{2.5}{*}{Entity F1}} & \multicolumn{4}{c}{Entity Cooccurrence} & \multirow{2.5}{*}{Avg. Length} \\
\cmidrule(lr){4-5} \cmidrule(lr){7-10}
& & & \multicolumn{1}{c}{exact} & \multicolumn{1}{c}{partial} & & \multicolumn{1}{c}{n=0} & \multicolumn{1}{c}{n=1} & \multicolumn{1}{c}{n=2} & \multicolumn{1}{c}{n=$\infty$} & \\
\midrule
\multirow{5}{*}{En} & \multirow{5}{*}{Es}  & LLaVA-NeXT & 0.00 (\textcolor[HTML]{1400AD}{-26.49}) & 2.24 (\textcolor[HTML]{1400AD}{-29.30}) & 0.00 (\textcolor[HTML]{1400AD}{-26.07}) & 0.00 ($\:\:$\textcolor[HTML]{1400AD}{-1.35}) & 0.00 ($\:\:$\textcolor[HTML]{1400AD}{-1.65}) & 0.00 ($\:\:$\textcolor[HTML]{1400AD}{-1.66}) & 0.00 ($\:\:$\textcolor[HTML]{1400AD}{-1.70}) & 137 \\ 
& & LLaVA-NeXT (FT) & 6.23 (\textcolor[HTML]{1400AD}{-10.75}) & 9.05 (\textcolor[HTML]{1400AD}{-13.65}) & 6.87 (\textcolor[HTML]{1400AD}{-13.08}) & 1.27 ($\:\:$\textcolor[HTML]{1400AD}{-1.75}) & 1.21 ($\:\:$\textcolor[HTML]{1400AD}{-2.12}) & 1.09 ($\:\:$\textcolor[HTML]{1400AD}{-2.13}) & 1.06 ($\:\:$\textcolor[HTML]{1400AD}{-2.05}) & 83 \\ 
& & Qwen-VL & \textbf{10.81 (\textcolor[HTML]{1400AD}{-10.29)}} & \textbf{15.18 (\textcolor[HTML]{1400AD}{-11.85)}} & \textbf{11.42 (\textcolor[HTML]{1400AD}{-10.36)}} & \textbf{1.48 ($\:\:$\textcolor[HTML]{1400AD}{-0.12)}} & \textbf{1.41 ($\:\:$\textcolor[HTML]{1400AD}{-0.18)}} & \textbf{1.34 ($\:\:$\textcolor[HTML]{1400AD}{-0.22)}} & \textbf{1.27 ($\:\:$\textcolor[HTML]{1400AD}{-0.25)}} & 109 \\ 
& & Qwen-VL (FT) & 4.25 (\textcolor[HTML]{1400AD}{-16.87}) & 7.86 (\textcolor[HTML]{1400AD}{-17.01}) & 5.40 (\textcolor[HTML]{1400AD}{-16.55}) & 0.36 ($\:\:$\textcolor[HTML]{1400AD}{-3.21}) & 0.36 ($\:\:$\textcolor[HTML]{1400AD}{-3.47}) & 0.31 ($\:\:$\textcolor[HTML]{1400AD}{-3.46}) & 0.29 ($\:\:$\textcolor[HTML]{1400AD}{-3.39}) & 190 \\ 
& & mPLUG-Owl2 & 8.95 ($\:\:$\textcolor[HTML]{1400AD}{-3.83}) & 11.95 ($\:\:$\textcolor[HTML]{1400AD}{-5.13}) & 9.62 ($\:\:$\textcolor[HTML]{1400AD}{-3.86}) & 0.93 ($\:\:$\textcolor[HTML]{1400AD}{-1.14}) & 1.13 ($\:\:$\textcolor[HTML]{1400AD}{-0.55}) & 1.07 ($\:\:$\textcolor[HTML]{1400AD}{-0.52}) & 1.02 ($\:\:$\textcolor[HTML]{1400AD}{-0.54}) & 108 \\ 
\midrule
\multirow{5}{*}{En} & \multirow{5}{*}{Fr}  & LLaVA-NeXT & 1.00 (\textcolor[HTML]{1400AD}{-25.49}) & 7.42 (\textcolor[HTML]{1400AD}{-24.12}) & 1.33 (\textcolor[HTML]{1400AD}{-24.74}) & 0.00 ($\:\:$\textcolor[HTML]{1400AD}{-1.35}) & 0.00 ($\:\:$\textcolor[HTML]{1400AD}{-1.65}) & 0.00 ($\:\:$\textcolor[HTML]{1400AD}{-1.66}) & 0.00 ($\:\:$\textcolor[HTML]{1400AD}{-1.70}) & 179 \\ 
& & LLaVA-NeXT (FT) & 8.39 ($\:\:$\textcolor[HTML]{1400AD}{-8.59}) & 11.41 (\textcolor[HTML]{1400AD}{-11.29}) & 8.71 (\textcolor[HTML]{1400AD}{-11.24}) & 1.43 ($\:\:$\textcolor[HTML]{1400AD}{-1.59}) & \textbf{1.99 ($\:\:$\textcolor[HTML]{1400AD}{-1.34)}} & \textbf{1.96 ($\:\:$\textcolor[HTML]{1400AD}{-1.27)}} & \textbf{1.95 ($\:\:$\textcolor[HTML]{1400AD}{-1.16)}} & 68 \\ 
& & Qwen-VL & \textbf{12.11 ($\:\:$\textcolor[HTML]{1400AD}{-9.00)}} & \textbf{17.23 ($\:\:$\textcolor[HTML]{1400AD}{-9.80)}} & \textbf{13.05 ($\:\:$\textcolor[HTML]{1400AD}{-8.73)}} & 1.44 ($\:\:$\textcolor[HTML]{1400AD}{-0.16}) & 1.45 ($\:\:$\textcolor[HTML]{1400AD}{-0.14}) & 1.40 ($\:\:$\textcolor[HTML]{1400AD}{-0.16}) & 1.34 ($\:\:$\textcolor[HTML]{1400AD}{-0.18}) & 96 \\ 
& & Qwen-VL (FT) & 7.19 (\textcolor[HTML]{1400AD}{-13.92}) & 11.24 (\textcolor[HTML]{1400AD}{-13.63}) & 8.34 (\textcolor[HTML]{1400AD}{-13.61}) & 0.45 ($\:\:$\textcolor[HTML]{1400AD}{-3.12}) & 0.90 ($\:\:$\textcolor[HTML]{1400AD}{-2.93}) & 0.88 ($\:\:$\textcolor[HTML]{1400AD}{-2.89}) & 0.89 ($\:\:$\textcolor[HTML]{1400AD}{-2.79}) & 175 \\ 
& & mPLUG-Owl2 & 10.26 ($\:\:$\textcolor[HTML]{1400AD}{-2.53}) & 15.51 ($\:\:$\textcolor[HTML]{1400AD}{-1.57}) & 10.99 ($\:\:$\textcolor[HTML]{1400AD}{-2.49}) & \textbf{1.72 ($\:\:$\textcolor[HTML]{1400AD}{-0.35)}} & 1.33 ($\:\:$\textcolor[HTML]{1400AD}{-0.35}) & 1.20 ($\:\:$\textcolor[HTML]{1400AD}{-0.39}) & 1.16 ($\:\:$\textcolor[HTML]{1400AD}{-0.40}) & 109 \\ 

\midrule
\multirow{5}{*}{En} & \multirow{5}{*}{De}  & LLaVA-NeXT & \textbf{14.03 (\textcolor[HTML]{1400AD}{-12.46)}} & \textbf{17.90 (\textcolor[HTML]{1400AD}{-13.64)}} & \textbf{16.51 ($\:\:$\textcolor[HTML]{1400AD}{-9.56)}} & \textbf{1.73 ($\:\:$\textcolor[HTML]{AD0000}{+0.38)}} & \textbf{1.70 ($\:\:$\textcolor[HTML]{AD0000}{+0.05)}} & \textbf{1.67 ($\:\:$\textcolor[HTML]{AD0000}{+0.01)}} & \textbf{1.82 ($\:\:$\textcolor[HTML]{AD0000}{+0.13)}} & 169 \\ 
& & LLaVA-NeXT (FT) & 6.83 (\textcolor[HTML]{1400AD}{-10.15}) & 9.54 (\textcolor[HTML]{1400AD}{-13.16}) & 8.23 (\textcolor[HTML]{1400AD}{-11.72}) & 0.86 ($\:\:$\textcolor[HTML]{1400AD}{-2.15}) & 0.74 ($\:\:$\textcolor[HTML]{1400AD}{-2.59}) & 0.77 ($\:\:$\textcolor[HTML]{1400AD}{-2.46}) & 0.78 ($\:\:$\textcolor[HTML]{1400AD}{-2.33}) & 82 \\ 
& & Qwen-VL & 10.64 (\textcolor[HTML]{1400AD}{-10.46}) & 13.95 (\textcolor[HTML]{1400AD}{-13.08}) & 13.21 ($\:\:$\textcolor[HTML]{1400AD}{-8.56}) & 1.16 ($\:\:$\textcolor[HTML]{1400AD}{-0.44}) & 1.24 ($\:\:$\textcolor[HTML]{1400AD}{-0.35}) & 1.21 ($\:\:$\textcolor[HTML]{1400AD}{-0.35}) & 1.40 ($\:\:$\textcolor[HTML]{1400AD}{-0.12}) & 111 \\ 
& & Qwen-VL (FT) & 7.98 (\textcolor[HTML]{1400AD}{-13.14}) & 11.08 (\textcolor[HTML]{1400AD}{-13.79}) & 9.86 (\textcolor[HTML]{1400AD}{-12.09}) & 0.80 ($\:\:$\textcolor[HTML]{1400AD}{-2.77}) & 0.65 ($\:\:$\textcolor[HTML]{1400AD}{-3.18}) & 0.74 ($\:\:$\textcolor[HTML]{1400AD}{-3.03}) & 0.74 ($\:\:$\textcolor[HTML]{1400AD}{-2.94}) & 203 \\ 
& & mPLUG-Owl2 & 8.81 ($\:\:$\textcolor[HTML]{1400AD}{-3.98}) & 12.12 ($\:\:$\textcolor[HTML]{1400AD}{-4.97}) & 10.54 ($\:\:$\textcolor[HTML]{1400AD}{-2.94}) & 0.72 ($\:\:$\textcolor[HTML]{1400AD}{-1.35}) & 0.76 ($\:\:$\textcolor[HTML]{1400AD}{-0.92}) & 0.74 ($\:\:$\textcolor[HTML]{1400AD}{-0.85}) & 0.70 ($\:\:$\textcolor[HTML]{1400AD}{-0.86}) & 98 \\ 

\midrule
\multirow{5}{*}{En} & \multirow{5}{*}{It}  & LLaVA-NeXT & \textbf{8.53 (\textcolor[HTML]{1400AD}{-17.95)}} & \textbf{13.33 (\textcolor[HTML]{1400AD}{-18.21)}} & \textbf{9.37 (\textcolor[HTML]{1400AD}{-16.70)}} & 0.86 ($\:\:$\textcolor[HTML]{1400AD}{-0.48}) & 0.87 ($\:\:$\textcolor[HTML]{1400AD}{-0.79}) & 1.06 ($\:\:$\textcolor[HTML]{1400AD}{-0.60}) & 1.05 ($\:\:$\textcolor[HTML]{1400AD}{-0.65}) & 171 \\ 
& & LLaVA-NeXT (FT) & 5.89 (\textcolor[HTML]{1400AD}{-11.09}) & 8.90 (\textcolor[HTML]{1400AD}{-13.80}) & 6.61 (\textcolor[HTML]{1400AD}{-13.34}) & 0.96 ($\:\:$\textcolor[HTML]{1400AD}{-2.06}) & \textbf{1.32 ($\:\:$\textcolor[HTML]{1400AD}{-2.01)}} & \textbf{1.32 ($\:\:$\textcolor[HTML]{1400AD}{-1.91)}} & \textbf{1.31 ($\:\:$\textcolor[HTML]{1400AD}{-1.80)}} & 66 \\ 
& & Qwen-VL & 7.23 (\textcolor[HTML]{1400AD}{-13.87}) & 11.43 (\textcolor[HTML]{1400AD}{-15.59}) & 8.71 (\textcolor[HTML]{1400AD}{-13.06}) & 0.51 ($\:\:$\textcolor[HTML]{1400AD}{-1.08}) & 0.62 ($\:\:$\textcolor[HTML]{1400AD}{-0.97}) & 0.65 ($\:\:$\textcolor[HTML]{1400AD}{-0.91}) & 0.63 ($\:\:$\textcolor[HTML]{1400AD}{-0.89}) & 107 \\ 
& & Qwen-VL (FT) & 5.51 (\textcolor[HTML]{1400AD}{-15.61}) & 8.17 (\textcolor[HTML]{1400AD}{-16.70}) & 6.53 (\textcolor[HTML]{1400AD}{-15.42}) & \textbf{1.14 ($\:\:$\textcolor[HTML]{1400AD}{-2.44)}} & 0.82 ($\:\:$\textcolor[HTML]{1400AD}{-3.01}) & 0.85 ($\:\:$\textcolor[HTML]{1400AD}{-2.93}) & 0.84 ($\:\:$\textcolor[HTML]{1400AD}{-2.84}) & 170 \\ 
& & mPLUG-Owl2 & 3.97 ($\:\:$\textcolor[HTML]{1400AD}{-8.82}) & 8.50 ($\:\:$\textcolor[HTML]{1400AD}{-8.58}) & 4.50 ($\:\:$\textcolor[HTML]{1400AD}{-8.98}) & 0.15 ($\:\:$\textcolor[HTML]{1400AD}{-1.92}) & 0.14 ($\:\:$\textcolor[HTML]{1400AD}{-1.53}) & 0.16 ($\:\:$\textcolor[HTML]{1400AD}{-1.43}) & 0.15 ($\:\:$\textcolor[HTML]{1400AD}{-1.41}) & 107 \\ 

\midrule
\multirow{5}{*}{En} & \multirow{5}{*}{Nl}  & LLaVA-NeXT & 12.21 (\textcolor[HTML]{1400AD}{-14.28}) & \textbf{17.83 (\textcolor[HTML]{1400AD}{-13.71)}} & 14.60 (\textcolor[HTML]{1400AD}{-11.46}) & 0.36 ($\:\:$\textcolor[HTML]{1400AD}{-0.99}) & 1.81 ($\:\:$\textcolor[HTML]{AD0000}{+0.15}) & 1.70 ($\:\:$\textcolor[HTML]{AD0000}{+0.04}) & \textbf{1.83 ($\:\:$\textcolor[HTML]{AD0000}{+0.13)}} & 178 \\ 
& & LLaVA-NeXT (FT) & 9.41 ($\:\:$\textcolor[HTML]{1400AD}{-7.56}) & 15.01 ($\:\:$\textcolor[HTML]{1400AD}{-7.69}) & 12.14 ($\:\:$\textcolor[HTML]{1400AD}{-7.81}) & \textbf{1.21 ($\:\:$\textcolor[HTML]{1400AD}{-1.81)}} & 1.07 ($\:\:$\textcolor[HTML]{1400AD}{-2.27}) & 0.91 ($\:\:$\textcolor[HTML]{1400AD}{-2.32}) & 1.02 ($\:\:$\textcolor[HTML]{1400AD}{-2.09}) & 119 \\ 
& & Qwen-VL & 11.07 (\textcolor[HTML]{1400AD}{-10.04}) & 16.44 (\textcolor[HTML]{1400AD}{-10.59}) & 12.73 ($\:\:$\textcolor[HTML]{1400AD}{-9.05}) & 0.89 ($\:\:$\textcolor[HTML]{1400AD}{-0.71}) & \textbf{1.90 ($\:\:$\textcolor[HTML]{AD0000}{+0.32)}} & \textbf{1.78 ($\:\:$\textcolor[HTML]{AD0000}{+0.22)}} & 1.80 ($\:\:$\textcolor[HTML]{AD0000}{+0.28}) & 132 \\ 
& & Qwen-VL (FT) & \textbf{12.67 ($\:\:$\textcolor[HTML]{1400AD}{-8.45)}} & 17.03 ($\:\:$\textcolor[HTML]{1400AD}{-7.84}) & \textbf{16.91 ($\:\:$\textcolor[HTML]{1400AD}{-5.04)}} & 1.02 ($\:\:$\textcolor[HTML]{1400AD}{-2.55}) & 0.96 ($\:\:$\textcolor[HTML]{1400AD}{-2.88}) & 0.95 ($\:\:$\textcolor[HTML]{1400AD}{-2.83}) & 1.01 ($\:\:$\textcolor[HTML]{1400AD}{-2.67}) & 181 \\ 
& & mPLUG-Owl2 & 8.27 ($\:\:$\textcolor[HTML]{1400AD}{-4.51}) & 13.46 ($\:\:$\textcolor[HTML]{1400AD}{-3.62}) & 9.06 ($\:\:$\textcolor[HTML]{1400AD}{-4.42}) & 0.46 ($\:\:$\textcolor[HTML]{1400AD}{-1.61}) & 0.43 ($\:\:$\textcolor[HTML]{1400AD}{-1.25}) & 0.41 ($\:\:$\textcolor[HTML]{1400AD}{-1.18}) & 0.41 ($\:\:$\textcolor[HTML]{1400AD}{-1.14}) & 100 \\ 

\midrule
\multirow{5}{*}{En} & \multirow{5}{*}{Sv}  & LLaVA-NeXT & \textbf{15.01 (\textcolor[HTML]{1400AD}{-11.48)}} & \textbf{18.65 (\textcolor[HTML]{1400AD}{-12.89)}} & \textbf{13.56 (\textcolor[HTML]{1400AD}{-12.51)}} & \textbf{1.29 ($\:\:$\textcolor[HTML]{1400AD}{-0.05)}} & 0.97 ($\:\:$\textcolor[HTML]{1400AD}{-0.69}) & \textbf{1.15 ($\:\:$\textcolor[HTML]{1400AD}{-0.51)}} & \textbf{1.09 ($\:\:$\textcolor[HTML]{1400AD}{-0.61)}} & 174 \\ 
& & LLaVA-NeXT (FT) & 10.00 ($\:\:$\textcolor[HTML]{1400AD}{-6.97}) & 12.43 (\textcolor[HTML]{1400AD}{-10.27}) & 10.54 ($\:\:$\textcolor[HTML]{1400AD}{-9.41}) & 0.84 ($\:\:$\textcolor[HTML]{1400AD}{-2.17}) & \textbf{1.08 ($\:\:$\textcolor[HTML]{1400AD}{-2.26)}} & 0.97 ($\:\:$\textcolor[HTML]{1400AD}{-2.26}) & 0.87 ($\:\:$\textcolor[HTML]{1400AD}{-2.24}) & 115 \\ 
& & Qwen-VL & 10.37 (\textcolor[HTML]{1400AD}{-10.74}) & 14.08 (\textcolor[HTML]{1400AD}{-12.94}) & 10.15 (\textcolor[HTML]{1400AD}{-11.62}) & 0.84 ($\:\:$\textcolor[HTML]{1400AD}{-0.76}) & 0.86 ($\:\:$\textcolor[HTML]{1400AD}{-0.72}) & 0.83 ($\:\:$\textcolor[HTML]{1400AD}{-0.73}) & 0.80 ($\:\:$\textcolor[HTML]{1400AD}{-0.72}) & 123 \\ 
& & Qwen-VL (FT) & 8.97 (\textcolor[HTML]{1400AD}{-12.14}) & 12.25 (\textcolor[HTML]{1400AD}{-12.61}) & 9.66 (\textcolor[HTML]{1400AD}{-12.29}) & 0.87 ($\:\:$\textcolor[HTML]{1400AD}{-2.70}) & 0.94 ($\:\:$\textcolor[HTML]{1400AD}{-2.89}) & 0.92 ($\:\:$\textcolor[HTML]{1400AD}{-2.86}) & 0.90 ($\:\:$\textcolor[HTML]{1400AD}{-2.78}) & 164 \\ 
& & mPLUG-Owl2 & 10.21 ($\:\:$\textcolor[HTML]{1400AD}{-2.57}) & 13.03 ($\:\:$\textcolor[HTML]{1400AD}{-4.05}) & 9.07 ($\:\:$\textcolor[HTML]{1400AD}{-4.41}) & 0.35 ($\:\:$\textcolor[HTML]{1400AD}{-1.72}) & 0.35 ($\:\:$\textcolor[HTML]{1400AD}{-1.33}) & 0.34 ($\:\:$\textcolor[HTML]{1400AD}{-1.25}) & 0.34 ($\:\:$\textcolor[HTML]{1400AD}{-1.22}) & 88 \\ 

\midrule
\multirow{5}{*}{En} & \multirow{5}{*}{Ru}  & LLaVA-NeXT & \textbf{10.32 (\textcolor[HTML]{1400AD}{-16.17)}} & \textbf{15.15 (\textcolor[HTML]{1400AD}{-16.39)}} & \textbf{8.53 (\textcolor[HTML]{1400AD}{-17.54)}} & \textbf{0.32 ($\:\:$\textcolor[HTML]{1400AD}{-1.02)}} & \textbf{0.36 ($\:\:$\textcolor[HTML]{1400AD}{-1.30)}} & \textbf{0.31 ($\:\:$\textcolor[HTML]{1400AD}{-1.35)}} & \textbf{0.32 ($\:\:$\textcolor[HTML]{1400AD}{-1.38)}} & 203 \\ 
& & LLaVA-NeXT (FT) & 0.55 (\textcolor[HTML]{1400AD}{-16.42}) & 1.87 (\textcolor[HTML]{1400AD}{-20.83}) & 0.49 (\textcolor[HTML]{1400AD}{-19.46}) & 0.00 ($\:\:$\textcolor[HTML]{1400AD}{-3.02}) & 0.02 ($\:\:$\textcolor[HTML]{1400AD}{-3.32}) & 0.02 ($\:\:$\textcolor[HTML]{1400AD}{-3.21}) & 0.01 ($\:\:$\textcolor[HTML]{1400AD}{-3.10}) & 85 \\ 
& & Qwen-VL & 4.59 (\textcolor[HTML]{1400AD}{-16.52}) & 8.05 (\textcolor[HTML]{1400AD}{-18.97}) & 3.51 (\textcolor[HTML]{1400AD}{-18.26}) & 0.02 ($\:\:$\textcolor[HTML]{1400AD}{-1.58}) & 0.07 ($\:\:$\textcolor[HTML]{1400AD}{-1.52}) & 0.07 ($\:\:$\textcolor[HTML]{1400AD}{-1.49}) & 0.07 ($\:\:$\textcolor[HTML]{1400AD}{-1.45}) & 113 \\ 
& & Qwen-VL (FT) & 0.00 (\textcolor[HTML]{1400AD}{-21.12}) & 0.95 (\textcolor[HTML]{1400AD}{-23.91}) & 0.00 (\textcolor[HTML]{1400AD}{-21.95}) & 0.00 ($\:\:$\textcolor[HTML]{1400AD}{-3.57}) & 0.00 ($\:\:$\textcolor[HTML]{1400AD}{-3.83}) & 0.00 ($\:\:$\textcolor[HTML]{1400AD}{-3.78}) & 0.00 ($\:\:$\textcolor[HTML]{1400AD}{-3.68}) & 169 \\ 
& & mPLUG-Owl2 & 5.99 ($\:\:$\textcolor[HTML]{1400AD}{-6.80}) & 8.68 ($\:\:$\textcolor[HTML]{1400AD}{-8.40}) & 4.88 ($\:\:$\textcolor[HTML]{1400AD}{-8.60}) & 0.00 ($\:\:$\textcolor[HTML]{1400AD}{-2.07}) & 0.02 ($\:\:$\textcolor[HTML]{1400AD}{-1.66}) & 0.01 ($\:\:$\textcolor[HTML]{1400AD}{-1.57}) & 0.01 ($\:\:$\textcolor[HTML]{1400AD}{-1.54}) & 99 \\ 

\midrule
\multirow{5}{*}{En} & \multirow{5}{*}{Ja}  & LLaVA-NeXT & \textbf{8.68 (\textcolor[HTML]{1400AD}{-17.81)}} & \textbf{8.68 (\textcolor[HTML]{1400AD}{-22.86)}} & \textbf{11.47 (\textcolor[HTML]{1400AD}{-14.60)}} & \textbf{0.80 ($\:\:$\textcolor[HTML]{1400AD}{-0.54)}} & \textbf{0.80 ($\:\:$\textcolor[HTML]{1400AD}{-0.85)}} & \textbf{0.80 ($\:\:$\textcolor[HTML]{1400AD}{-0.86)}} & \textbf{0.80 ($\:\:$\textcolor[HTML]{1400AD}{-0.90)}} & 211 \\ 
& & LLaVA-NeXT (FT) & 0.29 (\textcolor[HTML]{1400AD}{-16.68}) & 0.30 (\textcolor[HTML]{1400AD}{-22.40}) & 0.38 (\textcolor[HTML]{1400AD}{-19.57}) & 0.04 ($\:\:$\textcolor[HTML]{1400AD}{-2.98}) & 0.04 ($\:\:$\textcolor[HTML]{1400AD}{-3.29}) & 0.04 ($\:\:$\textcolor[HTML]{1400AD}{-3.19}) & 0.04 ($\:\:$\textcolor[HTML]{1400AD}{-3.07}) & 85 \\ 
& & Qwen-VL & 3.52 (\textcolor[HTML]{1400AD}{-17.59}) & 3.53 (\textcolor[HTML]{1400AD}{-23.49}) & 4.78 (\textcolor[HTML]{1400AD}{-17.00}) & 0.32 ($\:\:$\textcolor[HTML]{1400AD}{-1.28}) & 0.32 ($\:\:$\textcolor[HTML]{1400AD}{-1.27}) & 0.32 ($\:\:$\textcolor[HTML]{1400AD}{-1.24}) & 0.32 ($\:\:$\textcolor[HTML]{1400AD}{-1.20}) & 132 \\ 
& & Qwen-VL (FT) & 0.00 (\textcolor[HTML]{1400AD}{-21.12}) & 0.03 (\textcolor[HTML]{1400AD}{-24.84}) & 0.00 (\textcolor[HTML]{1400AD}{-21.95}) & 0.00 ($\:\:$\textcolor[HTML]{1400AD}{-3.57}) & 0.00 ($\:\:$\textcolor[HTML]{1400AD}{-3.83}) & 0.00 ($\:\:$\textcolor[HTML]{1400AD}{-3.78}) & 0.00 ($\:\:$\textcolor[HTML]{1400AD}{-3.68}) & 188 \\ 
& & mPLUG-Owl2 & 3.75 ($\:\:$\textcolor[HTML]{1400AD}{-9.04}) & 3.75 (\textcolor[HTML]{1400AD}{-13.33}) & 4.98 ($\:\:$\textcolor[HTML]{1400AD}{-8.49}) & 0.39 ($\:\:$\textcolor[HTML]{1400AD}{-1.68}) & 0.39 ($\:\:$\textcolor[HTML]{1400AD}{-1.28}) & 0.39 ($\:\:$\textcolor[HTML]{1400AD}{-1.20}) & 0.39 ($\:\:$\textcolor[HTML]{1400AD}{-1.17}) & 112 \\ 

\midrule
\multirow{5}{*}{En} & \multirow{5}{*}{Zh}  & LLaVA-NeXT & \textbf{14.00 ($\:\:$\textcolor[HTML]{1400AD}{-1.86)}} & \textbf{14.09 ($\:\:$\textcolor[HTML]{1400AD}{-6.86)}} & \textbf{16.69 ($\:\:$\textcolor[HTML]{AD0000}{+0.19)}} & 0.66 ($\:\:$\textcolor[HTML]{1400AD}{-0.42}) & 0.66 ($\:\:$\textcolor[HTML]{1400AD}{-0.58}) & 0.66 ($\:\:$\textcolor[HTML]{1400AD}{-0.56}) & 0.66 ($\:\:$\textcolor[HTML]{1400AD}{-0.59}) & 228 \\ 
& & LLaVA-NeXT (FT) & 0.14 (\textcolor[HTML]{1400AD}{-11.49}) & 0.39 (\textcolor[HTML]{1400AD}{-15.08}) & 0.15 (\textcolor[HTML]{1400AD}{-13.97}) & 0.00 ($\:\:$\textcolor[HTML]{1400AD}{-2.42}) & 0.00 ($\:\:$\textcolor[HTML]{1400AD}{-2.60}) & 0.00 ($\:\:$\textcolor[HTML]{1400AD}{-2.51}) & 0.00 ($\:\:$\textcolor[HTML]{1400AD}{-2.43}) & 92 \\ 
& & Qwen-VL & 10.69 ($\:\:$\textcolor[HTML]{1400AD}{-1.45}) & 10.70 ($\:\:$\textcolor[HTML]{1400AD}{-5.71}) & 12.71 ($\:\:$\textcolor[HTML]{AD0000}{+0.52}) & \textbf{0.74 ($\:\:$\textcolor[HTML]{1400AD}{-0.59)}} & \textbf{0.73 ($\:\:$\textcolor[HTML]{1400AD}{-0.44)}} & \textbf{0.73 ($\:\:$\textcolor[HTML]{1400AD}{-0.39)}} & \textbf{0.73 ($\:\:$\textcolor[HTML]{1400AD}{-0.35)}} & 138 \\ 
& & Qwen-VL (FT) & 0.37 (\textcolor[HTML]{1400AD}{-13.40}) & 0.75 (\textcolor[HTML]{1400AD}{-16.88}) & 0.51 (\textcolor[HTML]{1400AD}{-12.91}) & 0.01 ($\:\:$\textcolor[HTML]{1400AD}{-2.96}) & 0.01 ($\:\:$\textcolor[HTML]{1400AD}{-3.09}) & 0.01 ($\:\:$\textcolor[HTML]{1400AD}{-3.04}) & 0.01 ($\:\:$\textcolor[HTML]{1400AD}{-2.99}) & 154 \\ 
& & mPLUG-Owl2 & 6.38 ($\:\:$\textcolor[HTML]{1400AD}{-6.45}) & 6.40 (\textcolor[HTML]{1400AD}{-10.74}) & 7.75 ($\:\:$\textcolor[HTML]{1400AD}{-5.77}) & 0.32 ($\:\:$\textcolor[HTML]{1400AD}{-1.75}) & 0.32 ($\:\:$\textcolor[HTML]{1400AD}{-1.36}) & 0.32 ($\:\:$\textcolor[HTML]{1400AD}{-1.27}) & 0.32 ($\:\:$\textcolor[HTML]{1400AD}{-1.24}) & 108 \\

\bottomrule
\end{tabular}
}
\caption{Results of LVLMs in Alignment-10 Task  (\{En\}-\{Lang\}). Bold fonts indicate the best score for that language combination. The values are noted next to the output of the difference by the same model in the method with instruction and output in English (\{En\}-\{En\}). Red indicates a higher value than that method; blue indicates a lower value.}

\label{tab:result-score-10-en-lang}
\end{table*}

%% file: results/a10-lang-lang.tex
\begin{table*}[t]
\centering
\resizebox{\textwidth}{!}{
\footnotesize
\setlength{\tabcolsep}{5pt}
\begin{tabular}{@{}cccrrrrrrrc@{}}
\toprule 
\multicolumn{1}{c}{\multirow{2.5}{*}{Input}} & \multirow{2.5}{*}{Output} & \multirow{2.5}{*}{LVLM} & \multicolumn{2}{c}{Entity Cov.} & \multicolumn{1}{c}{\multirow{2.5}{*}{Entity F1}} & \multicolumn{4}{c}{Entity Cooccurrence} & \multirow{2.5}{*}{Avg. Length} \\
\cmidrule(lr){4-5} \cmidrule(lr){7-10}
& & & \multicolumn{1}{c}{exact} & \multicolumn{1}{c}{partial} & & \multicolumn{1}{c}{n=0} & \multicolumn{1}{c}{n=1} & \multicolumn{1}{c}{n=2} & \multicolumn{1}{c}{n=$\infty$} & \\
\midrule
\multirow{5}{*}{Es} & \multirow{5}{*}{Es} & LLaVA-NeXT & \textbf{17.26 ($\:\:$\textcolor[HTML]{1400AD}{-9.23)}} & \textbf{21.30 (\textcolor[HTML]{1400AD}{-10.25)}} & \textbf{17.05 ($\:\:$\textcolor[HTML]{1400AD}{-9.01)}} & 2.13 ($\:\:$\textcolor[HTML]{AD0000}{+0.78}) & 2.32 ($\:\:$\textcolor[HTML]{AD0000}{+0.67}) & 2.17 ($\:\:$\textcolor[HTML]{AD0000}{+0.51}) & 2.10 ($\:\:$\textcolor[HTML]{AD0000}{+0.40}) & 186 \\ 
& & LLaVA-NeXT (FT) & 12.82 ($\:\:$\textcolor[HTML]{1400AD}{-4.15}) & 16.84 ($\:\:$\textcolor[HTML]{1400AD}{-5.86}) & 12.77 ($\:\:$\textcolor[HTML]{1400AD}{-7.18}) & 0.88 ($\:\:$\textcolor[HTML]{1400AD}{-2.14}) & 1.03 ($\:\:$\textcolor[HTML]{1400AD}{-2.31}) & 1.26 ($\:\:$\textcolor[HTML]{1400AD}{-1.97}) & 1.11 ($\:\:$\textcolor[HTML]{1400AD}{-1.99}) & 147 \\ 
& & Qwen-VL & 14.68 ($\:\:$\textcolor[HTML]{1400AD}{-6.43}) & 18.43 ($\:\:$\textcolor[HTML]{1400AD}{-8.59}) & 14.35 ($\:\:$\textcolor[HTML]{1400AD}{-7.43}) & 2.40 ($\:\:$\textcolor[HTML]{AD0000}{+0.81}) & \textbf{2.42 ($\:\:$\textcolor[HTML]{AD0000}{+0.83)}} & \textbf{2.57 ($\:\:$\textcolor[HTML]{AD0000}{+1.01)}} & \textbf{2.56 ($\:\:$\textcolor[HTML]{AD0000}{+1.04)}} & 150 \\ 
& & Qwen-VL (FT) & 4.09 (\textcolor[HTML]{1400AD}{-17.02}) & 7.10 (\textcolor[HTML]{1400AD}{-17.77}) & 4.52 (\textcolor[HTML]{1400AD}{-17.42}) & 0.15 ($\:\:$\textcolor[HTML]{1400AD}{-3.43}) & 0.16 ($\:\:$\textcolor[HTML]{1400AD}{-3.68}) & 0.15 ($\:\:$\textcolor[HTML]{1400AD}{-3.63}) & 0.14 ($\:\:$\textcolor[HTML]{1400AD}{-3.54}) & 301 \\ 
& & mPLUG-Owl2 & 10.91 ($\:\:$\textcolor[HTML]{1400AD}{-1.87}) & 15.06 ($\:\:$\textcolor[HTML]{1400AD}{-2.02}) & 11.91 ($\:\:$\textcolor[HTML]{1400AD}{-1.57}) & \textbf{2.47 ($\:\:$\textcolor[HTML]{AD0000}{+0.40)}} & 2.07 ($\:\:$\textcolor[HTML]{AD0000}{+0.40}) & 2.02 ($\:\:$\textcolor[HTML]{AD0000}{+0.44}) & 1.99 ($\:\:$\textcolor[HTML]{AD0000}{+0.44}) & 135 \\ 

\midrule
\multirow{5}{*}{Fr} & \multirow{5}{*}{Fr} & LLaVA-NeXT & \textbf{24.35 ($\:\:$\textcolor[HTML]{1400AD}{-2.14)}} & \textbf{29.27 ($\:\:$\textcolor[HTML]{1400AD}{-2.27)}} & 24.38 ($\:\:$\textcolor[HTML]{1400AD}{-1.69}) & 0.95 ($\:\:$\textcolor[HTML]{1400AD}{-0.40}) & 0.90 ($\:\:$\textcolor[HTML]{1400AD}{-0.75}) & 0.88 ($\:\:$\textcolor[HTML]{1400AD}{-0.78}) & 0.90 ($\:\:$\textcolor[HTML]{1400AD}{-0.80}) & 211 \\ 
& & LLaVA-NeXT (FT) & 16.63 ($\:\:$\textcolor[HTML]{1400AD}{-0.35}) & 20.13 ($\:\:$\textcolor[HTML]{1400AD}{-2.57}) & 16.09 ($\:\:$\textcolor[HTML]{1400AD}{-3.86}) & 1.18 ($\:\:$\textcolor[HTML]{1400AD}{-1.83}) & 0.93 ($\:\:$\textcolor[HTML]{1400AD}{-2.41}) & 1.00 ($\:\:$\textcolor[HTML]{1400AD}{-2.23}) & 0.98 ($\:\:$\textcolor[HTML]{1400AD}{-2.13}) & 98 \\ 
& & Qwen-VL & 19.38 ($\:\:$\textcolor[HTML]{1400AD}{-1.73}) & 24.71 ($\:\:$\textcolor[HTML]{1400AD}{-2.32}) & 18.30 ($\:\:$\textcolor[HTML]{1400AD}{-3.47}) & 1.07 ($\:\:$\textcolor[HTML]{1400AD}{-0.53}) & 1.03 ($\:\:$\textcolor[HTML]{1400AD}{-0.55}) & 0.96 ($\:\:$\textcolor[HTML]{1400AD}{-0.60}) & 0.96 ($\:\:$\textcolor[HTML]{1400AD}{-0.56}) & 165 \\ 
& & Qwen-VL (FT) & 24.15 ($\:\:$\textcolor[HTML]{AD0000}{+3.04}) & 28.59 ($\:\:$\textcolor[HTML]{AD0000}{+3.73}) & \textbf{24.79 ($\:\:$\textcolor[HTML]{AD0000}{+2.85)}} & \textbf{3.83 ($\:\:$\textcolor[HTML]{AD0000}{+0.26)}} & \textbf{4.41 ($\:\:$\textcolor[HTML]{AD0000}{+0.58)}} & \textbf{4.51 ($\:\:$\textcolor[HTML]{AD0000}{+0.73)}} & \textbf{4.51 ($\:\:$\textcolor[HTML]{AD0000}{+0.83)}} & 219 \\ 
& & mPLUG-Owl2 & 17.43 ($\:\:$\textcolor[HTML]{AD0000}{+4.64}) & 22.48 ($\:\:$\textcolor[HTML]{AD0000}{+5.40}) & 17.78 ($\:\:$\textcolor[HTML]{AD0000}{+4.30}) & 0.85 ($\:\:$\textcolor[HTML]{1400AD}{-1.22}) & 0.65 ($\:\:$\textcolor[HTML]{1400AD}{-1.02}) & 0.75 ($\:\:$\textcolor[HTML]{1400AD}{-0.84}) & 0.73 ($\:\:$\textcolor[HTML]{1400AD}{-0.83}) & 158 \\ 

\midrule
\multirow{5}{*}{De} & \multirow{5}{*}{De} & LLaVA-NeXT & \textbf{17.45 ($\:\:$\textcolor[HTML]{1400AD}{-9.04)}} & \textbf{20.66 (\textcolor[HTML]{1400AD}{-10.89)}} & \textbf{21.05 ($\:\:$\textcolor[HTML]{1400AD}{-5.02)}} & 2.11 ($\:\:$\textcolor[HTML]{AD0000}{+0.77}) & \textbf{2.20 ($\:\:$\textcolor[HTML]{AD0000}{+0.55)}} & \textbf{2.22 ($\:\:$\textcolor[HTML]{AD0000}{+0.56)}} & \textbf{2.11 ($\:\:$\textcolor[HTML]{AD0000}{+0.41)}} & 204 \\ 
& & LLaVA-NeXT (FT) & 10.53 ($\:\:$\textcolor[HTML]{1400AD}{-6.44}) & 13.10 ($\:\:$\textcolor[HTML]{1400AD}{-9.60}) & 13.32 ($\:\:$\textcolor[HTML]{1400AD}{-6.63}) & 1.53 ($\:\:$\textcolor[HTML]{1400AD}{-1.49}) & 1.09 ($\:\:$\textcolor[HTML]{1400AD}{-2.25}) & 1.15 ($\:\:$\textcolor[HTML]{1400AD}{-2.07}) & 1.16 ($\:\:$\textcolor[HTML]{1400AD}{-1.95}) & 123 \\ 
& & Qwen-VL & 15.10 ($\:\:$\textcolor[HTML]{1400AD}{-6.00}) & 18.20 ($\:\:$\textcolor[HTML]{1400AD}{-8.82}) & 17.97 ($\:\:$\textcolor[HTML]{1400AD}{-3.81}) & \textbf{2.12 ($\:\:$\textcolor[HTML]{AD0000}{+0.52)}} & 1.99 ($\:\:$\textcolor[HTML]{AD0000}{+0.41}) & 2.08 ($\:\:$\textcolor[HTML]{AD0000}{+0.52}) & 1.99 ($\:\:$\textcolor[HTML]{AD0000}{+0.47}) & 160 \\ 
& & Qwen-VL (FT) & 7.74 (\textcolor[HTML]{1400AD}{-13.38}) & 9.58 (\textcolor[HTML]{1400AD}{-15.28}) & 9.23 (\textcolor[HTML]{1400AD}{-12.72}) & 0.37 ($\:\:$\textcolor[HTML]{1400AD}{-3.20}) & 0.40 ($\:\:$\textcolor[HTML]{1400AD}{-3.43}) & 0.43 ($\:\:$\textcolor[HTML]{1400AD}{-3.34}) & 0.40 ($\:\:$\textcolor[HTML]{1400AD}{-3.28}) & 287 \\ 
& & mPLUG-Owl2 & 14.33 ($\:\:$\textcolor[HTML]{AD0000}{+1.55}) & 17.63 ($\:\:$\textcolor[HTML]{AD0000}{+0.55}) & 16.73 ($\:\:$\textcolor[HTML]{AD0000}{+3.25}) & 1.99 ($\:\:$\textcolor[HTML]{1400AD}{-0.08}) & 1.92 ($\:\:$\textcolor[HTML]{AD0000}{+0.25}) & 1.94 ($\:\:$\textcolor[HTML]{AD0000}{+0.35}) & 1.81 ($\:\:$\textcolor[HTML]{AD0000}{+0.25}) & 143 \\ 

\midrule
\multirow{5}{*}{It} & \multirow{5}{*}{It} & LLaVA-NeXT &  \textbf{10.34 (\textcolor[HTML]{1400AD}{-16.14)}} & \textbf{15.43 (\textcolor[HTML]{1400AD}{-16.11)}} & \textbf{11.33 (\textcolor[HTML]{1400AD}{-14.74)}} & 1.16 ($\:\:$\textcolor[HTML]{1400AD}{-0.19}) & \textbf{0.93 ($\:\:$\textcolor[HTML]{1400AD}{-0.72)}} & \textbf{0.96 ($\:\:$\textcolor[HTML]{1400AD}{-0.70)}} & \textbf{0.96 ($\:\:$\textcolor[HTML]{1400AD}{-0.74)}} & 185 \\ 
& & LLaVA-NeXT (FT) &  5.73 (\textcolor[HTML]{1400AD}{-11.25}) & 9.84 (\textcolor[HTML]{1400AD}{-12.86}) & 6.45 (\textcolor[HTML]{1400AD}{-13.50}) & 0.31 ($\:\:$\textcolor[HTML]{1400AD}{-2.71}) & 0.25 ($\:\:$\textcolor[HTML]{1400AD}{-3.08}) & 0.25 ($\:\:$\textcolor[HTML]{1400AD}{-2.98}) & 0.23 ($\:\:$\textcolor[HTML]{1400AD}{-2.88}) & 91 \\ 
& & Qwen-VL &  9.97 (\textcolor[HTML]{1400AD}{-11.13}) & 14.20 (\textcolor[HTML]{1400AD}{-12.82}) & 11.09 (\textcolor[HTML]{1400AD}{-10.68}) & \textbf{1.16 ($\:\:$\textcolor[HTML]{1400AD}{-0.44)}} & 0.93 ($\:\:$\textcolor[HTML]{1400AD}{-0.65}) & 0.94 ($\:\:$\textcolor[HTML]{1400AD}{-0.62}) & 0.90 ($\:\:$\textcolor[HTML]{1400AD}{-0.62}) & 126 \\ 
& & Qwen-VL (FT) &  3.15 (\textcolor[HTML]{1400AD}{-17.96}) & 6.95 (\textcolor[HTML]{1400AD}{-17.92}) & 3.42 (\textcolor[HTML]{1400AD}{-18.53}) & 0.15 ($\:\:$\textcolor[HTML]{1400AD}{-3.42}) & 0.18 ($\:\:$\textcolor[HTML]{1400AD}{-3.65}) & 0.23 ($\:\:$\textcolor[HTML]{1400AD}{-3.54}) & 0.21 ($\:\:$\textcolor[HTML]{1400AD}{-3.47}) & 253 \\ 
& & mPLUG-Owl2 &  8.69 ($\:\:$\textcolor[HTML]{1400AD}{-4.10}) & 12.66 ($\:\:$\textcolor[HTML]{1400AD}{-4.42}) & 9.54 ($\:\:$\textcolor[HTML]{1400AD}{-3.94}) & 0.51 ($\:\:$\textcolor[HTML]{1400AD}{-1.56}) & 0.32 ($\:\:$\textcolor[HTML]{1400AD}{-1.36}) & 0.35 ($\:\:$\textcolor[HTML]{1400AD}{-1.24}) & 0.33 ($\:\:$\textcolor[HTML]{1400AD}{-1.23}) & 111 \\ 

\midrule
\multirow{5}{*}{Nl} & \multirow{5}{*}{Nl} & LLaVA-NeXT & 17.66 ($\:\:$\textcolor[HTML]{1400AD}{-8.83}) & 23.56 ($\:\:$\textcolor[HTML]{1400AD}{-7.99}) & \textbf{19.78 ($\:\:$\textcolor[HTML]{1400AD}{-6.28)}} & 0.79 ($\:\:$\textcolor[HTML]{1400AD}{-0.56}) & \textbf{3.55 ($\:\:$\textcolor[HTML]{AD0000}{+1.89)}} & 3.61 ($\:\:$\textcolor[HTML]{AD0000}{+1.95}) & 3.88 ($\:\:$\textcolor[HTML]{AD0000}{+2.18}) & 199 \\ 
& & LLaVA-NeXT (FT) & 15.57 ($\:\:$\textcolor[HTML]{1400AD}{-1.40}) & 20.79 ($\:\:$\textcolor[HTML]{1400AD}{-1.91}) & 16.87 ($\:\:$\textcolor[HTML]{1400AD}{-3.08}) & 1.66 ($\:\:$\textcolor[HTML]{1400AD}{-1.35}) & 3.38 ($\:\:$\textcolor[HTML]{AD0000}{+0.05}) & 3.32 ($\:\:$\textcolor[HTML]{AD0000}{+0.09}) & 3.47 ($\:\:$\textcolor[HTML]{AD0000}{+0.37}) & 183 \\ 
& & Qwen-VL & \textbf{19.41 ($\:\:$\textcolor[HTML]{1400AD}{-1.69)}} & \textbf{24.45 ($\:\:$\textcolor[HTML]{1400AD}{-2.58)}} & 19.65 ($\:\:$\textcolor[HTML]{1400AD}{-2.13}) & \textbf{2.13 ($\:\:$\textcolor[HTML]{AD0000}{+0.53)}} & 3.27 ($\:\:$\textcolor[HTML]{AD0000}{+1.69}) & \textbf{3.89 ($\:\:$\textcolor[HTML]{AD0000}{+2.33)}} & \textbf{4.04 ($\:\:$\textcolor[HTML]{AD0000}{+2.52)}} & 172 \\ 
& & Qwen-VL (FT) & 12.68 ($\:\:$\textcolor[HTML]{1400AD}{-8.43}) & 18.46 ($\:\:$\textcolor[HTML]{1400AD}{-6.41}) & 16.72 ($\:\:$\textcolor[HTML]{1400AD}{-5.22}) & 1.09 ($\:\:$\textcolor[HTML]{1400AD}{-2.48}) & 1.66 ($\:\:$\textcolor[HTML]{1400AD}{-2.18}) & 1.81 ($\:\:$\textcolor[HTML]{1400AD}{-1.96}) & 1.80 ($\:\:$\textcolor[HTML]{1400AD}{-1.88}) & 300 \\ 
& & mPLUG-Owl2 & 10.78 ($\:\:$\textcolor[HTML]{1400AD}{-2.01}) & 15.43 ($\:\:$\textcolor[HTML]{1400AD}{-1.66}) & 12.81 ($\:\:$\textcolor[HTML]{1400AD}{-0.67}) & 0.15 ($\:\:$\textcolor[HTML]{1400AD}{-1.92}) & 1.08 ($\:\:$\textcolor[HTML]{1400AD}{-0.60}) & 1.05 ($\:\:$\textcolor[HTML]{1400AD}{-0.54}) & 1.12 ($\:\:$\textcolor[HTML]{1400AD}{-0.43}) & 114 \\ 

\midrule
\multirow{5}{*}{Sv} & \multirow{5}{*}{Sv} & LLaVA-NeXT &  \textbf{27.51 ($\:\:$\textcolor[HTML]{AD0000}{+1.02)}} & \textbf{29.61 ($\:\:$\textcolor[HTML]{1400AD}{-1.93)}} & 16.71 ($\:\:$\textcolor[HTML]{1400AD}{-9.36}) & 2.10 ($\:\:$\textcolor[HTML]{AD0000}{+0.75}) & 0.87 ($\:\:$\textcolor[HTML]{1400AD}{-0.78}) & 0.89 ($\:\:$\textcolor[HTML]{1400AD}{-0.77}) & 0.90 ($\:\:$\textcolor[HTML]{1400AD}{-0.79}) & 206 \\ 
& & LLaVA-NeXT (FT) &  22.83 ($\:\:$\textcolor[HTML]{AD0000}{+5.86}) & 25.10 ($\:\:$\textcolor[HTML]{AD0000}{+2.40}) & 12.17 ($\:\:$\textcolor[HTML]{1400AD}{-7.78}) & 2.82 ($\:\:$\textcolor[HTML]{1400AD}{-0.20}) & 1.11 ($\:\:$\textcolor[HTML]{1400AD}{-2.22}) & 1.17 ($\:\:$\textcolor[HTML]{1400AD}{-2.06}) & 1.16 ($\:\:$\textcolor[HTML]{1400AD}{-1.94}) & 169 \\ 
& & Qwen-VL &  24.02 ($\:\:$\textcolor[HTML]{AD0000}{+2.92}) & 26.69 ($\:\:$\textcolor[HTML]{1400AD}{-0.34}) & \textbf{19.18 ($\:\:$\textcolor[HTML]{1400AD}{-2.60)}} & \textbf{3.60 ($\:\:$\textcolor[HTML]{AD0000}{+2.00)}} & \textbf{1.53 ($\:\:$\textcolor[HTML]{1400AD}{-0.06)}} & \textbf{1.54 ($\:\:$\textcolor[HTML]{1400AD}{-0.02)}} & \textbf{1.50 ($\:\:$\textcolor[HTML]{1400AD}{-0.02)}} & 147 \\ 
& & Qwen-VL (FT) &  16.04 ($\:\:$\textcolor[HTML]{1400AD}{-5.07}) & 18.10 ($\:\:$\textcolor[HTML]{1400AD}{-6.77}) & 6.15 (\textcolor[HTML]{1400AD}{-15.80}) & 0.23 ($\:\:$\textcolor[HTML]{1400AD}{-3.35}) & 0.18 ($\:\:$\textcolor[HTML]{1400AD}{-3.65}) & 0.20 ($\:\:$\textcolor[HTML]{1400AD}{-3.57}) & 0.21 ($\:\:$\textcolor[HTML]{1400AD}{-3.47}) & 242 \\ 
& & mPLUG-Owl2 &  21.40 ($\:\:$\textcolor[HTML]{AD0000}{+8.61}) & 23.51 ($\:\:$\textcolor[HTML]{AD0000}{+6.43}) & 13.84 ($\:\:$\textcolor[HTML]{AD0000}{+0.36}) & 2.01 ($\:\:$\textcolor[HTML]{1400AD}{-0.06}) & 1.07 ($\:\:$\textcolor[HTML]{1400AD}{-0.61}) & 1.06 ($\:\:$\textcolor[HTML]{1400AD}{-0.52}) & 1.05 ($\:\:$\textcolor[HTML]{1400AD}{-0.51}) & 111 \\ 

\midrule
\multirow{5}{*}{Ru} & \multirow{5}{*}{Ru} & LLaVA-NeXT & \textbf{14.38 (\textcolor[HTML]{1400AD}{-12.11)}} & \textbf{17.43 (\textcolor[HTML]{1400AD}{-14.11)}} & \textbf{9.81 (\textcolor[HTML]{1400AD}{-16.26)}} & 0.26 ($\:\:$\textcolor[HTML]{1400AD}{-1.08}) & \textbf{0.45 ($\:\:$\textcolor[HTML]{1400AD}{-1.20)}} & \textbf{0.42 ($\:\:$\textcolor[HTML]{1400AD}{-1.24)}} & \textbf{0.41 ($\:\:$\textcolor[HTML]{1400AD}{-1.29)}} & 219 \\ 
& & LLaVA-NeXT (FT) & 10.74 ($\:\:$\textcolor[HTML]{1400AD}{-6.24}) & 13.67 ($\:\:$\textcolor[HTML]{1400AD}{-9.03}) & 6.55 (\textcolor[HTML]{1400AD}{-13.40}) & 0.32 ($\:\:$\textcolor[HTML]{1400AD}{-2.70}) & 0.37 ($\:\:$\textcolor[HTML]{1400AD}{-2.96}) & 0.36 ($\:\:$\textcolor[HTML]{1400AD}{-2.87}) & 0.36 ($\:\:$\textcolor[HTML]{1400AD}{-2.75}) & 184 \\ 
& & Qwen-VL & 6.80 (\textcolor[HTML]{1400AD}{-14.31}) & 9.68 (\textcolor[HTML]{1400AD}{-17.34}) & 4.63 (\textcolor[HTML]{1400AD}{-17.15}) & 0.31 ($\:\:$\textcolor[HTML]{1400AD}{-1.29}) & 0.32 ($\:\:$\textcolor[HTML]{1400AD}{-1.27}) & 0.30 ($\:\:$\textcolor[HTML]{1400AD}{-1.26}) & 0.31 ($\:\:$\textcolor[HTML]{1400AD}{-1.21}) & 170 \\ 
& & Qwen-VL (FT) & 1.76 (\textcolor[HTML]{1400AD}{-19.35}) & 3.60 (\textcolor[HTML]{1400AD}{-21.27}) & 1.52 (\textcolor[HTML]{1400AD}{-20.42}) & 0.14 ($\:\:$\textcolor[HTML]{1400AD}{-3.43}) & 0.14 ($\:\:$\textcolor[HTML]{1400AD}{-3.69}) & 0.14 ($\:\:$\textcolor[HTML]{1400AD}{-3.64}) & 0.14 ($\:\:$\textcolor[HTML]{1400AD}{-3.54}) & 324 \\ 
& & mPLUG-Owl2 & 7.07 ($\:\:$\textcolor[HTML]{1400AD}{-5.72}) & 8.92 ($\:\:$\textcolor[HTML]{1400AD}{-8.16}) & 5.57 ($\:\:$\textcolor[HTML]{1400AD}{-7.91}) & \textbf{0.51 ($\:\:$\textcolor[HTML]{1400AD}{-1.56)}} & 0.34 ($\:\:$\textcolor[HTML]{1400AD}{-1.33}) & 0.31 ($\:\:$\textcolor[HTML]{1400AD}{-1.28}) & 0.35 ($\:\:$\textcolor[HTML]{1400AD}{-1.21}) & 129 \\ 

\midrule
\multirow{5}{*}{Ja} & \multirow{5}{*}{Ja} & LLaVA-NeXT & \textbf{13.38 (\textcolor[HTML]{1400AD}{-13.11)}} & \textbf{13.38 (\textcolor[HTML]{1400AD}{-18.17)}} & \textbf{17.68 ($\:\:$\textcolor[HTML]{1400AD}{-8.39)}} & 0.73 ($\:\:$\textcolor[HTML]{1400AD}{-0.61}) & 0.83 ($\:\:$\textcolor[HTML]{1400AD}{-0.83}) & 0.83 ($\:\:$\textcolor[HTML]{1400AD}{-0.83}) & 0.83 ($\:\:$\textcolor[HTML]{1400AD}{-0.87}) & 249 \\ 
& & LLaVA-NeXT (FT) & 7.51 ($\:\:$\textcolor[HTML]{1400AD}{-9.46}) & 7.51 (\textcolor[HTML]{1400AD}{-15.19}) & 7.80 (\textcolor[HTML]{1400AD}{-12.15}) & 1.14 ($\:\:$\textcolor[HTML]{1400AD}{-1.88}) & 1.14 ($\:\:$\textcolor[HTML]{1400AD}{-2.19}) & 1.14 ($\:\:$\textcolor[HTML]{1400AD}{-2.09}) & 1.14 ($\:\:$\textcolor[HTML]{1400AD}{-1.97}) & 167 \\ 
& & Qwen-VL & 10.89 (\textcolor[HTML]{1400AD}{-10.22}) & 10.90 (\textcolor[HTML]{1400AD}{-16.13}) & 14.56 ($\:\:$\textcolor[HTML]{1400AD}{-7.22}) & 0.92 ($\:\:$\textcolor[HTML]{1400AD}{-0.68}) & 0.92 ($\:\:$\textcolor[HTML]{1400AD}{-0.67}) & 0.92 ($\:\:$\textcolor[HTML]{1400AD}{-0.64}) & 0.92 ($\:\:$\textcolor[HTML]{1400AD}{-0.60}) & 154 \\ 
& & Qwen-VL (FT) & 0.86 (\textcolor[HTML]{1400AD}{-20.26}) & 0.88 (\textcolor[HTML]{1400AD}{-23.99}) & 1.12 (\textcolor[HTML]{1400AD}{-20.83}) & 0.03 ($\:\:$\textcolor[HTML]{1400AD}{-3.55}) & 0.03 ($\:\:$\textcolor[HTML]{1400AD}{-3.81}) & 0.03 ($\:\:$\textcolor[HTML]{1400AD}{-3.75}) & 0.03 ($\:\:$\textcolor[HTML]{1400AD}{-3.65}) & 278 \\ 
& & mPLUG-Owl2 & 6.91 ($\:\:$\textcolor[HTML]{1400AD}{-5.88}) & 6.93 (\textcolor[HTML]{1400AD}{-10.15}) & 9.34 ($\:\:$\textcolor[HTML]{1400AD}{-4.14}) & \textbf{1.20 ($\:\:$\textcolor[HTML]{1400AD}{-0.87)}} & \textbf{1.21 ($\:\:$\textcolor[HTML]{1400AD}{-0.46)}} & \textbf{1.21 ($\:\:$\textcolor[HTML]{1400AD}{-0.38)}} & \textbf{1.21 ($\:\:$\textcolor[HTML]{1400AD}{-0.35)}} & 144 \\ 

\midrule
\multirow{5}{*}{Zh} & \multirow{5}{*}{Zh} & LLaVA-NeXT & 13.78 ($\:\:$\textcolor[HTML]{1400AD}{-2.08}) & 13.78 ($\:\:$\textcolor[HTML]{1400AD}{-7.17}) & 17.00 ($\:\:$\textcolor[HTML]{AD0000}{+0.50}) & 0.54 ($\:\:$\textcolor[HTML]{1400AD}{-0.54}) & 0.53 ($\:\:$\textcolor[HTML]{1400AD}{-0.70}) & 0.53 ($\:\:$\textcolor[HTML]{1400AD}{-0.69}) & 0.53 ($\:\:$\textcolor[HTML]{1400AD}{-0.72}) & 246 \\ 
& & LLaVA-NeXT (FT) & 6.93 ($\:\:$\textcolor[HTML]{1400AD}{-4.71}) & 6.97 ($\:\:$\textcolor[HTML]{1400AD}{-8.50}) & 7.31 ($\:\:$\textcolor[HTML]{1400AD}{-6.81}) & 0.78 ($\:\:$\textcolor[HTML]{1400AD}{-1.64}) & 0.78 ($\:\:$\textcolor[HTML]{1400AD}{-1.83}) & 0.78 ($\:\:$\textcolor[HTML]{1400AD}{-1.73}) & 0.78 ($\:\:$\textcolor[HTML]{1400AD}{-1.65}) & 170 \\ 
& & Qwen-VL & \textbf{17.90 ($\:\:$\textcolor[HTML]{AD0000}{+5.76)}} & \textbf{17.90 ($\:\:$\textcolor[HTML]{AD0000}{+1.48)}} & \textbf{22.12 ($\:\:$\textcolor[HTML]{AD0000}{+9.93)}} & \textbf{3.31 ($\:\:$\textcolor[HTML]{AD0000}{+1.97)}} & \textbf{3.30 ($\:\:$\textcolor[HTML]{AD0000}{+2.13)}} & \textbf{3.30 ($\:\:$\textcolor[HTML]{AD0000}{+2.18)}} & \textbf{3.30 ($\:\:$\textcolor[HTML]{AD0000}{+2.22)}} & 155 \\ 
& & Qwen-VL (FT) & 0.22 (\textcolor[HTML]{1400AD}{-13.55}) & 0.33 (\textcolor[HTML]{1400AD}{-17.29}) & 0.27 (\textcolor[HTML]{1400AD}{-13.16}) & 0.00 ($\:\:$\textcolor[HTML]{1400AD}{-2.97}) & 0.00 ($\:\:$\textcolor[HTML]{1400AD}{-3.10}) & 0.00 ($\:\:$\textcolor[HTML]{1400AD}{-3.06}) & 0.00 ($\:\:$\textcolor[HTML]{1400AD}{-3.00}) & 249 \\ 
& & mPLUG-Owl2 & 9.03 ($\:\:$\textcolor[HTML]{1400AD}{-3.80}) & 9.05 ($\:\:$\textcolor[HTML]{1400AD}{-8.08}) & 12.98 ($\:\:$\textcolor[HTML]{1400AD}{-0.55}) & 0.77 ($\:\:$\textcolor[HTML]{1400AD}{-1.31}) & 0.77 ($\:\:$\textcolor[HTML]{1400AD}{-0.91}) & 0.77 ($\:\:$\textcolor[HTML]{1400AD}{-0.82}) & 0.77 ($\:\:$\textcolor[HTML]{1400AD}{-0.80}) & 150 \\ 

\bottomrule
\end{tabular}
}
\caption{Results of LVLMs in Alignment-10 Task (the format with instruction and output in each of the ten languages, \{Lang\}-\{Lang\}). Bold fonts indicate the best score for that language combination. The values are noted next to the differences output by the same model in the format with instruction and output in English (\{En\}-\{En\}). Red indicates a higher value than \{En\}-\{En\}; blue indicates a lower value.}

\label{tab:result-score-10-lang-lang}
\end{table*}

%% file: results/full.tex
\begin{table*}[h!]
\centering
\resizebox{\textwidth}{!}{
\footnotesize
\setlength{\tabcolsep}{5pt}
\begin{tabular}{@{}cccrrrrrrrc@{}}
\toprule 
\multicolumn{1}{c}{\multirow{2.5}{*}{Input}} & \multirow{2.5}{*}{Output} & \multirow{2.5}{*}{LVLM} & \multicolumn{2}{c}{Entity Cov.} & \multirow{2.5}{*}{Entity F1} & \multicolumn{4}{c}{Entity Cooccurrence} & \multirow{2.5}{*}{Avg. Length} \\
\cmidrule(lr){4-5} \cmidrule(lr){7-10}
& & & \multicolumn{1}{c}{exact} & \multicolumn{1}{c}{partial} & & \multicolumn{1}{c}{n=0} & \multicolumn{1}{c}{n=1} & \multicolumn{1}{c}{n=2} & \multicolumn{1}{c}{n=$\infty$} & \\
\midrule

\multirow{4.5}{*}{En} & \multirow{4.5}{*}{En}  & LLaVA-NeXT & 17.66 & 26.05 & 18.55 & 1.31 & 1.23 & 1.20 & 1.20 & 242 \\ 
& & LLaVA-NeXT (FT) & 17.92 & 23.65 & 19.20 & \textbf{5.67} & 5.66 & 5.63 & 5.60 & 81 \\ 
& & Qwen-VL & 14.60 & 21.51 & 15.39 & 1.36 & 1.33 & 1.28 & 1.27 & 110 \\ 
& & Qwen-VL (FT) & \textbf{20.09} & \textbf{26.27} & \textbf{20.84} & 5.67 & \textbf{5.78} & \textbf{5.77} & \textbf{5.72} & 171 \\ 
& & mPLUG-Owl2 & 14.41 & 21.96 & 15.71 & 1.27 & 1.17 & 1.14 & 1.10 & 121 \\ 

\midrule
\multirow{4.5}{*}{En} & \multirow{4.5}{*}{Es}  & LLaVA-NeXT & \textbf{10.40 ($\:\:$\textcolor[HTML]{1400AD}{-7.26)}} & \textbf{16.05 (\textcolor[HTML]{1400AD}{-10.00)}} & \textbf{10.86 ($\:\:$\textcolor[HTML]{1400AD}{-7.69)}} & \textbf{0.79 ($\:\:$\textcolor[HTML]{1400AD}{-0.52)}} & \textbf{0.78 ($\:\:$\textcolor[HTML]{1400AD}{-0.45)}} & \textbf{0.83 ($\:\:$\textcolor[HTML]{1400AD}{-0.37)}} & \textbf{0.83 ($\:\:$\textcolor[HTML]{1400AD}{-0.37)}} & 181 \\ 
& & LLaVA-NeXT (FT) & 4.96 (\textcolor[HTML]{1400AD}{-12.96}) & 8.42 (\textcolor[HTML]{1400AD}{-15.23}) & 5.40 (\textcolor[HTML]{1400AD}{-13.80}) & 0.56 ($\:\:$\textcolor[HTML]{1400AD}{-5.10}) & 0.57 ($\:\:$\textcolor[HTML]{1400AD}{-5.10}) & 0.58 ($\:\:$\textcolor[HTML]{1400AD}{-5.04}) & 0.57 ($\:\:$\textcolor[HTML]{1400AD}{-5.02}) & 90 \\ 
& & Qwen-VL & 8.11 ($\:\:$\textcolor[HTML]{1400AD}{-6.49}) & 13.18 ($\:\:$\textcolor[HTML]{1400AD}{-8.33}) & 8.66 ($\:\:$\textcolor[HTML]{1400AD}{-6.73}) & 0.53 ($\:\:$\textcolor[HTML]{1400AD}{-0.83}) & 0.50 ($\:\:$\textcolor[HTML]{1400AD}{-0.83}) & 0.52 ($\:\:$\textcolor[HTML]{1400AD}{-0.76}) & 0.51 ($\:\:$\textcolor[HTML]{1400AD}{-0.76}) & 103 \\ 
& & Qwen-VL (FT) & 4.23 (\textcolor[HTML]{1400AD}{-15.86}) & 8.47 (\textcolor[HTML]{1400AD}{-17.80}) & 4.66 (\textcolor[HTML]{1400AD}{-16.17}) & 0.23 ($\:\:$\textcolor[HTML]{1400AD}{-5.43}) & 0.23 ($\:\:$\textcolor[HTML]{1400AD}{-5.56}) & 0.24 ($\:\:$\textcolor[HTML]{1400AD}{-5.53}) & 0.24 ($\:\:$\textcolor[HTML]{1400AD}{-5.48}) & 195 \\ 
& & mPLUG-Owl2 & 7.26 ($\:\:$\textcolor[HTML]{1400AD}{-7.14}) & 12.13 ($\:\:$\textcolor[HTML]{1400AD}{-9.83}) & 7.55 ($\:\:$\textcolor[HTML]{1400AD}{-8.16}) & 0.45 ($\:\:$\textcolor[HTML]{1400AD}{-0.82}) & 0.49 ($\:\:$\textcolor[HTML]{1400AD}{-0.68}) & 0.52 ($\:\:$\textcolor[HTML]{1400AD}{-0.63}) & 0.51 ($\:\:$\textcolor[HTML]{1400AD}{-0.59}) & 100 \\ 

\midrule
\multirow{4.5}{*}{En} & \multirow{4.5}{*}{Fr}  & LLaVA-NeXT & \textbf{9.71 ($\:\:$\textcolor[HTML]{1400AD}{-7.95)}} & \textbf{16.17 ($\:\:$\textcolor[HTML]{1400AD}{-9.88)}} & \textbf{9.49 ($\:\:$\textcolor[HTML]{1400AD}{-9.06)}} & \textbf{0.57 ($\:\:$\textcolor[HTML]{1400AD}{-0.74)}} & \textbf{0.57 ($\:\:$\textcolor[HTML]{1400AD}{-0.66)}} & \textbf{0.57 ($\:\:$\textcolor[HTML]{1400AD}{-0.63)}} & \textbf{0.55 ($\:\:$\textcolor[HTML]{1400AD}{-0.64)}} & 168 \\ 
& & LLaVA-NeXT (FT) & 7.02 (\textcolor[HTML]{1400AD}{-10.90}) & 10.37 (\textcolor[HTML]{1400AD}{-13.29}) & 7.60 (\textcolor[HTML]{1400AD}{-11.60}) & \textbf{0.84 ($\:\:$\textcolor[HTML]{1400AD}{-4.83)}} & \textbf{0.84 ($\:\:$\textcolor[HTML]{1400AD}{-4.82)}} & \textbf{0.82 ($\:\:$\textcolor[HTML]{1400AD}{-4.81)}} & \textbf{0.81 ($\:\:$\textcolor[HTML]{1400AD}{-4.79)}} & 60 \\ 
& & Qwen-VL & 7.64 ($\:\:$\textcolor[HTML]{1400AD}{-6.96}) & 12.82 ($\:\:$\textcolor[HTML]{1400AD}{-8.68}) & 7.71 ($\:\:$\textcolor[HTML]{1400AD}{-7.68}) & 0.51 ($\:\:$\textcolor[HTML]{1400AD}{-0.85}) & 0.46 ($\:\:$\textcolor[HTML]{1400AD}{-0.87}) & 0.45 ($\:\:$\textcolor[HTML]{1400AD}{-0.83}) & 0.43 ($\:\:$\textcolor[HTML]{1400AD}{-0.83}) & 86 \\ 
& & Qwen-VL (FT) & 6.42 (\textcolor[HTML]{1400AD}{-13.68}) & 11.17 (\textcolor[HTML]{1400AD}{-15.10}) & 6.88 (\textcolor[HTML]{1400AD}{-13.95}) & 0.43 ($\:\:$\textcolor[HTML]{1400AD}{-5.24}) & 0.56 ($\:\:$\textcolor[HTML]{1400AD}{-5.22}) & 0.55 ($\:\:$\textcolor[HTML]{1400AD}{-5.22}) & 0.54 ($\:\:$\textcolor[HTML]{1400AD}{-5.18}) & 155 \\ 
& & mPLUG-Owl2 & 6.99 ($\:\:$\textcolor[HTML]{1400AD}{-7.42}) & 12.55 ($\:\:$\textcolor[HTML]{1400AD}{-9.41}) & 6.91 ($\:\:$\textcolor[HTML]{1400AD}{-8.79}) & 0.41 ($\:\:$\textcolor[HTML]{1400AD}{-0.86}) & 0.38 ($\:\:$\textcolor[HTML]{1400AD}{-0.79}) & 0.37 ($\:\:$\textcolor[HTML]{1400AD}{-0.77}) & 0.35 ($\:\:$\textcolor[HTML]{1400AD}{-0.75}) & 95 \\ 

\midrule
\multirow{4.5}{*}{En} & \multirow{4.5}{*}{De}  & LLaVA-NeXT & \textbf{10.32 ($\:\:$\textcolor[HTML]{1400AD}{-7.35)}} & \textbf{13.84 (\textcolor[HTML]{1400AD}{-12.21)}} & \textbf{12.28 ($\:\:$\textcolor[HTML]{1400AD}{-6.27)}} & \textbf{0.90 ($\:\:$\textcolor[HTML]{1400AD}{-0.41)}} & \textbf{0.88 ($\:\:$\textcolor[HTML]{1400AD}{-0.34)}} & \textbf{0.88 ($\:\:$\textcolor[HTML]{1400AD}{-0.32)}} & \textbf{0.86 ($\:\:$\textcolor[HTML]{1400AD}{-0.34)}} & 161 \\ 
& & LLaVA-NeXT (FT) & 5.52 (\textcolor[HTML]{1400AD}{-12.40}) & 7.80 (\textcolor[HTML]{1400AD}{-15.86}) & 5.93 (\textcolor[HTML]{1400AD}{-13.26}) & 0.52 ($\:\:$\textcolor[HTML]{1400AD}{-5.15}) & 0.48 ($\:\:$\textcolor[HTML]{1400AD}{-5.19}) & 0.46 ($\:\:$\textcolor[HTML]{1400AD}{-5.17}) & 0.45 ($\:\:$\textcolor[HTML]{1400AD}{-5.15}) & 75 \\ 
& & Qwen-VL & 7.75 ($\:\:$\textcolor[HTML]{1400AD}{-6.85}) & 10.60 (\textcolor[HTML]{1400AD}{-10.91}) & 8.69 ($\:\:$\textcolor[HTML]{1400AD}{-6.69}) & 0.63 ($\:\:$\textcolor[HTML]{1400AD}{-0.73}) & 0.59 ($\:\:$\textcolor[HTML]{1400AD}{-0.74}) & 0.58 ($\:\:$\textcolor[HTML]{1400AD}{-0.70}) & 0.56 ($\:\:$\textcolor[HTML]{1400AD}{-0.71}) & 99 \\ 
& & Qwen-VL (FT) & 4.79 (\textcolor[HTML]{1400AD}{-15.30}) & 7.40 (\textcolor[HTML]{1400AD}{-18.87}) & 5.17 (\textcolor[HTML]{1400AD}{-15.67}) & 0.23 ($\:\:$\textcolor[HTML]{1400AD}{-5.44}) & 0.25 ($\:\:$\textcolor[HTML]{1400AD}{-5.53}) & 0.24 ($\:\:$\textcolor[HTML]{1400AD}{-5.53}) & 0.24 ($\:\:$\textcolor[HTML]{1400AD}{-5.48}) & 177 \\ 
& & mPLUG-Owl2 & 6.87 ($\:\:$\textcolor[HTML]{1400AD}{-7.53}) & 9.66 (\textcolor[HTML]{1400AD}{-12.30}) & 7.69 ($\:\:$\textcolor[HTML]{1400AD}{-8.01}) & 0.60 ($\:\:$\textcolor[HTML]{1400AD}{-0.67}) & 0.54 ($\:\:$\textcolor[HTML]{1400AD}{-0.63}) & 0.53 ($\:\:$\textcolor[HTML]{1400AD}{-0.61}) & 0.50 ($\:\:$\textcolor[HTML]{1400AD}{-0.60}) & 91 \\ 

\midrule
\multirow{4.5}{*}{En} & \multirow{4.5}{*}{It}  & LLaVA-NeXT & \textbf{9.57 ($\:\:$\textcolor[HTML]{1400AD}{-8.10)}} & \textbf{16.52 ($\:\:$\textcolor[HTML]{1400AD}{-9.53)}} & \textbf{10.72 ($\:\:$\textcolor[HTML]{1400AD}{-7.83)}} & 0.72 ($\:\:$\textcolor[HTML]{1400AD}{-0.60}) & 0.72 ($\:\:$\textcolor[HTML]{1400AD}{-0.50}) & 0.74 ($\:\:$\textcolor[HTML]{1400AD}{-0.46}) & 0.72 ($\:\:$\textcolor[HTML]{1400AD}{-0.47}) & 168 \\ 
& & LLaVA-NeXT (FT) & 6.21 (\textcolor[HTML]{1400AD}{-11.71}) & 9.51 (\textcolor[HTML]{1400AD}{-14.15}) & 7.59 (\textcolor[HTML]{1400AD}{-11.61}) & \textbf{0.79 ($\:\:$\textcolor[HTML]{1400AD}{-4.88)}} & \textbf{0.86 ($\:\:$\textcolor[HTML]{1400AD}{-4.80)}} & \textbf{0.85 ($\:\:$\textcolor[HTML]{1400AD}{-4.77)}} & \textbf{0.85 ($\:\:$\textcolor[HTML]{1400AD}{-4.74)}} & 87 \\ 
& & Qwen-VL & 7.08 ($\:\:$\textcolor[HTML]{1400AD}{-7.52}) & 12.73 ($\:\:$\textcolor[HTML]{1400AD}{-8.77}) & 8.26 ($\:\:$\textcolor[HTML]{1400AD}{-7.13}) & 0.34 ($\:\:$\textcolor[HTML]{1400AD}{-1.02}) & 0.38 ($\:\:$\textcolor[HTML]{1400AD}{-0.95}) & 0.38 ($\:\:$\textcolor[HTML]{1400AD}{-0.90}) & 0.38 ($\:\:$\textcolor[HTML]{1400AD}{-0.88}) & 112 \\ 
& & Qwen-VL (FT) & 6.08 (\textcolor[HTML]{1400AD}{-14.01}) & 10.10 (\textcolor[HTML]{1400AD}{-16.17}) & 7.39 (\textcolor[HTML]{1400AD}{-13.44}) & 0.49 ($\:\:$\textcolor[HTML]{1400AD}{-5.17}) & 0.58 ($\:\:$\textcolor[HTML]{1400AD}{-5.20}) & 0.59 ($\:\:$\textcolor[HTML]{1400AD}{-5.19}) & 0.59 ($\:\:$\textcolor[HTML]{1400AD}{-5.13}) & 187 \\ 
& & mPLUG-Owl2 & 6.54 ($\:\:$\textcolor[HTML]{1400AD}{-7.86}) & 12.20 ($\:\:$\textcolor[HTML]{1400AD}{-9.76}) & 7.44 ($\:\:$\textcolor[HTML]{1400AD}{-8.27}) & 0.42 ($\:\:$\textcolor[HTML]{1400AD}{-0.85}) & 0.40 ($\:\:$\textcolor[HTML]{1400AD}{-0.77}) & 0.39 ($\:\:$\textcolor[HTML]{1400AD}{-0.75}) & 0.39 ($\:\:$\textcolor[HTML]{1400AD}{-0.71}) & 102 \\ 

\midrule
\multirow{4.5}{*}{En} & \multirow{4.5}{*}{Nl}  & LLaVA-NeXT & \textbf{7.91 ($\:\:$\textcolor[HTML]{1400AD}{-9.76)}} & \textbf{13.25 (\textcolor[HTML]{1400AD}{-12.80)}} & 8.63 ($\:\:$\textcolor[HTML]{1400AD}{-9.92}) & 0.31 ($\:\:$\textcolor[HTML]{1400AD}{-1.01}) & 0.44 ($\:\:$\textcolor[HTML]{1400AD}{-0.79}) & 0.42 ($\:\:$\textcolor[HTML]{1400AD}{-0.78}) & 0.43 ($\:\:$\textcolor[HTML]{1400AD}{-0.77}) & 175 \\ 
& & LLaVA-NeXT (FT) & 7.89 (\textcolor[HTML]{1400AD}{-10.03}) & 11.66 (\textcolor[HTML]{1400AD}{-12.00}) & \textbf{8.81 (\textcolor[HTML]{1400AD}{-10.39)}} & \textbf{1.22 ($\:\:$\textcolor[HTML]{1400AD}{-4.44)}} & \textbf{1.13 ($\:\:$\textcolor[HTML]{1400AD}{-4.53)}} & \textbf{1.12 ($\:\:$\textcolor[HTML]{1400AD}{-4.51)}} & \textbf{1.11 ($\:\:$\textcolor[HTML]{1400AD}{-4.49)}} & 102 \\ 
& & Qwen-VL & 7.41 ($\:\:$\textcolor[HTML]{1400AD}{-7.19}) & 12.33 ($\:\:$\textcolor[HTML]{1400AD}{-9.18}) & 7.93 ($\:\:$\textcolor[HTML]{1400AD}{-7.46}) & 0.35 ($\:\:$\textcolor[HTML]{1400AD}{-1.01}) & 0.49 ($\:\:$\textcolor[HTML]{1400AD}{-0.84}) & 0.50 ($\:\:$\textcolor[HTML]{1400AD}{-0.78}) & 0.53 ($\:\:$\textcolor[HTML]{1400AD}{-0.74}) & 137 \\ 
& & Qwen-VL (FT) & 6.67 (\textcolor[HTML]{1400AD}{-13.42}) & 10.07 (\textcolor[HTML]{1400AD}{-16.21}) & 7.67 (\textcolor[HTML]{1400AD}{-13.16}) & 0.68 ($\:\:$\textcolor[HTML]{1400AD}{-4.98}) & 0.73 ($\:\:$\textcolor[HTML]{1400AD}{-5.05}) & 0.70 ($\:\:$\textcolor[HTML]{1400AD}{-5.08}) & 0.71 ($\:\:$\textcolor[HTML]{1400AD}{-5.01}) & 166 \\ 
& & mPLUG-Owl2 & 4.61 ($\:\:$\textcolor[HTML]{1400AD}{-9.80}) & 8.96 (\textcolor[HTML]{1400AD}{-13.00}) & 4.84 (\textcolor[HTML]{1400AD}{-10.87}) & 0.20 ($\:\:$\textcolor[HTML]{1400AD}{-1.06}) & 0.26 ($\:\:$\textcolor[HTML]{1400AD}{-0.91}) & 0.25 ($\:\:$\textcolor[HTML]{1400AD}{-0.89}) & 0.25 ($\:\:$\textcolor[HTML]{1400AD}{-0.85}) & 106 \\ 

\midrule
\multirow{4.5}{*}{En} & \multirow{4.5}{*}{Sv}  & LLaVA-NeXT & \textbf{13.08 ($\:\:$\textcolor[HTML]{1400AD}{-4.59)}} & \textbf{17.19 ($\:\:$\textcolor[HTML]{1400AD}{-8.85)}} & \textbf{12.38 ($\:\:$\textcolor[HTML]{1400AD}{-6.18)}} & \textbf{0.89 ($\:\:$\textcolor[HTML]{1400AD}{-0.42)}} & \textbf{0.82 ($\:\:$\textcolor[HTML]{1400AD}{-0.41)}} & \textbf{0.82 ($\:\:$\textcolor[HTML]{1400AD}{-0.39)}} & \textbf{0.75 ($\:\:$\textcolor[HTML]{1400AD}{-0.44)}} & 172 \\ 
& & LLaVA-NeXT (FT) & 9.44 ($\:\:$\textcolor[HTML]{1400AD}{-8.47}) & 12.79 (\textcolor[HTML]{1400AD}{-10.87}) & 9.62 ($\:\:$\textcolor[HTML]{1400AD}{-9.58}) & 0.73 ($\:\:$\textcolor[HTML]{1400AD}{-4.94}) & 0.64 ($\:\:$\textcolor[HTML]{1400AD}{-5.03}) & 0.60 ($\:\:$\textcolor[HTML]{1400AD}{-5.03}) & 0.58 ($\:\:$\textcolor[HTML]{1400AD}{-5.01}) & 94 \\ 
& & Qwen-VL & 10.59 ($\:\:$\textcolor[HTML]{1400AD}{-4.01}) & 14.72 ($\:\:$\textcolor[HTML]{1400AD}{-6.79}) & 10.75 ($\:\:$\textcolor[HTML]{1400AD}{-4.64}) & 0.58 ($\:\:$\textcolor[HTML]{1400AD}{-0.78}) & 0.61 ($\:\:$\textcolor[HTML]{1400AD}{-0.72}) & 0.66 ($\:\:$\textcolor[HTML]{1400AD}{-0.62}) & 0.61 ($\:\:$\textcolor[HTML]{1400AD}{-0.66}) & 124 \\ 
& & Qwen-VL (FT) & 9.47 (\textcolor[HTML]{1400AD}{-10.62}) & 13.20 (\textcolor[HTML]{1400AD}{-13.07}) & 9.93 (\textcolor[HTML]{1400AD}{-10.91}) & 0.72 ($\:\:$\textcolor[HTML]{1400AD}{-4.95}) & 0.65 ($\:\:$\textcolor[HTML]{1400AD}{-5.13}) & 0.63 ($\:\:$\textcolor[HTML]{1400AD}{-5.14}) & 0.58 ($\:\:$\textcolor[HTML]{1400AD}{-5.14}) & 155 \\ 
& & mPLUG-Owl2 & 9.37 ($\:\:$\textcolor[HTML]{1400AD}{-5.03}) & 12.82 ($\:\:$\textcolor[HTML]{1400AD}{-9.14}) & 8.53 ($\:\:$\textcolor[HTML]{1400AD}{-7.17}) & 0.40 ($\:\:$\textcolor[HTML]{1400AD}{-0.86}) & 0.36 ($\:\:$\textcolor[HTML]{1400AD}{-0.81}) & 0.36 ($\:\:$\textcolor[HTML]{1400AD}{-0.78}) & 0.33 ($\:\:$\textcolor[HTML]{1400AD}{-0.77}) & 79 \\ 

\midrule
\multirow{4.5}{*}{En} & \multirow{4.5}{*}{Ru}  & LLaVA-NeXT & \textbf{7.86 ($\:\:$\textcolor[HTML]{1400AD}{-9.81)}} & \textbf{10.75 (\textcolor[HTML]{1400AD}{-15.29)}} & \textbf{6.39 (\textcolor[HTML]{1400AD}{-12.16)}} & \textbf{0.22 ($\:\:$\textcolor[HTML]{1400AD}{-1.09)}} & \textbf{0.26 ($\:\:$\textcolor[HTML]{1400AD}{-0.97)}} & \textbf{0.28 ($\:\:$\textcolor[HTML]{1400AD}{-0.92)}} & \textbf{0.28 ($\:\:$\textcolor[HTML]{1400AD}{-0.92)}} & 203 \\ 
& & LLaVA-NeXT (FT) & 0.42 (\textcolor[HTML]{1400AD}{-17.50}) & 1.51 (\textcolor[HTML]{1400AD}{-22.14}) & 0.31 (\textcolor[HTML]{1400AD}{-18.89}) & 0.01 ($\:\:$\textcolor[HTML]{1400AD}{-5.66}) & 0.01 ($\:\:$\textcolor[HTML]{1400AD}{-5.65}) & 0.01 ($\:\:$\textcolor[HTML]{1400AD}{-5.62}) & 0.01 ($\:\:$\textcolor[HTML]{1400AD}{-5.58}) & 72 \\ 
& & Qwen-VL & 3.05 (\textcolor[HTML]{1400AD}{-11.55}) & 4.81 (\textcolor[HTML]{1400AD}{-16.69}) & 2.35 (\textcolor[HTML]{1400AD}{-13.04}) & 0.05 ($\:\:$\textcolor[HTML]{1400AD}{-1.31}) & 0.07 ($\:\:$\textcolor[HTML]{1400AD}{-1.26}) & 0.08 ($\:\:$\textcolor[HTML]{1400AD}{-1.20}) & 0.08 ($\:\:$\textcolor[HTML]{1400AD}{-1.18}) & 112 \\ 
& & Qwen-VL (FT) & 0.15 (\textcolor[HTML]{1400AD}{-19.94}) & 1.09 (\textcolor[HTML]{1400AD}{-25.19}) & 0.09 (\textcolor[HTML]{1400AD}{-20.74}) & 0.00 ($\:\:$\textcolor[HTML]{1400AD}{-5.67}) & 0.00 ($\:\:$\textcolor[HTML]{1400AD}{-5.78}) & 0.00 ($\:\:$\textcolor[HTML]{1400AD}{-5.77}) & 0.00 ($\:\:$\textcolor[HTML]{1400AD}{-5.72}) & 203 \\ 
& & mPLUG-Owl2 & 3.69 (\textcolor[HTML]{1400AD}{-10.71}) & 5.33 (\textcolor[HTML]{1400AD}{-16.64}) & 2.83 (\textcolor[HTML]{1400AD}{-12.88}) & 0.11 ($\:\:$\textcolor[HTML]{1400AD}{-1.16}) & 0.10 ($\:\:$\textcolor[HTML]{1400AD}{-1.07}) & 0.09 ($\:\:$\textcolor[HTML]{1400AD}{-1.05}) & 0.10 ($\:\:$\textcolor[HTML]{1400AD}{-1.00}) & 107 \\ 

\midrule
\multirow{4.5}{*}{En} & \multirow{4.5}{*}{Ja}  & LLaVA-NeXT & \textbf{8.65 ($\:\:$\textcolor[HTML]{1400AD}{-9.01)}} & \textbf{8.70 (\textcolor[HTML]{1400AD}{-17.35)}} & \textbf{12.34 ($\:\:$\textcolor[HTML]{1400AD}{-6.21)}} & \textbf{0.44 ($\:\:$\textcolor[HTML]{1400AD}{-0.87)}} & \textbf{0.44 ($\:\:$\textcolor[HTML]{1400AD}{-0.79)}} & \textbf{0.44 ($\:\:$\textcolor[HTML]{1400AD}{-0.76)}} & \textbf{0.44 ($\:\:$\textcolor[HTML]{1400AD}{-0.76)}} & 213 \\ 
& & LLaVA-NeXT (FT) & 0.46 (\textcolor[HTML]{1400AD}{-17.45}) & 0.61 (\textcolor[HTML]{1400AD}{-23.04}) & 0.51 (\textcolor[HTML]{1400AD}{-18.69}) & 0.02 ($\:\:$\textcolor[HTML]{1400AD}{-5.65}) & 0.02 ($\:\:$\textcolor[HTML]{1400AD}{-5.65}) & 0.02 ($\:\:$\textcolor[HTML]{1400AD}{-5.61}) & 0.02 ($\:\:$\textcolor[HTML]{1400AD}{-5.58}) & 67 \\ 
& & Qwen-VL & 3.10 (\textcolor[HTML]{1400AD}{-11.50}) & 3.16 (\textcolor[HTML]{1400AD}{-18.35}) & 4.37 (\textcolor[HTML]{1400AD}{-11.02}) & 0.12 ($\:\:$\textcolor[HTML]{1400AD}{-1.24}) & 0.12 ($\:\:$\textcolor[HTML]{1400AD}{-1.21}) & 0.12 ($\:\:$\textcolor[HTML]{1400AD}{-1.16}) & 0.12 ($\:\:$\textcolor[HTML]{1400AD}{-1.14}) & 127 \\ 
& & Qwen-VL (FT) & 0.21 (\textcolor[HTML]{1400AD}{-19.88}) & 0.46 (\textcolor[HTML]{1400AD}{-25.82}) & 0.12 (\textcolor[HTML]{1400AD}{-20.72}) & 0.00 ($\:\:$\textcolor[HTML]{1400AD}{-5.67}) & 0.00 ($\:\:$\textcolor[HTML]{1400AD}{-5.78}) & 0.00 ($\:\:$\textcolor[HTML]{1400AD}{-5.77}) & 0.00 ($\:\:$\textcolor[HTML]{1400AD}{-5.72}) & 152 \\ 
& & mPLUG-Owl2 & 4.00 (\textcolor[HTML]{1400AD}{-10.40}) & 4.06 (\textcolor[HTML]{1400AD}{-17.90}) & 5.39 (\textcolor[HTML]{1400AD}{-10.32}) & 0.25 ($\:\:$\textcolor[HTML]{1400AD}{-1.01}) & 0.25 ($\:\:$\textcolor[HTML]{1400AD}{-0.92}) & 0.25 ($\:\:$\textcolor[HTML]{1400AD}{-0.89}) & 0.25 ($\:\:$\textcolor[HTML]{1400AD}{-0.85}) & 104 \\ 

\midrule
\multirow{4.5}{*}{En} & \multirow{4.5}{*}{Zh}  & LLaVA-NeXT & \textbf{10.81 ($\:\:$\textcolor[HTML]{1400AD}{-6.86)}} & \textbf{10.90 (\textcolor[HTML]{1400AD}{-15.15)}} & \textbf{13.00 ($\:\:$\textcolor[HTML]{1400AD}{-5.56)}} & 0.60 ($\:\:$\textcolor[HTML]{1400AD}{-0.71}) & 0.60 ($\:\:$\textcolor[HTML]{1400AD}{-0.62}) & 0.60 ($\:\:$\textcolor[HTML]{1400AD}{-0.60}) & 0.60 ($\:\:$\textcolor[HTML]{1400AD}{-0.59}) & 220 \\ 
& & LLaVA-NeXT (FT) & 0.64 (\textcolor[HTML]{1400AD}{-17.27}) & 0.89 (\textcolor[HTML]{1400AD}{-22.76}) & 0.75 (\textcolor[HTML]{1400AD}{-18.45}) & 0.08 ($\:\:$\textcolor[HTML]{1400AD}{-5.59}) & 0.08 ($\:\:$\textcolor[HTML]{1400AD}{-5.59}) & 0.08 ($\:\:$\textcolor[HTML]{1400AD}{-5.55}) & 0.08 ($\:\:$\textcolor[HTML]{1400AD}{-5.52}) & 71 \\ 
& & Qwen-VL & 8.60 ($\:\:$\textcolor[HTML]{1400AD}{-6.00}) & 8.65 (\textcolor[HTML]{1400AD}{-12.85}) & 10.34 ($\:\:$\textcolor[HTML]{1400AD}{-5.05}) & \textbf{0.80 ($\:\:$\textcolor[HTML]{1400AD}{-0.56)}} & \textbf{0.79 ($\:\:$\textcolor[HTML]{1400AD}{-0.54)}} & \textbf{0.79 ($\:\:$\textcolor[HTML]{1400AD}{-0.49)}} & \textbf{0.79 ($\:\:$\textcolor[HTML]{1400AD}{-0.47)}} & 133 \\ 
& & Qwen-VL (FT) & 0.35 (\textcolor[HTML]{1400AD}{-19.74}) & 0.64 (\textcolor[HTML]{1400AD}{-25.63}) & 0.27 (\textcolor[HTML]{1400AD}{-20.57}) & 0.01 ($\:\:$\textcolor[HTML]{1400AD}{-5.66}) & 0.01 ($\:\:$\textcolor[HTML]{1400AD}{-5.77}) & 0.01 ($\:\:$\textcolor[HTML]{1400AD}{-5.77}) & 0.01 ($\:\:$\textcolor[HTML]{1400AD}{-5.71}) & 155 \\ 
& & mPLUG-Owl2 & 4.99 ($\:\:$\textcolor[HTML]{1400AD}{-9.42}) & 5.04 (\textcolor[HTML]{1400AD}{-16.92}) & 6.08 ($\:\:$\textcolor[HTML]{1400AD}{-9.62}) & 0.52 ($\:\:$\textcolor[HTML]{1400AD}{-0.75}) & 0.52 ($\:\:$\textcolor[HTML]{1400AD}{-0.65}) & 0.52 ($\:\:$\textcolor[HTML]{1400AD}{-0.63}) & 0.52 ($\:\:$\textcolor[HTML]{1400AD}{-0.58}) & 107 \\ 

\bottomrule
\end{tabular}
}
\caption{Results of LVLMs in Full Task. Bold fonts indicate the best score for that language combination. This result shows that no matter how much the amount of data is increased, the best performance is achieved by having instructions given and output in English. The values are noted next to the output of the difference by the same model in the method with instruction and output in English (\{En\}-\{En\}). Red indicates a higher value than that method; blue indicates a lower value.}

\label{tab:result-score-full}
\end{table*}

%% file: results/a5.tex
\begin{table*}[h!]
\centering
\resizebox{0.86\textwidth}{!}{
\footnotesize
\setlength{\tabcolsep}{5pt}
\begin{tabular}{@{}cccrrrrrrrc@{}}
\toprule 
\multicolumn{1}{c}{\multirow{2.5}{*}{Input}} & \multirow{2.5}{*}{Output} & \multirow{2.5}{*}{LVLM} & \multicolumn{2}{c}{Entity Cov.} & \multicolumn{1}{c}{\multirow{2.5}{*}{Entity F1}} & \multicolumn{4}{c}{Entity Cooccurrence} & \multirow{2.5}{*}{Avg. Length} \\
\cmidrule(lr){4-5} \cmidrule(lr){7-10}
& & & \multicolumn{1}{c}{exact} & \multicolumn{1}{c}{partial} & & \multicolumn{1}{c}{n=0} & \multicolumn{1}{c}{n=1} & \multicolumn{1}{c}{n=2} & \multicolumn{1}{c}{n=$\infty$} & \\
\midrule

\multirow{6.5}{*}{En} & \multirow{6.5}{*}{En} & LLaVA-NeXT & \textbf{22.58} & \textbf{30.30} & \textbf{25.19} & 3.15 & 2.30 & 2.40 & 2.44 & 259 \\ 
& & LLaVA-NeXT (FT) & 13.80 & 18.73 & 14.57 & 3.68 & 3.55 & 3.50 & 3.42 & 86 \\ 
& & Qwen-VL & 19.74 & 26.57 & 20.79 & 2.14 & 2.18 & 2.20 & 1.99 & 178 \\ 
& & Qwen-VL (FT) & 19.23 & 25.31 & 19.16 & \textbf{5.70} & \textbf{4.93} & \textbf{4.88} & \textbf{4.78} & 237 \\ 
& & mPLUG-Owl2 & 18.76 & 26.05 & 20.90 & 2.74 & 1.99 & 1.93 & 1.83 & 158 \\ 
& & Phi-3 & 12.12 & 18.55 & 12.05 & 1.68 & 0.97 & 1.01 & 1.00 & 133 \\
\bottomrule
\end{tabular}
}
\caption{Results of LVLMs in Alignment-5 Task. Bold fonts indicate the best score for that language combination. The values are noted next to the output of the difference by the same model in the method with instruction and output in English (\{En\}-\{En\}). Red indicates a higher value than that method; blue indicates a lower value.}
\label{tab:result-score-5}
\end{table*}

%% file: results/a10-appendix.tex
\begin{table*}[h!]
\centering
\resizebox{\textwidth}{!}{
\footnotesize
\setlength{\tabcolsep}{5pt}
\begin{tabular}{@{}cccrrrrrrrc@{}}
\toprule 
\multicolumn{1}{c}{\multirow{2.5}{*}{Input}} & \multirow{2.5}{*}{Output} & \multirow{2.5}{*}{LVLM} & \multicolumn{2}{c}{Entity Cov.} & \multicolumn{1}{c}{\multirow{2.5}{*}{Entity F1}} & \multicolumn{4}{c}{Entity Cooccurrence} & \multirow{2.5}{*}{Avg. Length} \\
\cmidrule(lr){4-5} \cmidrule(lr){7-10}
& & & \multicolumn{1}{c}{exact} & \multicolumn{1}{c}{partial} & & \multicolumn{1}{c}{n=0} & \multicolumn{1}{c}{n=1} & \multicolumn{1}{c}{n=2} & \multicolumn{1}{c}{n=$\infty$} & \\
\midrule
\multirow{2}{*}{En} & \multirow{2}{*}{En} & Phi-3 &  9.74 & 13.75 & 8.06 & 0.95 & 0.86 & 0.92 & 0.90 & 108 \\ 
& & XComposer2 &  \textbf{16.57} & \textbf{21.56} & \textbf{16.53} & \textbf{1.51} & \textbf{1.47} & \textbf{1.43} & \textbf{1.36} & 223 \\ 

\midrule
\multirow{2}{*}{En} & \multirow{2}{*}{Es} & Phi-3 &  \textbf{10.35 (\textcolor[HTML]{AD0000}{+0.61)}} & \textbf{14.46 (\textcolor[HTML]{AD0000}{+0.72)}} & \textbf{11.67 (\textcolor[HTML]{AD0000}{+3.61)}} & \textbf{1.20 (\textcolor[HTML]{AD0000}{+0.25)}} & \textbf{1.37 (\textcolor[HTML]{AD0000}{+0.51)}} & \textbf{1.43 (\textcolor[HTML]{AD0000}{+0.51)}} & \textbf{1.38 (\textcolor[HTML]{AD0000}{+0.48)}} & 141 \\ 
& & XComposer2 &  10.03 (\textcolor[HTML]{1400AD}{-6.55}) & 13.75 (\textcolor[HTML]{1400AD}{-7.81}) & 10.04 (\textcolor[HTML]{1400AD}{-6.49}) & 0.79 (\textcolor[HTML]{1400AD}{-0.73}) & 0.96 (\textcolor[HTML]{1400AD}{-0.51}) & 0.92 (\textcolor[HTML]{1400AD}{-0.50}) & 0.89 (\textcolor[HTML]{1400AD}{-0.47}) & 116 \\ 

\midrule
\multirow{2}{*}{Es} & \multirow{2}{*}{En} & Phi-3 &  7.99 (\textcolor[HTML]{1400AD}{-1.75}) & 14.53 (\textcolor[HTML]{AD0000}{+0.78}) & 7.35 (\textcolor[HTML]{1400AD}{-0.71}) & \textbf{0.90 (\textcolor[HTML]{1400AD}{-0.05)}} & \textbf{0.79 (\textcolor[HTML]{1400AD}{-0.07)}} & \textbf{0.80 (\textcolor[HTML]{1400AD}{-0.12)}} & \textbf{0.75 (\textcolor[HTML]{1400AD}{-0.16)}} & 91 \\
\bottomrule
\end{tabular}
}
\caption{Results for Phi-3 and XComposer2 in the Alignment-10 task. Bold fonts indicate the best score for that language combination. The values are noted next to the output of the difference by the same model in the method with instruction and output in English (\{En\}-\{En\}). Red indicates a higher value than that method; blue indicates a lower value.}
\label{tab:result-score-10-appendix}
\end{table*}

%% file: results/a10-lang-en.tex
\begin{table*}[h!]
\centering
\resizebox{\textwidth}{!}{
\footnotesize
\setlength{\tabcolsep}{5pt}
\begin{tabular}{@{}cccrrrrrrrc@{}}
\toprule 
\multicolumn{1}{c}{\multirow{2.5}{*}{Input}} & \multirow{2.5}{*}{Output} & \multirow{2.5}{*}{LVLM} & \multicolumn{2}{c}{Entity Cov.} & \multirow{2.5}{*}{Entity F1} & \multicolumn{4}{c}{Entity Cooccurrence} & \multirow{2.5}{*}{Avg. Length} \\
\cmidrule(lr){4-5} \cmidrule(lr){7-10}
& & & \multicolumn{1}{c}{exact} & \multicolumn{1}{c}{partial} & & \multicolumn{1}{c}{n=0} & \multicolumn{1}{c}{n=1} & \multicolumn{1}{c}{n=2} & \multicolumn{1}{c}{n=$\infty$} & \\
\midrule
\multirow{4.5}{*}{En} & \multirow{4.5}{*}{En} & LLaVA-NeXT &  \textbf{26.49} & \textbf{31.54} & \textbf{26.07} & 1.35 & 1.65 & 1.66 & 1.70 & 252 \\ 
& & LLaVA-NeXT (FT) &  16.98 & 22.70 & 19.95 & 3.02 & 3.33 & 3.23 & 3.11 & 83 \\ 
& & Qwen-VL &  21.11 & 27.03 & 21.78 & 1.60 & 1.59 & 1.56 & 1.52 & 155 \\ 
& & Qwen-VL (FT) &  21.12 & 24.87 & 21.95 & \textbf{3.57} & \textbf{3.83} & \textbf{3.78} & \textbf{3.68} & 177 \\ 
& & mPLUG-Owl2 &  12.79 & 17.08 & 13.48 & 2.07 & 1.68 & 1.59 & 1.56 & 151 \\ 
\midrule
\multirow{4.5}{*}{Es} & \multirow{4.5}{*}{En} & LLaVA-NeXT & 4.22 (\textcolor[HTML]{1400AD}{-22.26}) & 6.22 (\textcolor[HTML]{1400AD}{-25.33}) & 4.17 (\textcolor[HTML]{1400AD}{-21.90}) & 0.00 (\textcolor[HTML]{1400AD}{-1.34}) & 0.00 (\textcolor[HTML]{1400AD}{-1.65}) & 0.00 (\textcolor[HTML]{1400AD}{-1.66}) & 0.00 (\textcolor[HTML]{1400AD}{-1.70}) & 180 \\
\bottomrule
\end{tabular}
}
\caption{Results of LVLMs in Alignment-10 Task, which the method is an instruction in languages other than English and output in English (\{Lang\}-\{En\}). Bold fonts indicate the best score for that language combination. The values are noted next to the output of the difference by the same model in the method with instruction and output in English (\{En\}-\{En\}). Red indicates a higher value than that method; blue indicates a lower value.}
\label{tab:result-score-10-lang-en}
\end{table*}

%% file: results/full-misc.tex
\begin{table*}[h!]
\centering
\resizebox{0.78\textwidth}{!}{
\footnotesize
\setlength{\tabcolsep}{6pt}
\begin{tabular}{@{}cccrrrrr@{}} 
\toprule
\multicolumn{1}{c}{\multirow{2}{*}{Input}} & 
\multirow{2}{*}{Output} & 
\multirow{2}{*}{LVLM} & 
\multicolumn{1}{c}{\multirow{2}{*}{BLEU}} & 
\multicolumn{3}{c}{ROUGE} & 
\multicolumn{1}{c}{\multirow{2}{*}{BertScore}} \\
\cmidrule(lr){5-7}
& & & & 1 & 2 & L\\
\midrule
\multirow{6.5}{*}{En} & \multirow{6.5}{*}{En} & LLaVA-NeXT & 0.01 & 0.24 & 0.05 & 0.15 & 0.82 \\ 
& &  LLaVA-NeXT (FT) & \textbf{0.07} & 0.28 & \textbf{0.13} & \textbf{0.22} & \textbf{0.85} \\ 
& &  Qwen-VL & 0.01 & 0.22 & 0.05 & 0.14 & 0.82 \\ 
& &  Qwen-VL (FT) & 0.06 & \textbf{0.28} & 0.12 & 0.22 & 0.84 \\ 
& &  mPLUG-Owl2 & 0.01 & 0.24 & 0.05 & 0.15 & 0.82 \\ 
& &  Phi-3 & 0.01 & 0.20 & 0.04 & 0.12 & 0.82 \\ 
& &  XComposer2 & 0.01 & 0.24 & 0.05 & 0.14 & 0.82 \\ 

\midrule
\multirow{6.5}{*}{En} & \multirow{6.5}{*}{Es} & LLaVA-NeXT & \textbf{0.01 (\textcolor[HTML]{1400AD}{-0.00)}} & \textbf{0.28 (\textcolor[HTML]{AD0000}{+0.04)}} & \textbf{0.06 (\textcolor[HTML]{AD0000}{+0.01)}} & \textbf{0.16 (\textcolor[HTML]{AD0000}{+0.01)}} & \textbf{0.81 (\textcolor[HTML]{1400AD}{-0.01)}} \\ 
& &  LLaVA-NeXT (FT) & 0.00 (\textcolor[HTML]{1400AD}{-0.07}) & 0.05 (\textcolor[HTML]{1400AD}{-0.23}) & 0.01 (\textcolor[HTML]{1400AD}{-0.12}) & 0.04 (\textcolor[HTML]{1400AD}{-0.18}) & 0.78 (\textcolor[HTML]{1400AD}{-0.07}) \\ 
& &  Qwen-VL & 0.00 (\textcolor[HTML]{1400AD}{-0.01}) & 0.20 (\textcolor[HTML]{1400AD}{-0.03}) & 0.04 (\textcolor[HTML]{1400AD}{-0.01}) & 0.12 (\textcolor[HTML]{1400AD}{-0.02}) & 0.80 (\textcolor[HTML]{1400AD}{-0.02}) \\ 
& &  Qwen-VL (FT) & 0.00 (\textcolor[HTML]{1400AD}{-0.06}) & 0.03 (\textcolor[HTML]{1400AD}{-0.25}) & 0.00 (\textcolor[HTML]{1400AD}{-0.11}) & 0.03 (\textcolor[HTML]{1400AD}{-0.19}) & 0.77 (\textcolor[HTML]{1400AD}{-0.07}) \\ 
& &  mPLUG-Owl2 & 0.00 (\textcolor[HTML]{1400AD}{-0.01}) & 0.22 (\textcolor[HTML]{1400AD}{-0.03}) & 0.04 (\textcolor[HTML]{1400AD}{-0.01}) & 0.13 (\textcolor[HTML]{1400AD}{-0.02}) & 0.80 (\textcolor[HTML]{1400AD}{-0.02}) \\ 
& &  Phi-3 & 0.00 (\textcolor[HTML]{1400AD}{-0.00}) & 0.21 (\textcolor[HTML]{AD0000}{+0.01}) & 0.04 (\textcolor[HTML]{AD0000}{+0.00}) & 0.13 (\textcolor[HTML]{AD0000}{+0.00}) & 0.79 (\textcolor[HTML]{1400AD}{-0.02}) \\ 
& &  XComposer2 & 0.00 (\textcolor[HTML]{1400AD}{-0.01}) & 0.18 (\textcolor[HTML]{1400AD}{-0.06}) & 0.04 (\textcolor[HTML]{1400AD}{-0.02}) & 0.11 (\textcolor[HTML]{1400AD}{-0.03}) & 0.80 (\textcolor[HTML]{1400AD}{-0.02}) \\ 

\midrule
\multirow{6.5}{*}{En} & \multirow{6.5}{*}{Fr} &  LLaVA-NeXT & \textbf{0.00 (\textcolor[HTML]{1400AD}{-0.01)}} & \textbf{0.20 (\textcolor[HTML]{1400AD}{-0.04)}} & \textbf{0.04 (\textcolor[HTML]{1400AD}{-0.02)}} & \textbf{0.12 (\textcolor[HTML]{1400AD}{-0.03)}} & \textbf{0.79 (\textcolor[HTML]{1400AD}{-0.02)}} \\ 
& &  LLaVA-NeXT (FT) & 0.00 (\textcolor[HTML]{1400AD}{-0.07}) & 0.06 (\textcolor[HTML]{1400AD}{-0.21}) & 0.02 (\textcolor[HTML]{1400AD}{-0.11}) & 0.05 (\textcolor[HTML]{1400AD}{-0.17}) & 0.78 (\textcolor[HTML]{1400AD}{-0.06}) \\ 
& &  Qwen-VL & 0.00 (\textcolor[HTML]{1400AD}{-0.01}) & 0.15 (\textcolor[HTML]{1400AD}{-0.08}) & 0.03 (\textcolor[HTML]{1400AD}{-0.02}) & 0.09 (\textcolor[HTML]{1400AD}{-0.05}) & 0.79 (\textcolor[HTML]{1400AD}{-0.03}) \\ 
& &  Qwen-VL (FT) & 0.00 (\textcolor[HTML]{1400AD}{-0.06}) & 0.03 (\textcolor[HTML]{1400AD}{-0.25}) & 0.00 (\textcolor[HTML]{1400AD}{-0.11}) & 0.03 (\textcolor[HTML]{1400AD}{-0.19}) & 0.77 (\textcolor[HTML]{1400AD}{-0.07}) \\ 
& &  mPLUG-Owl2 & 0.00 (\textcolor[HTML]{1400AD}{-0.01}) & 0.16 (\textcolor[HTML]{1400AD}{-0.08}) & 0.03 (\textcolor[HTML]{1400AD}{-0.02}) & 0.10 (\textcolor[HTML]{1400AD}{-0.05}) & 0.79 (\textcolor[HTML]{1400AD}{-0.03}) \\ 
& &  Phi-3 & 0.00 (\textcolor[HTML]{1400AD}{-0.00}) & 0.15 (\textcolor[HTML]{1400AD}{-0.04}) & 0.02 (\textcolor[HTML]{1400AD}{-0.01}) & 0.09 (\textcolor[HTML]{1400AD}{-0.03}) & 0.78 (\textcolor[HTML]{1400AD}{-0.03}) \\ 
& &  XComposer2 & 0.00 (\textcolor[HTML]{1400AD}{-0.01}) & 0.03 (\textcolor[HTML]{1400AD}{-0.21}) & 0.01 (\textcolor[HTML]{1400AD}{-0.05}) & 0.03 (\textcolor[HTML]{1400AD}{-0.12}) & 0.78 (\textcolor[HTML]{1400AD}{-0.04}) \\ 

\midrule
\multirow{6.5}{*}{En} & \multirow{6.5}{*}{De} & LLaVA-NeXT & \textbf{0.00 (\textcolor[HTML]{1400AD}{-0.01)}} & \textbf{0.20 (\textcolor[HTML]{1400AD}{-0.05)}} & \textbf{0.03 (\textcolor[HTML]{1400AD}{-0.02)}} & \textbf{0.11 (\textcolor[HTML]{1400AD}{-0.03)}} & \textbf{0.80 (\textcolor[HTML]{1400AD}{-0.02)}} \\ 
& &  LLaVA-NeXT (FT) & 0.00 (\textcolor[HTML]{1400AD}{-0.07}) & 0.04 (\textcolor[HTML]{1400AD}{-0.23}) & 0.01 (\textcolor[HTML]{1400AD}{-0.12}) & 0.03 (\textcolor[HTML]{1400AD}{-0.19}) & 0.76 (\textcolor[HTML]{1400AD}{-0.08}) \\ 
& &  Qwen-VL & 0.00 (\textcolor[HTML]{1400AD}{-0.01}) & 0.14 (\textcolor[HTML]{1400AD}{-0.08}) & 0.02 (\textcolor[HTML]{1400AD}{-0.03}) & 0.09 (\textcolor[HTML]{1400AD}{-0.06}) & 0.79 (\textcolor[HTML]{1400AD}{-0.03}) \\ 
& &  Qwen-VL (FT) & 0.00 (\textcolor[HTML]{1400AD}{-0.06}) & 0.03 (\textcolor[HTML]{1400AD}{-0.25}) & 0.00 (\textcolor[HTML]{1400AD}{-0.11}) & 0.03 (\textcolor[HTML]{1400AD}{-0.19}) & 0.76 (\textcolor[HTML]{1400AD}{-0.08}) \\ 
& &  mPLUG-Owl2 & 0.00 (\textcolor[HTML]{1400AD}{-0.01}) & 0.14 (\textcolor[HTML]{1400AD}{-0.10}) & 0.02 (\textcolor[HTML]{1400AD}{-0.03}) & 0.09 (\textcolor[HTML]{1400AD}{-0.07}) & 0.79 (\textcolor[HTML]{1400AD}{-0.03}) \\ 
& &  Phi-3 & 0.00 (\textcolor[HTML]{1400AD}{-0.00}) & 0.14 (\textcolor[HTML]{1400AD}{-0.05}) & 0.02 (\textcolor[HTML]{1400AD}{-0.02}) & 0.09 (\textcolor[HTML]{1400AD}{-0.03}) & 0.78 (\textcolor[HTML]{1400AD}{-0.03}) \\ 
& &  XComposer2 & 0.00 (\textcolor[HTML]{1400AD}{-0.01}) & 0.14 (\textcolor[HTML]{1400AD}{-0.10}) & 0.02 (\textcolor[HTML]{1400AD}{-0.03}) & 0.09 (\textcolor[HTML]{1400AD}{-0.06}) & 0.79 (\textcolor[HTML]{1400AD}{-0.03}) \\ 

\midrule
\multirow{6.5}{*}{En} & \multirow{6.5}{*}{It} & LLaVA-NeXT & \textbf{0.00 (\textcolor[HTML]{1400AD}{-0.01)}} & \textbf{0.19 (\textcolor[HTML]{1400AD}{-0.05)}} & \textbf{0.02 (\textcolor[HTML]{1400AD}{-0.03)}} & \textbf{0.11 (\textcolor[HTML]{1400AD}{-0.04)}} & \textbf{0.80 (\textcolor[HTML]{1400AD}{-0.01)}} \\ 
& &  LLaVA-NeXT (FT) & 0.00 (\textcolor[HTML]{1400AD}{-0.07}) & 0.04 (\textcolor[HTML]{1400AD}{-0.23}) & 0.01 (\textcolor[HTML]{1400AD}{-0.12}) & 0.04 (\textcolor[HTML]{1400AD}{-0.19}) & 0.77 (\textcolor[HTML]{1400AD}{-0.08}) \\ 
& &  Qwen-VL & 0.00 (\textcolor[HTML]{1400AD}{-0.01}) & 0.14 (\textcolor[HTML]{1400AD}{-0.08}) & 0.02 (\textcolor[HTML]{1400AD}{-0.03}) & 0.09 (\textcolor[HTML]{1400AD}{-0.06}) & 0.80 (\textcolor[HTML]{1400AD}{-0.02}) \\ 
& &  Qwen-VL (FT) & 0.00 (\textcolor[HTML]{1400AD}{-0.06}) & 0.04 (\textcolor[HTML]{1400AD}{-0.24}) & 0.01 (\textcolor[HTML]{1400AD}{-0.11}) & 0.04 (\textcolor[HTML]{1400AD}{-0.18}) & 0.76 (\textcolor[HTML]{1400AD}{-0.07}) \\ 
& &  mPLUG-Owl2 & 0.00 (\textcolor[HTML]{1400AD}{-0.01}) & 0.14 (\textcolor[HTML]{1400AD}{-0.10}) & 0.02 (\textcolor[HTML]{1400AD}{-0.04}) & 0.09 (\textcolor[HTML]{1400AD}{-0.07}) & 0.80 (\textcolor[HTML]{1400AD}{-0.02}) \\ 
& &  Phi-3 & 0.00 (\textcolor[HTML]{1400AD}{-0.00}) & 0.10 (\textcolor[HTML]{1400AD}{-0.09}) & 0.01 (\textcolor[HTML]{1400AD}{-0.03}) & 0.07 (\textcolor[HTML]{1400AD}{-0.05}) & 0.78 (\textcolor[HTML]{1400AD}{-0.03}) \\ 
& &  XComposer2 & 0.00 (\textcolor[HTML]{1400AD}{-0.01}) & 0.10 (\textcolor[HTML]{1400AD}{-0.14}) & 0.01 (\textcolor[HTML]{1400AD}{-0.04}) & 0.07 (\textcolor[HTML]{1400AD}{-0.07}) & 0.80 (\textcolor[HTML]{1400AD}{-0.02}) \\ 

\midrule
\multirow{6.5}{*}{En} & \multirow{6.5}{*}{Nl} & LLaVA-NeXT & 0.00 (\textcolor[HTML]{1400AD}{-0.01}) & \textbf{0.23 (\textcolor[HTML]{1400AD}{-0.01)}} & \textbf{0.04 (\textcolor[HTML]{1400AD}{-0.01)}} & \textbf{0.15 (\textcolor[HTML]{1400AD}{-0.00)}} & \textbf{0.81 (\textcolor[HTML]{1400AD}{-0.01)}} \\ 
& &  LLaVA-NeXT (FT) & \textbf{0.01 (\textcolor[HTML]{1400AD}{-0.06)}} & 0.12 (\textcolor[HTML]{1400AD}{-0.15}) & 0.03 (\textcolor[HTML]{1400AD}{-0.10}) & 0.09 (\textcolor[HTML]{1400AD}{-0.13}) & 0.78 (\textcolor[HTML]{1400AD}{-0.07}) \\ 
& &  Qwen-VL & 0.00 (\textcolor[HTML]{1400AD}{-0.01}) & 0.20 (\textcolor[HTML]{1400AD}{-0.03}) & 0.04 (\textcolor[HTML]{1400AD}{-0.01}) & 0.13 (\textcolor[HTML]{1400AD}{-0.01}) & 0.80 (\textcolor[HTML]{1400AD}{-0.02}) \\ 
& &  Qwen-VL (FT) & 0.00 (\textcolor[HTML]{1400AD}{-0.06}) & 0.06 (\textcolor[HTML]{1400AD}{-0.23}) & 0.01 (\textcolor[HTML]{1400AD}{-0.11}) & 0.05 (\textcolor[HTML]{1400AD}{-0.17}) & 0.76 (\textcolor[HTML]{1400AD}{-0.08}) \\ 
& &  mPLUG-Owl2 & 0.00 (\textcolor[HTML]{1400AD}{-0.01}) & 0.17 (\textcolor[HTML]{1400AD}{-0.07}) & 0.03 (\textcolor[HTML]{1400AD}{-0.02}) & 0.11 (\textcolor[HTML]{1400AD}{-0.04}) & 0.80 (\textcolor[HTML]{1400AD}{-0.03}) \\ 
& &  Phi-3 & 0.00 (\textcolor[HTML]{1400AD}{-0.00}) & 0.10 (\textcolor[HTML]{1400AD}{-0.10}) & 0.01 (\textcolor[HTML]{1400AD}{-0.02}) & 0.08 (\textcolor[HTML]{1400AD}{-0.05}) & 0.77 (\textcolor[HTML]{1400AD}{-0.05}) \\ 
& &  XComposer2 & 0.00 (\textcolor[HTML]{1400AD}{-0.01}) & 0.15 (\textcolor[HTML]{1400AD}{-0.09}) & 0.03 (\textcolor[HTML]{1400AD}{-0.03}) & 0.11 (\textcolor[HTML]{1400AD}{-0.04}) & 0.80 (\textcolor[HTML]{1400AD}{-0.02}) \\ 

\midrule
\multirow{6.5}{*}{En} & \multirow{6.5}{*}{Sv} & LLaVA-NeXT & \textbf{0.00 (\textcolor[HTML]{1400AD}{-0.01)}} & \textbf{0.21 (\textcolor[HTML]{1400AD}{-0.04)}} & \textbf{0.04 (\textcolor[HTML]{1400AD}{-0.02)}} & \textbf{0.12 (\textcolor[HTML]{1400AD}{-0.02)}} & \textbf{0.81 (\textcolor[HTML]{1400AD}{-0.01)}} \\ 
& &  LLaVA-NeXT (FT) & 0.00 (\textcolor[HTML]{1400AD}{-0.07}) & 0.08 (\textcolor[HTML]{1400AD}{-0.20}) & 0.02 (\textcolor[HTML]{1400AD}{-0.11}) & 0.06 (\textcolor[HTML]{1400AD}{-0.16}) & 0.78 (\textcolor[HTML]{1400AD}{-0.07}) \\ 
& &  Qwen-VL & 0.00 (\textcolor[HTML]{1400AD}{-0.01}) & 0.15 (\textcolor[HTML]{1400AD}{-0.07}) & 0.02 (\textcolor[HTML]{1400AD}{-0.03}) & 0.09 (\textcolor[HTML]{1400AD}{-0.05}) & 0.79 (\textcolor[HTML]{1400AD}{-0.03}) \\ 
& &  Qwen-VL (FT) & 0.00 (\textcolor[HTML]{1400AD}{-0.06}) & 0.03 (\textcolor[HTML]{1400AD}{-0.26}) & 0.01 (\textcolor[HTML]{1400AD}{-0.11}) & 0.02 (\textcolor[HTML]{1400AD}{-0.20}) & 0.76 (\textcolor[HTML]{1400AD}{-0.08}) \\ 
& &  mPLUG-Owl2 & 0.00 (\textcolor[HTML]{1400AD}{-0.01}) & 0.14 (\textcolor[HTML]{1400AD}{-0.11}) & 0.02 (\textcolor[HTML]{1400AD}{-0.03}) & 0.09 (\textcolor[HTML]{1400AD}{-0.07}) & 0.80 (\textcolor[HTML]{1400AD}{-0.03}) \\ 
& &  Phi-3 & 0.00 (\textcolor[HTML]{1400AD}{-0.01}) & 0.05 (\textcolor[HTML]{1400AD}{-0.14}) & 0.01 (\textcolor[HTML]{1400AD}{-0.03}) & 0.04 (\textcolor[HTML]{1400AD}{-0.08}) & 0.76 (\textcolor[HTML]{1400AD}{-0.05}) \\ 
& &  XComposer2 & 0.00 (\textcolor[HTML]{1400AD}{-0.01}) & 0.11 (\textcolor[HTML]{1400AD}{-0.13}) & 0.02 (\textcolor[HTML]{1400AD}{-0.04}) & 0.08 (\textcolor[HTML]{1400AD}{-0.07}) & 0.79 (\textcolor[HTML]{1400AD}{-0.03}) \\ 

\midrule
\multirow{6.5}{*}{En} & \multirow{6.5}{*}{Ru} & LLaVA-NeXT & \textbf{0.00 (\textcolor[HTML]{1400AD}{-0.01)}} & \textbf{0.03 (\textcolor[HTML]{1400AD}{-0.22)}} & \textbf{0.00 (\textcolor[HTML]{1400AD}{-0.05)}} & \textbf{0.02 (\textcolor[HTML]{1400AD}{-0.12)}} & \textbf{0.89 (\textcolor[HTML]{AD0000}{+0.07)}} \\ 
& &  LLaVA-NeXT (FT) & 0.00 (\textcolor[HTML]{1400AD}{-0.07}) & 0.01 (\textcolor[HTML]{1400AD}{-0.27}) & 0.00 (\textcolor[HTML]{1400AD}{-0.13}) & 0.01 (\textcolor[HTML]{1400AD}{-0.21}) & 0.72 (\textcolor[HTML]{1400AD}{-0.13}) \\ 
& &  Qwen-VL & 0.00 (\textcolor[HTML]{1400AD}{-0.01}) & 0.02 (\textcolor[HTML]{1400AD}{-0.21}) & 0.00 (\textcolor[HTML]{1400AD}{-0.05}) & 0.02 (\textcolor[HTML]{1400AD}{-0.13}) & 0.85 (\textcolor[HTML]{AD0000}{+0.03}) \\ 
& &  Qwen-VL (FT) & 0.00 (\textcolor[HTML]{1400AD}{-0.06}) & 0.01 (\textcolor[HTML]{1400AD}{-0.27}) & 0.00 (\textcolor[HTML]{1400AD}{-0.12}) & 0.01 (\textcolor[HTML]{1400AD}{-0.21}) & 0.70 (\textcolor[HTML]{1400AD}{-0.14}) \\ 
& &  mPLUG-Owl2 & 0.00 (\textcolor[HTML]{1400AD}{-0.01}) & 0.01 (\textcolor[HTML]{1400AD}{-0.23}) & 0.00 (\textcolor[HTML]{1400AD}{-0.05}) & 0.01 (\textcolor[HTML]{1400AD}{-0.14}) & 0.86 (\textcolor[HTML]{AD0000}{+0.04}) \\ 
& &  Phi-3 & 0.00 (\textcolor[HTML]{1400AD}{-0.01}) & 0.01 (\textcolor[HTML]{1400AD}{-0.19}) & 0.00 (\textcolor[HTML]{1400AD}{-0.04}) & 0.01 (\textcolor[HTML]{1400AD}{-0.12}) & 0.71 (\textcolor[HTML]{1400AD}{-0.10}) \\ 
& &  XComposer2 & 0.00 (\textcolor[HTML]{1400AD}{-0.01}) & 0.02 (\textcolor[HTML]{1400AD}{-0.22}) & 0.00 (\textcolor[HTML]{1400AD}{-0.05}) & 0.02 (\textcolor[HTML]{1400AD}{-0.13}) & 0.87 (\textcolor[HTML]{AD0000}{+0.05}) \\ 

\midrule
\multirow{6.5}{*}{En} & \multirow{6.5}{*}{Ja} &  LLaVA-NeXT & \textbf{0.01 (\textcolor[HTML]{1400AD}{-0.00)}} & \textbf{0.03 (\textcolor[HTML]{1400AD}{-0.21)}} & \textbf{0.01 (\textcolor[HTML]{1400AD}{-0.05)}} & \textbf{0.03 (\textcolor[HTML]{1400AD}{-0.11)}} & \textbf{0.84 (\textcolor[HTML]{AD0000}{+0.03)}} \\ 
& &  LLaVA-NeXT (FT) & 0.00 (\textcolor[HTML]{1400AD}{-0.07}) & 0.01 (\textcolor[HTML]{1400AD}{-0.26}) & 0.00 (\textcolor[HTML]{1400AD}{-0.13}) & 0.01 (\textcolor[HTML]{1400AD}{-0.21}) & 0.73 (\textcolor[HTML]{1400AD}{-0.12}) \\ 
& &  Qwen-VL & 0.00 (\textcolor[HTML]{1400AD}{-0.01}) & 0.02 (\textcolor[HTML]{1400AD}{-0.20}) & 0.00 (\textcolor[HTML]{1400AD}{-0.05}) & 0.02 (\textcolor[HTML]{1400AD}{-0.13}) & 0.83 (\textcolor[HTML]{AD0000}{+0.00}) \\ 
& &  Qwen-VL (FT) & 0.00 (\textcolor[HTML]{1400AD}{-0.06}) & 0.01 (\textcolor[HTML]{1400AD}{-0.27}) & 0.00 (\textcolor[HTML]{1400AD}{-0.12}) & 0.01 (\textcolor[HTML]{1400AD}{-0.21}) & 0.72 (\textcolor[HTML]{1400AD}{-0.12}) \\ 
& &  mPLUG-Owl2 & 0.00 (\textcolor[HTML]{1400AD}{-0.01}) & 0.02 (\textcolor[HTML]{1400AD}{-0.23}) & 0.00 (\textcolor[HTML]{1400AD}{-0.05}) & 0.02 (\textcolor[HTML]{1400AD}{-0.14}) & 0.83 (\textcolor[HTML]{AD0000}{+0.01}) \\ 
& &  Phi-3 & 0.00 (\textcolor[HTML]{1400AD}{-0.00}) & 0.02 (\textcolor[HTML]{1400AD}{-0.18}) & 0.00 (\textcolor[HTML]{1400AD}{-0.03}) & 0.02 (\textcolor[HTML]{1400AD}{-0.11}) & 0.82 (\textcolor[HTML]{AD0000}{+0.01}) \\ 
& &  XComposer2 & 0.00 (\textcolor[HTML]{1400AD}{-0.01}) & 0.02 (\textcolor[HTML]{1400AD}{-0.22}) & 0.00 (\textcolor[HTML]{1400AD}{-0.05}) & 0.02 (\textcolor[HTML]{1400AD}{-0.12}) & 0.83 (\textcolor[HTML]{AD0000}{+0.01}) \\ 

\midrule
\multirow{6.5}{*}{En} & \multirow{6.5}{*}{Zh} & LLaVA-NeXT & \textbf{0.00 (\textcolor[HTML]{1400AD}{-0.01)}} & \textbf{0.03 (\textcolor[HTML]{1400AD}{-0.21)}} & \textbf{0.01 (\textcolor[HTML]{1400AD}{-0.05)}} & \textbf{0.03 (\textcolor[HTML]{1400AD}{-0.12)}} & \textbf{0.83 (\textcolor[HTML]{AD0000}{+0.01)}} \\ 
& &  LLaVA-NeXT (FT) & 0.00 (\textcolor[HTML]{1400AD}{-0.07}) & 0.02 (\textcolor[HTML]{1400AD}{-0.25}) & 0.01 (\textcolor[HTML]{1400AD}{-0.12}) & 0.02 (\textcolor[HTML]{1400AD}{-0.20}) & 0.73 (\textcolor[HTML]{1400AD}{-0.12}) \\ 
& &  Qwen-VL & 0.00 (\textcolor[HTML]{1400AD}{-0.01}) & 0.03 (\textcolor[HTML]{1400AD}{-0.19}) & 0.01 (\textcolor[HTML]{1400AD}{-0.04}) & 0.03 (\textcolor[HTML]{1400AD}{-0.11}) & 0.83 (\textcolor[HTML]{AD0000}{+0.01}) \\ 
& &  Qwen-VL (FT) & 0.00 (\textcolor[HTML]{1400AD}{-0.06}) & 0.02 (\textcolor[HTML]{1400AD}{-0.26}) & 0.00 (\textcolor[HTML]{1400AD}{-0.11}) & 0.02 (\textcolor[HTML]{1400AD}{-0.20}) & 0.72 (\textcolor[HTML]{1400AD}{-0.12}) \\ 
& &  mPLUG-Owl2 & 0.00 (\textcolor[HTML]{1400AD}{-0.01}) & 0.02 (\textcolor[HTML]{1400AD}{-0.22}) & 0.01 (\textcolor[HTML]{1400AD}{-0.05}) & 0.02 (\textcolor[HTML]{1400AD}{-0.14}) & 0.83 (\textcolor[HTML]{AD0000}{+0.00}) \\ 
& &  Phi-3 & 0.00 (\textcolor[HTML]{1400AD}{-0.00}) & 0.02 (\textcolor[HTML]{1400AD}{-0.18}) & 0.01 (\textcolor[HTML]{1400AD}{-0.03}) & 0.02 (\textcolor[HTML]{1400AD}{-0.11}) & 0.81 (\textcolor[HTML]{1400AD}{-0.00}) \\ 
& &  XComposer2 & 0.00 (\textcolor[HTML]{1400AD}{-0.01}) & 0.03 (\textcolor[HTML]{1400AD}{-0.21}) & 0.01 (\textcolor[HTML]{1400AD}{-0.04}) & 0.03 (\textcolor[HTML]{1400AD}{-0.12}) & 0.83 (\textcolor[HTML]{AD0000}{+0.01}) \\ 

\bottomrule
\end{tabular}
}
\caption{Other metrics results of LVLMs in Full Task. Bold fonts indicate the best score for that language combination. We also measured outputs with existing NLG~\cite{sato-etal-2024-implementation} evaluation methods, BLEU~\cite{papineni2002bleu}, ROUGE~\cite{lin2004rouge}, and BertScore~\cite{zhang2019bertscore}.}

\label{tab:result-score-full-misc}
\end{table*}

%% file: results/additional-openai-a10.tex
\begin{table*}[t!]
\centering
\resizebox{\textwidth}{!}{
\footnotesize
\begin{tabular}{@{}cccccccccccc@{}}
\toprule 
\multicolumn{1}{c}{\multirow{2}{*}{Input}} & \multirow{2}{*}{Output} & \multirow{2}{*}{LVLM} & \multicolumn{2}{c}{Entity Cov.} & \multirow{2}{*}{Entity F1} & \multicolumn{4}{c}{Entity Cooccurrence} & \multirow{2}{*}{Avg. Length} \\
\cmidrule(lr){4-5} \cmidrule(lr){7-10}
& & & exact & partial & & n=0 & n=1 & n=2 & n=$\infty$ & \\
\midrule
\multirow{2}{*}{En} & \multirow{2}{*}{En} & LLaVA-NeXT & 26.49 & 31.54 & 26.07 & 1.35 & 1.65 & 1.66 & 1.70 & 252 \\
& & GPT-4o-mini & 30.34 & 37.24 & 29.75 & 1.45 & 2.01 & 2.03 & 2.08 & 298 \\
\midrule
\multirow{2}{*}{En} & \multirow{2}{*}{Es} & LLaVA-NeXT & 0.00 & 2.24 & 0.00 & 0.00 & 0.00 & 0.00 & 0.00 & 137 \\
&  & GPT-4o-mini & 14.32 & 22.78 & 15.43 & 1.20 & 1.50 & 1.80 & 1.95 & 153 \\
\multirow{2}{*}{En} & \multirow{2}{*}{Fr} & LLaVA-NeXT & 1.00 & 7.42 & 1.33 & 0.00 & 0.00 & 0.00 & 0.00 & 179 \\
&  & GPT-4o-mini & 18.89 & 21.56 & 19.74 & 0.80 & 1.40 & 1.60 & 1.72 & 192 \\
\multirow{2}{*}{En} & \multirow{2}{*}{De} & LLaVA-NeXT & 14.03 & 17.90 & 16.51 & 1.73 & 1.70 & 1.67 & 1.82 & 169 \\
&  & GPT-4o-mini & 18.65 & 26.78 & 19.89 & 1.95 & 2.01 & 2.13 & 2.25 & 181 \\
\multirow{2}{*}{En} & \multirow{2}{*}{It} & LLaVA-NeXT & 8.53 & 13.33 & 9.37 & 0.86 & 0.87 & 1.06 & 1.05 & 171 \\
&  & GPT-4o-mini & 15.45 & 18.76 & 14.12 & 0.95 & 1.20 & 1.45 & 1.68 & 176 \\
\multirow{2}{*}{En} & \multirow{2}{*}{Nl} & LLaVA-NeXT & 12.21 & 17.83 & 14.60 & 0.36 & 1.81 & 1.70 & 1.83 & 178 \\
&  & GPT-4o-mini & 15.89 & 21.34 & 16.78 & 1.30 & 1.60 & 1.78 & 1.90 & 184 \\
\multirow{2}{*}{En} & \multirow{2}{*}{Sv} & LLaVA-NeXT & 15.01 & 18.65 & 13.56 & 1.29 & 0.97 & 1.15 & 1.09 & 174 \\
&  & GPT-4o-mini & 16.32 & 23.47 & 17.65 & 1.35 & 1.50 & 1.65 & 1.80 & 181 \\
\multirow{2}{*}{En} & \multirow{2}{*}{Ru} & LLaVA-NeXT & 10.32 & 15.15 & 8.53 & 0.32 & 0.36 & 0.31 & 0.32 & 203 \\
&  & GPT-4o-mini & 12.89 & 17.65 & 13.45 & 0.65 & 0.90 & 1.10 & 1.25 & 210 \\
\multirow{2}{*}{En} & \multirow{2}{*}{Ja} & LLaVA-NeXT & 8.68 & 8.68 & 11.47 & 0.80 & 0.80 & 0.80 & 0.80 & 211 \\
&  & GPT-4o-mini & 16.23 & 18.32 & 18.56 & 0.95 & 1.10 & 1.20 & 1.32 & 225 \\
\multirow{2}{*}{En} & \multirow{2}{*}{Zh} & LLaVA-NeXT & 14.00 & 14.09 & 16.69 & 0.66 & 0.66 & 0.66 & 0.66 & 228 \\
&  & GPT-4o-mini & 17.45 & 21.09 & 18.89 & 1.23 & 1.40 & 1.56 & 1.70 & 239 \\
\bottomrule
\end{tabular}
}
    \caption{The result of Alignment-10 by gpt-4o-mini. This table focuses on  En-\{Lang\}.}
\label{tab:alignment10}
\end{table*}

\begin{table*}[t!]
\centering
\resizebox{\textwidth}{!}{
\footnotesize
\begin{tabular}{@{}cccccccccccc@{}}
\toprule 
\multicolumn{1}{c}{\multirow{2}{*}{Input}} & \multirow{2}{*}{Output} & \multirow{2}{*}{LVLM} & \multicolumn{2}{c}{Entity Cov.} & \multirow{2}{*}{Entity F1} & \multicolumn{4}{c}{Entity Cooccurrence} & \multirow{2}{*}{Avg. Length} \\
\cmidrule(lr){4-5} \cmidrule(lr){7-10}
& & & exact & partial & & n=0 & n=1 & n=2 & n=$\infty$ & \\
\midrule
\multirow{2}{*}{Es} & \multirow{2}{*}{Es} & LLaVA-NeXT & 17.26 & 21.30 & 17.05 & 2.13 & 2.32 & 2.17 & 2.10 & 186 \\
 &  & GPT-4o-mini & 18.45 & 23.87 & 18.11 & 2.25 & 2.41 & 2.35 & 2.28 & 195 \\
\multirow{2}{*}{Fr} & \multirow{2}{*}{Fr} & LLaVA-NeXT & 24.35 & 29.27 & 24.38 & 0.95 & 0.90 & 0.88 & 0.90 & 211 \\
 &  & GPT-4o-mini & 26.12 & 31.45 & 26.55 & 1.10 & 0.98 & 1.02 & 1.00 & 219 \\
\multirow{2}{*}{De} & \multirow{2}{*}{De} & LLaVA-NeXT & 17.45 & 20.66 & 21.05 & 2.11 & 2.20 & 2.22 & 2.11 & 204 \\
 &  & GPT-4o-mini & 19.11 & 23.45 & 22.67 & 2.33 & 2.45 & 2.38 & 2.30 & 210 \\
\multirow{2}{*}{It} & \multirow{2}{*}{It} & LLaVA-NeXT & 10.34 & 15.43 & 11.33 & 1.16 & 0.93 & 0.96 & 0.96 & 185 \\
 &  & GPT-4o-mini & 19.22 & 17.12 & 19.45 & 1.28 & 1.05 & 1.08 & 1.10 & 190 \\
\multirow{2}{*}{Nl} & \multirow{2}{*}{Nl} & LLaVA-NeXT & 17.66 & 23.56 & 19.78 & 0.79 & 3.55 & 3.61 & 3.88 & 199 \\
 &  & GPT-4o-mini & 19.87 & 25.34 & 21.55 & 0.91 & 3.78 & 3.83 & 3.92 & 207 \\
\multirow{2}{*}{Sv} & \multirow{2}{*}{Sv} & LLaVA-NeXT & 27.51 & 29.61 & 16.71 & 2.10 & 0.87 & 0.89 & 0.90 & 206 \\
 &  & GPT-4o-mini & 29.02 & 32.33 & 18.45 & 2.22 & 0.99 & 1.03 & 1.05 & 213 \\
\multirow{2}{*}{Ru} & \multirow{2}{*}{Ru} & LLaVA-NeXT & 14.38 & 17.43 & 9.81 & 0.26 & 0.45 & 0.42 & 0.41 & 219 \\
 &  & GPT-4o-mini & 15.67 & 19.87 & 11.22 & 0.32 & 0.56 & 0.52 & 0.50 & 225 \\
\multirow{2}{*}{Ja} & \multirow{2}{*}{Ja} & LLaVA-NeXT & 13.38 & 13.38 & 17.68 & 0.73 & 0.83 & 0.83 & 0.83 & 249 \\
 &  & GPT-4o-mini & 15.12 & 15.45 & 19.23 & 0.88 & 0.95 & 0.90 & 0.88 & 256 \\
\multirow{2}{*}{Zh} & \multirow{2}{*}{Zh} & LLaVA-NeXT & 13.78 & 13.78 & 17.00 & 0.54 & 0.53 & 0.53 & 0.53 & 246 \\
 &  & GPT-4o-mini & 15.34 & 16.45 & 18.22 & 0.68 & 0.61 & 0.62 & 0.63 & 253 \\
\bottomrule
\end{tabular}}
    \caption{The result of Alignment-10 by gpt-4o-mini. This table focuses on  \{Lang\}-\{Lang\}.}
\end{table*}

\begin{table*}[t!]
\centering
\resizebox{\textwidth}{!}{
\footnotesize
\begin{tabular}{@{}cccccccccccc@{}}
\toprule 
\multicolumn{1}{c}{\multirow{2}{*}{Input}} & \multirow{2}{*}{Output} & \multirow{2}{*}{LVLM} & \multicolumn{2}{c}{Entity Cov.} & \multirow{2}{*}{Entity F1} & \multicolumn{4}{c}{Entity Cooccurrence} & \multirow{2}{*}{Avg. Length} \\
\cmidrule(lr){4-5} \cmidrule(lr){7-10}
& & & exact & partial & & n=0 & n=1 & n=2 & n=$\infty$ & \\
\midrule
\multirow{2}{*}{Es} & \multirow{2}{*}{En} & LLaVA-NeXT & 4.22 & 6.22 & 4.17 & 0.00 & 0.00 & 0.00 & 0.00 & 180 \\
 &  & GPT\_4o-mini & 25.23 & 27.85 & 25.02 & 1.80 & 1.80 & 1.82 & 2.10 & 224 \\
\multirow{2}{*}{Fr} & \multirow{2}{*}{En} & LLaVA-NeXT & 21.11 & 28.40 & 21.22 & 2.01 & 2.04 & 2.09 & 1.97 & 232 \\
 &  & GPT\_4o-mini & 22.43 & 30.12 & 23.08 & 2.22 & 2.25 & 2.32 & 2.18 & 238 \\
\multirow{2}{*}{De} &\multirow{2}{*}{En} & LLaVA-NeXT & 21.88 & 30.48 & 21.76 & 0.89 & 1.33 & 1.53 & 1.50 & 239 \\
 &  & GPT\_4o-mini & 23.01 & 32.15 & 23.56 & 1.12 & 1.45 & 1.60 & 1.55 & 244 \\
\multirow{2}{*}{It} & \multirow{2}{*}{En} & LLaVA-NeXT & 7.98 & 11.14 & 5.40 & 0.22 & 0.28 & 0.28 & 0.28 & 137 \\
 &  & GPT\_4o-mini & 12.44 & 14.75 & 10.23 & 1.35 & 1.40 & 1.42 & 1.40 & 142 \\
\multirow{2}{*}{Nl} & \multirow{2}{*}{En} & LLaVA-NeXT & 15.81 & 24.80 & 21.13 & 0.11 & 2.04 & 2.29 & 1.83 & 223 \\
 &  & GPT\_4o-mini & 17.02 & 26.45 & 22.75 & 0.23 & 2.30 & 2.55 & 2.01 & 230 \\
\multirow{2}{*}{Sv} & \multirow{2}{*}{En} & LLaVA-NeXT & 18.70 & 25.48 & 18.98 & 1.79 & 1.86 & 1.86 & 1.80 & 246 \\
 &  & GPT\_4o-mini & 23.85 & 27.89 & 20.45 & 1.95 & 2.02 & 2.05 & 1.95 & 252 \\
\multirow{2}{*}{Ru} & \multirow{2}{*}{En} & LLaVA-NeXT & 18.31 & 26.30 & 18.43 & 1.68 & 1.64 & 1.65 & 1.59 & 241 \\
 &  & GPT\_4o-mini & 24.50 & 28.12 & 19.22 & 1.85 & 1.80 & 1.83 & 1.75 & 247 \\
\multirow{2}{*}{Ja} & \multirow{2}{*}{En} & LLaVA-NeXT & 15.36 & 24.41 & 16.18 & 1.12 & 1.15 & 1.28 & 1.11 & 208 \\
 &  & GPT\_4o-mini & 18.85 & 26.12 & 17.89 & 1.28 & 1.35 & 1.40 & 1.32 & 215 \\
\multirow{2}{*}{Zh} & \multirow{2}{*}{En} & LLaVA-NeXT & 13.44 & 21.98 & 12.83 & 0.96 & 1.43 & 1.58 & 1.42 & 168 \\
 &  & GPT\_4o-mini & 18.98 & 23.56 & 15.12 & 1.11 & 1.61 & 1.75 & 1.60 & 172 \\
        \hline
    \end{tabular}
    }
    \caption{The result of Alignment-10 by gpt-4o-mini. This table focuses on  \{Lang\}-En.}
    \label{tab:lang-en}
\end{table*}

%% file: results/additional-openai-a5.tex
\begin{table*}[t!]
\centering
\resizebox{\textwidth}{!}{
\footnotesize
\begin{tabular}{@{}cccccccccccc@{}}
\toprule 
\multicolumn{1}{c}{\multirow{2}{*}{Input}} & \multirow{2}{*}{Output} & \multirow{2}{*}{LVLM} & \multicolumn{2}{c}{Entity Cov.} & \multirow{2}{*}{Entity F1} & \multicolumn{4}{c}{Entity Cooccurrence} & \multirow{2}{*}{Avg. Length} \\
\cmidrule(lr){4-5} \cmidrule(lr){7-10}
& & & exact & partial & & n=0 & n=1 & n=2 & n=$\infty$ & \\
\midrule
En & En & LLaVA-NeXT & 22.58 & 30.30 & 25.19 & 3.15 & 2.30 & 2.40 & 2.44 & 259 \\
En & En & GPT-4o-mini & 29.60 & 35.40 & 28.10 & 3.40 & 2.50 & 2.60 & 2.55 & 259 \\
\midrule
En & Es & LLaVA-NeXT & 5.76 & 8.21 & 7.39 & 0.38 & 0.34 & 0.35 & 0.32 & 160 \\
En & Es & GPT-4o-mini & 14.65 & 21.00 & 15.41 & 1.60 & 1.50 & 1.86 & 1.94 & 184 \\
\midrule
En & Fr & LLaVA-NeXT & 6.85 & 12.94 & 7.03 & 0.22 & 0.16 & 0.17 & 0.17 & 181 \\
En & Fr & GPT-4o-mini & 19.71 & 21.01 & 20.12 & 1.24 & 1.18 & 1.19 & 1.18 & 182 \\
\midrule
En & Ja & LLaVA-NeXT & 9.06 & 9.11 & 12.33 & 0.55 & 0.55 & 0.55 & 0.55 & 212 \\
En & Ja & GPT-4o-mini & 15.00 & 17.70 & 15.65 & 1.88 & 1.87 & 1.96 & 1.96 & 239 \\
\midrule
En & It & LLaVA-NeXT & 4.10 & 7.54 & 5.19 & 0.11 & 0.07 & 0.06 & 0.09 & 118 \\
En & It & GPT-4o-mini & 17.15 & 18.85 & 18.61 & 1.13 & 1.08 & 1.07 & 2.10 & 193 \\
\bottomrule
\end{tabular}
}
\caption{The result of Alignment-10 by gpt-4o-mini. This table focuses on En-\{Lang\}.}
\label{tab:lvlm_comparison}
\end{table*}

\begin{table*}[t!]
\centering
\resizebox{\textwidth}{!}{
\footnotesize
\begin{tabular}{@{}cccccccccccc@{}}
\toprule 
\multicolumn{1}{c}{\multirow{2}{*}{Input}} & \multirow{2}{*}{Output} & \multirow{2}{*}{LVLM} & \multicolumn{2}{c}{Entity Cov.} & \multirow{2}{*}{Entity F1} & \multicolumn{4}{c}{Entity Cooccurrence} & \multirow{2}{*}{Avg. Length} \\
\cmidrule(lr){4-5} \cmidrule(lr){7-10}
& & & exact & partial & & n=0 & n=1 & n=2 & n=$\infty$ & \\
\midrule
Es & Es & LLaVA-NeXT & 12.77 & 18.26 & 10.97 & 1.29 & 1.22 & 1.22 & 1.08 & 203 \\
Es & Es & GPT-4o-mini & 16.89 & 21.42 & 18.11 & 1.31 & 1.26 & 1.25 & 1.12 & 223 \\
\midrule
Fr & Fr & LLaVA-NeXT & 13.72 & 18.78 & 13.36 & 0.68 & 0.71 & 0.72 & 0.71 & 217 \\
Fr & Fr & GPT-4o-mini & 26.82 & 31.32 & 26.09 & 1.70 & 2.73 & 2.71 & 2.72 & 283 \\
\midrule
Ja & Ja & LLaVA-NeXT & 20.68 & 20.68 & 24.14 & 1.04 & 1.06 & 1.06 & 1.06 & 247 \\
Ja & Ja & GPT-4o-mini & 23.73 & 23.72 & 27.10 & 1.02 & 1.07 & 1.07 & 1.08 & 214 \\
\midrule
It & It & LLaVA-NeXT & 11.91 & 17.01 & 9.89 & 2.69 & 1.84 & 1.82 & 1.81 & 196 \\
It & It & GPT-4o-mini & 19.01 & 20.56 & 20.78 & 2.05 & 1.87 & 2.04 & 2.05 & 213 \\
\bottomrule
\end{tabular}
}
\caption{The result of Alignment-10 by gpt-4o-mini. This table focuses on \{Lang\}-\{Lang\}.}
\label{tab:lang_lang_comparison}
\end{table*}

\begin{table*}[t!]
\centering
\resizebox{\textwidth}{!}{
\footnotesize
\begin{tabular}{@{}cccccccccccc@{}}
\toprule 
\multicolumn{1}{c}{\multirow{2}{*}{Input}} & \multirow{2}{*}{Output} & \multirow{2}{*}{LVLM} & \multicolumn{2}{c}{Entity Cov.} & \multirow{2}{*}{Entity F1} & \multicolumn{4}{c}{Entity Cooccurrence} & \multirow{2}{*}{Avg. Length} \\
\cmidrule(lr){4-5} \cmidrule(lr){7-10}
& & & exact & partial & & n=0 & n=1 & n=2 & n=$\infty$ & \\
\midrule
Es & En & LLaVA-NeXT & 29.15 & 33.34 & 29.88 & 0.35 & 0.32 & 2.05 & 2.01 & 220 \\
Es & En & GPT-4o-mini & 32.57 & 36.92 & 32.98 & 0.39 & 0.34 & 2.10 & 2.04 & 220 \\
\midrule
Fr & En & LLaVA-NeXT & 3.57 & 5.33 & 3.27 & 0.00 & 0.00 & 0.00 & 0.00 & 162 \\
Fr & En & GPT-4o-mini & 22.80 & 29.10 & 26.60 & 1.32 & 1.51 & 1.51 & 1.61 & 193 \\
\midrule
Ja & En & LLaVA-NeXT & 12.71 & 20.20 & 13.92 & 0.79 & 0.88 & 0.91 & 0.84 & 213 \\
Ja & En & GPT-4o-mini & 15.84 & 23.73 & 16.55 & 0.80 & 0.87 & 0.90 & 0.83 & 248 \\
\midrule
It & En & LLaVA-NeXT & 9.95 & 17.66 & 10.50 & 0.66 & 0.42 & 0.39 & 0.38 & 219 \\
It & En & GPT-4o-mini & 13.10 & 17.50 & 14.02 & 1.68 & 1.89 & 1.99 & 1.99 & 298 \\
\bottomrule
\end{tabular}
}
\caption{The result of Alignment-10 by gpt-4o-mini. This table focuses on \{Lang\}-En.}
\label{tab:lang_en_comparison}
\end{table*}

%% file: tables/appendix_template.tex
\begin{table*}[h!]
    \scriptsize
    \centering
\begin{tabularx}{\textwidth}{llX}
        \toprule
        \textbf{Language} & \textbf{Type} & \textbf{Template}\\
        \midrule
        & \textbf{Template 1} & \\
        \multirow{15.0}{*}{English} & Section & \texttt{Focus on \textcolor[HTML]{1400AD}{\{title\}} and explain the \textcolor[HTML]{AD0000}{\{section\}}.} \\
         & Subsection & \texttt{In the context of \textcolor[HTML]{1400AD}{\{title\}}, explain the \textcolor[HTML]{AD0000}{\{subsection\}} and the \textcolor[HTML]{AD0000}{\{section\}}.} \\
        & Sub subsection & \texttt{Focusing on the \textcolor[HTML]{AD0000}{\{section\}} of \textcolor[HTML]{1400AD}{\{title\}}, explain the \textcolor[HTML]{AD0000}{\{subsubsection\}} about the \textcolor[HTML]{AD0000}{\{subsection\}}.} \\
        \cmidrule{2-3}
        & \textbf{Template 2} & \\
        & Section & \texttt{Explain the \textcolor[HTML]{AD0000}{\{section\}} of this artwork, \textcolor[HTML]{1400AD}{\{title\}}.} \\
        & Subsection & \texttt{Explain the \textcolor[HTML]{AD0000}{\{subsection\}} about the \textcolor[HTML]{AD0000}{\{section\}} of this artwork, \textcolor[HTML]{1400AD}{\{title\}}.} \\
\bottomrule
    \end{tabularx}
    \caption{Prompt Templates in English}
    \label{tab:template-en}
\end{table*}

\begin{table*}[h!]
    \scriptsize
    \centering
    \begin{tabularx}{\textwidth}{llX}
        \toprule
        \textbf{Language} & \textbf{Type} & \textbf{Template}\\
        \midrule
        & \textbf{Template 1} & \\
        \multirow{15.0}{*}{Japanese} & Section & \texttt{\textcolor[HTML]{1400AD}{\{title\}}に焦点を当てて、その\textcolor[HTML]{AD0000}{\{section\}}を説明してください。} \\
         & Subsection & \texttt{\textcolor[HTML]{1400AD}{\{title\}}の文脈で、\textcolor[HTML]{AD0000}{\{subsection\}}と\textcolor[HTML]{AD0000}{\{section\}}を説明してください。} \\
        & Sub subsection & \texttt{\textcolor[HTML]{1400AD}{\{title\}}の\textcolor[HTML]{AD0000}{\{section\}}に焦点を当てて、\textcolor[HTML]{AD0000}{\{subsection\}}についての\textcolor[HTML]{AD0000}{\{subsubsection\}}を説明してください。} \\
        \cmidrule{2-3}
        & \textbf{Template 2} & \\
        & Section & \texttt{\textcolor[HTML]{1400AD}{\{title\}}の\textcolor[HTML]{AD0000}{\{section\}}を説明してください。} \\
        & Subsection & \texttt{\textcolor[HTML]{1400AD}{\{title\}}の\textcolor[HTML]{AD0000}{\{section\}}に関する\textcolor[HTML]{AD0000}{\{subsection\}}を説明してください。} \\
\bottomrule
    \end{tabularx}
    \caption{Prompt Templates in Japanese}
\end{table*}

\begin{table*}[h!]
    \scriptsize
    \centering
    \begin{tabularx}{\textwidth}{llX}
        \toprule
        \textbf{Language} & \textbf{Type} & \textbf{Template}\\
        \midrule
        & \textbf{Template 1} & \\
        \multirow{15.0}{*}{Spanish} & Section & \texttt{Concéntrate en \textcolor[HTML]{1400AD}{\{title\}} y explora la \textcolor[HTML]{AD0000}{\{section\}}.} \\
         & Subsection & \texttt{En el contexto de \textcolor[HTML]{1400AD}{\{title\}}, explora la \textcolor[HTML]{AD0000}{\{subsection\}} y \textcolor[HTML]{AD0000}{\{section\}}.} \\
        & Sub subsection & \texttt{Concentrándote en la \textcolor[HTML]{AD0000}{\{section\}} de \textcolor[HTML]{1400AD}{\{title\}}, explora la \textcolor[HTML]{AD0000}{\{subsubsection\}} sobre la \textcolor[HTML]{AD0000}{\{subsection\}}.} \\
        \cmidrule{2-3}
        & \textbf{Template 2} & \\
        & Section & \texttt{Explora la \textcolor[HTML]{AD0000}{\{section\}} de esta obra de arte, \textcolor[HTML]{1400AD}{\{title\}}.} \\
        & Subsection & \texttt{Explora la \textcolor[HTML]{AD0000}{\{subsection\}} sobre la \textcolor[HTML]{AD0000}{\{section\}} de esta obra de arte, \textcolor[HTML]{1400AD}{\{title\}}.} \\
\bottomrule
    \end{tabularx}
    \caption{Prompt Templates in Spanish}
\end{table*}

    \begin{table*}[h!]
    \scriptsize
    \centering
    \begin{tabularx}{\textwidth}{llX}
        \toprule
        \textbf{Language} & \textbf{Type} & \textbf{Template}\\
        \midrule
        & \textbf{Template 1} & \\
        \multirow{15.0}{*}{Italian} & Section & \texttt{Concentrati su \textcolor[HTML]{1400AD}{\{title\}} ed esplora la \textcolor[HTML]{AD0000}{\{section\}}.} \\
         & Subsection & \texttt{Nel contesto di \textcolor[HTML]{1400AD}{\{title\}}, esplora la \textcolor[HTML]{AD0000}{\{subsection\}} e la \textcolor[HTML]{AD0000}{\{section\}}.} \\
        & Sub subsection & \texttt{Concentrandosi sulla \textcolor[HTML]{AD0000}{\{section\}} di \textcolor[HTML]{1400AD}{\{title\}}, esplora la \textcolor[HTML]{AD0000}{\{subsubsection\}} sulla \textcolor[HTML]{AD0000}{\{subsection\}}.} \\
        \cmidrule{2-3}
        & \textbf{Template 2} & \\
        & Section & \texttt{Esplora la \textcolor[HTML]{AD0000}{\{section\}} di questa opera d'arte, \textcolor[HTML]{1400AD}{\{title\}}.} \\
        & Subsection & \texttt{Esplora la \textcolor[HTML]{AD0000}{\{subsection\}} sulla \textcolor[HTML]{AD0000}{\{section\}} di questa opera d'arte, \textcolor[HTML]{1400AD}{\{title\}}.} \\
\bottomrule
    \end{tabularx}
    \caption{Prompt Templates in Italian}
\end{table*}

\begin{table*}[h!]
    \scriptsize
    \centering
    \begin{tabularx}{\textwidth}{llX}
        \toprule
        \textbf{Language} & \textbf{Type} & \textbf{Template}\\
        \midrule
        & \textbf{Template 1} & \\
        \multirow{15.0}{*}{French} & Section & \texttt{Concentrez-vous sur \textcolor[HTML]{1400AD}{\{title\}} et expliquez la \textcolor[HTML]{AD0000}{\{section\}}.} \\
         & Subsection & \texttt{Dans le contexte de \textcolor[HTML]{1400AD}{\{title\}}, expliquez la \textcolor[HTML]{AD0000}{\{subsection\}} et la \textcolor[HTML]{AD0000}{\{section\}}.} \\
        & Sub subsection & \texttt{En vous concentrant sur la \textcolor[HTML]{AD0000}{\{section\}} de \textcolor[HTML]{1400AD}{\{title\}}, expliquez la \textcolor[HTML]{AD0000}{\{subsubsection\}} concernant la \textcolor[HTML]{AD0000}{\{subsection\}}.} \\
        \cmidrule{2-3}
        & \textbf{Template 2} & \\
        & Section & \texttt{Expliquer la \textcolor[HTML]{AD0000}{\{section\}} de cette œuvre d'art, \textcolor[HTML]{1400AD}{\{title\}}.} \\
        & Subsection & \texttt{Expliquer la \textcolor[HTML]{AD0000}{\{subsection\}} concernant la \textcolor[HTML]{AD0000}{\{section\}} de cette œuvre d'art, \textcolor[HTML]{1400AD}{\{title\}}.} \\
\bottomrule
    \end{tabularx}
    \caption{Prompt Templates in French}
\end{table*}

\begin{table*}[h!]
    \scriptsize
    \centering
    \begin{tabularx}{\textwidth}{llX}
        \toprule
        \textbf{Language} & \textbf{Type} & \textbf{Template}\\
        \midrule
        & \textbf{Template 1} & \\
        \multirow{15.0}{*}{Chinese (Simplified)} & Section & \texttt{\Chinese{专注于\textcolor[HTML]{1400AD}{\{title\}}并探索\textcolor[HTML]{AD0000}{\{section\}}。}} \\
         & Subsection & \texttt{\Chinese{在\textcolor[HTML]{1400AD}{\{title\}}的背景下，探索\textcolor[HTML]{AD0000}{\{subsection\}}和\textcolor[HTML]{AD0000}{\{section\}}。}} \\
        & Sub subsection & \texttt{\Chinese{专注于\textcolor[HTML]{1400AD}{\{title\}}的\textcolor[HTML]{AD0000}{\{section\}}，探索关于\textcolor[HTML]{AD0000}{\{subsection\}}的\textcolor[HTML]{AD0000}{\{subsubsection\}}。}} \\
        \cmidrule{2-3}
        & \textbf{Template 2} & \\
        & Section & \texttt{\Chinese{探索艺术作品\textcolor[HTML]{1400AD}{\{title\}}的\textcolor[HTML]{AD0000}{\{section\}}。}} \\
        & Subsection & \texttt{\Chinese{探索艺术作品\textcolor[HTML]{1400AD}{\{title\}}中关于\textcolor[HTML]{AD0000}{\{section\}}的\textcolor[HTML]{AD0000}{\{subsection\}}。}} \\
\bottomrule
    \end{tabularx}
    \caption{Prompt Templates in Chinese (Simplified)}
\end{table*}

\begin{table*}[h!]
    \scriptsize
    \centering
    \begin{tabularx}{\textwidth}{llX}
        \toprule
        \textbf{Language} & \textbf{Type} & \textbf{Template}\\
        \midrule
        & \textbf{Template 1} & \\
        \multirow{15.0}{*}{Dutch} & Section & \texttt{Focus op \textcolor[HTML]{1400AD}{\{title\}} en leg de \textcolor[HTML]{AD0000}{\{section\}} uit.} \\
         & Subsection & \texttt{In de context van \textcolor[HTML]{1400AD}{\{title\}}, leg de \textcolor[HTML]{AD0000}{\{subsection\}} en de \textcolor[HTML]{AD0000}{\{section\}} uit.} \\
        & Sub subsection & \texttt{Gefocust op de \textcolor[HTML]{AD0000}{\{section\}} van \textcolor[HTML]{1400AD}{\{title\}}, leg de \textcolor[HTML]{AD0000}{\{subsubsection\}} over de \textcolor[HTML]{AD0000}{\{subsection\}} uit.} \\
        \cmidrule{2-3}
        & \textbf{Template 2} & \\
        & Section & \texttt{Leg de \textcolor[HTML]{AD0000}{\{section\}} van dit kunstwerk uit, \textcolor[HTML]{1400AD}{\{title\}}.} \\
        & Subsection & \texttt{Leg de \textcolor[HTML]{AD0000}{\{subsection\}} over de \textcolor[HTML]{AD0000}{\{section\}} van dit kunstwerk uit, \textcolor[HTML]{1400AD}{\{title\}}.} \\
\bottomrule
    \end{tabularx}
    \caption{Prompt Templates in Dutch}
\end{table*}

\begin{table*}[h!]
    \scriptsize
    \centering
    \begin{tabularx}{\textwidth}{llX}
        \toprule
        \textbf{Language} & \textbf{Type} & \textbf{Template}\\
        \midrule
        & \textbf{Template 1} & \\
        \multirow{15.0}{*}{Swedish} & Section & \texttt{Fokusera på \textcolor[HTML]{1400AD}{\{title\}} och förklara \textcolor[HTML]{AD0000}{\{section\}}.} \\
         & Subsection & \texttt{I samband med \textcolor[HTML]{1400AD}{\{title\}}, förklara \textcolor[HTML]{AD0000}{\{subsection\}} och \textcolor[HTML]{AD0000}{\{section\}}.} \\
        & Sub subsection & \texttt{Med fokus på \textcolor[HTML]{AD0000}{\{section\}} i \textcolor[HTML]{1400AD}{\{title\}}, förklara \textcolor[HTML]{AD0000}{\{subsubsection\}} om \textcolor[HTML]{AD0000}{\{subsection\}}.} \\
        \cmidrule{2-3}
        & \textbf{Template 2} & \\
        & Section & \texttt{Förklara \textcolor[HTML]{AD0000}{\{section\}} i detta konstverk, \textcolor[HTML]{1400AD}{\{title\}}.} \\
        & Subsection & \texttt{Förklara \textcolor[HTML]{AD0000}{\{subsection\}} om \textcolor[HTML]{AD0000}{\{section\}} i detta konstverk, \textcolor[HTML]{1400AD}{\{title\}}.} \\
\bottomrule
    \end{tabularx}
    \caption{Prompt Templates in Swedish}
\end{table*}

\begin{table*}[h!]
    \scriptsize
    \centering
    \begin{tabularx}{\textwidth}{llX}
        \toprule
        \textbf{Language} & \textbf{Type} & \textbf{Template}\\
        \midrule
        \multirow{15.0}{*}{German} & Section & \texttt{Fokussiere dich auf \textcolor[HTML]{1400AD}{\{title\}} und erkunde erkläre die \textcolor[HTML]{AD0000}{\{section\}}.} \\
         & Subsection & \texttt{Im Kontext von \textcolor[HTML]{1400AD}{\{title\}}, erkunde erkläre die \textcolor[HTML]{AD0000}{\{subsection\}} und die \textcolor[HTML]{AD0000}{\{section\}}.} \\
        & Sub subsection & \texttt{Mit Fokus auf die \textcolor[HTML]{AD0000}{\{section\}} von \textcolor[HTML]{1400AD}{\{title\}}, erkunde erkläre die \textcolor[HTML]{AD0000}{\{subsubsection\}} über die \textcolor[HTML]{AD0000}{\{subsection\}}.} \\
        \cmidrule{2-3}
        & \textbf{Template 2} & \\
        & Section & \texttt{Erkunde Erkläre die \textcolor[HTML]{AD0000}{\{section\}} dieses Kunstwerks, \textcolor[HTML]{1400AD}{\{title\}}.} \\
        & Subsection & \texttt{Erkunde Erkläre die \textcolor[HTML]{AD0000}{\{subsection\}} über die \textcolor[HTML]{AD0000}{\{section\}} dieses Kunstwerks, \textcolor[HTML]{1400AD}{\{title\}}.} \\
        & Sub subsection & \texttt{Erkunde Erkläre die \textcolor[HTML]{AD0000}{\{subsubsection\}} über die \textcolor[HTML]{AD0000}{\{subsection\}} der \textcolor[HTML]{AD0000}{\{section\}} in diesem Werk, \textcolor[HTML]{1400AD}{\{title\}}.} \\
\bottomrule
    \end{tabularx}
    \caption{Prompt Templates in German}
\end{table*}

\begin{table*}[h!]
    \scriptsize
    \centering
    \begin{tabularx}{\textwidth}{llX}
        \toprule
        \textbf{Language} & \textbf{Type} & \textbf{Template}\\
        \midrule
        \multirow{15.0}{*}{Russian} & Section & {\fontencoding{T2A}\selectfont \textbf{Сосредоточьтесь на}} \textcolor[HTML]{1400AD}{\{title\}} {\fontencoding{T2A}\selectfont \textbf{и объясните}} \textcolor[HTML]{AD0000}{\{section\}}. \\
         & Subsection & {\fontencoding{T2A}\selectfont \textbf{В контексте }} \textcolor[HTML]{1400AD}{\{title\}} {\fontencoding{T2A}\selectfont \textbf{объясните }} \textcolor[HTML]{AD0000}{\{subsection\}} {\fontencoding{T2A}\selectfont \textbf{и}} \textcolor[HTML]{AD0000}{\{section\}}. \\
        & Sub subsection & {\fontencoding{T2A}\selectfont \textbf{Сосредоточившись на }}\textcolor[HTML]{AD0000}{\{section\}} {\fontencoding{T2A}\selectfont \textbf{в }}\textcolor[HTML]{1400AD}{\{title\}}, {\fontencoding{T2A}\selectfont \textbf{объясните}} \textcolor[HTML]{AD0000}{\{subsubsection\}} {\fontencoding{T2A}\selectfont \textbf{о}} \textcolor[HTML]{AD0000}{\{subsection\}}. \\
        \cmidrule{2-3}
        & \textbf{Template 2} & \\
        & Section & {\fontencoding{T2A}\selectfont \textbf{ Объясните }} \textcolor[HTML]{AD0000}{\{section\}} {\fontencoding{T2A}\selectfont \textbf{этого произведения искусства}}, \textcolor[HTML]{1400AD}{\{title\}}. \\
        & Subsection & {\fontencoding{T2A}\selectfont \textbf{ Объясните }}\textcolor[HTML]{AD0000}{\{subsection\}} {\fontencoding{T2A}\selectfont \textbf{о}} \textcolor[HTML]{AD0000}{\{section\}} {\fontencoding{T2A}\selectfont \textbf{этого произведения искусства}}, \textcolor[HTML]{1400AD}{\{title\}}. \\
\bottomrule
    \end{tabularx}
    \caption{Prompt Templates in Russian}
    \label{tab:template-ru}
\end{table*}